\documentclass[journal,twoside]{IEEEtran}
\usepackage[T1]{fontenc}
\usepackage[utf8]{inputenc}
\usepackage{microtype}
\usepackage{amsmath,amssymb}
\usepackage{graphicx}
\usepackage[table]{xcolor}
\usepackage{booktabs}
\usepackage{multirow}
\usepackage{array}
\usepackage{tabularx}
\usepackage{longtable}
\usepackage{caption}
\usepackage{subcaption}
\usepackage{tikz}
\usepackage{url}
\usepackage[breaklinks]{hyperref}
\usepackage{balance}

\definecolor{accent}{HTML}{2F5D7C}
\definecolor{wash}{HTML}{F2F4F6}
\definecolor{rule}{HTML}{B9C2C9}
\definecolor{warn}{HTML}{8C4A32}

\hypersetup{colorlinks=true, linkcolor=black, citecolor=black, urlcolor=accent}

\newcolumntype{L}[1]{>{\raggedright\arraybackslash}p{#1}}
\newcolumntype{C}[1]{>{\centering\arraybackslash}p{#1}}
\newcommand{\hdr}[1]{\textbf{#1}}
\newcommand{\na}{\textcolor{black!35}{--}}

\makeatletter
\newcommand{\licensed}[4][]{%
  \par\vspace{1pt}{\scriptsize\textcolor{black!55}{%
    #2, \href{#4}{#3}%
    \def\lic@changes{#1}\ifx\lic@changes\@empty.\else; #1.\fi}}}
\makeatother

\makeatletter
\g@addto@macro{\UrlBreaks}{\UrlOrds}
\makeatother

\newcommand{\plateGraspPenetration}{%
  \begin{figure}[t]
    \centering
    \includegraphics[width=0.86\columnwidth]{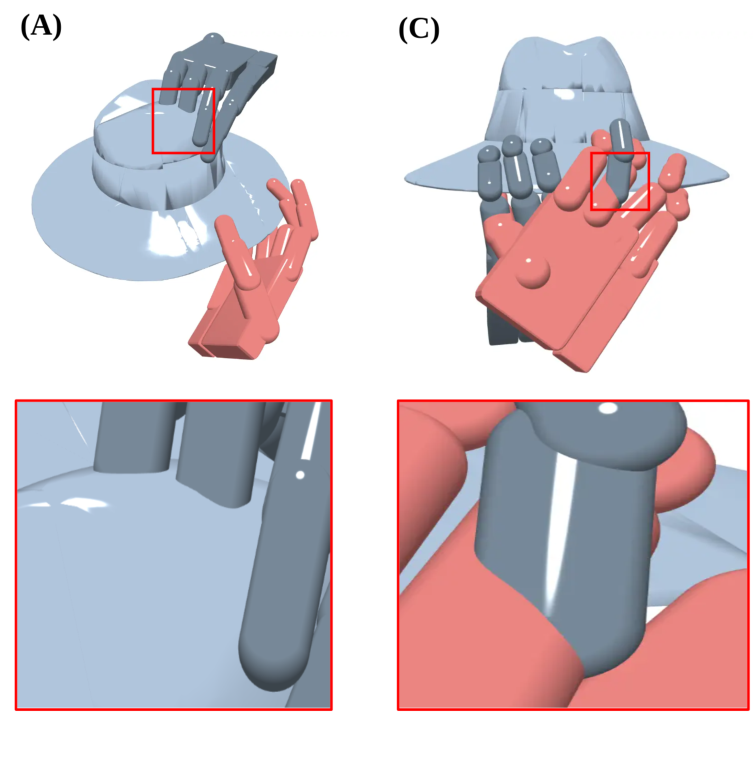}
    \caption{Two failures of synthesised two-hand grasps, each enlarged at the
      contact: (A) a finger inside the object, (C) a finger inside the other hand.
      Both look plausible whole~\cite{bimangrasp_2024}.}
    \label{fig:bimanual-penetration}
    \licensed[panels (A) and (C) of Fig.~9]{Fig.~9 of Shao and Xiao, arXiv:2411.15903}{CC BY 4.0}{https://creativecommons.org/licenses/by/4.0/}
  \end{figure}}

\begin{document}

\title{Learning Dexterous Manipulation:\\
Hands, Simulators, Training, and Evaluation}

\author{%
  Luai Abuelsamen\\[2pt]%
  \normalsize Independent Researcher\\[2pt]%
  \normalsize\texttt{luai\_abuelsamen@berkeley.edu}%
  \thanks{Correspondence: \texttt{luai\_abuelsamen@berkeley.edu}. This work was done
  independently. The full per-paper tabulation behind the tables in this survey is available from
  the author on request.}%
}

\markboth{Preprint, October 2026}{Abuelsamen: Learning Dexterous Manipulation}
\maketitle

\begin{abstract}
A dexterous hand can change an object's pose without setting it down. This survey covers 113
learning-based methods that attempt it, the hands they run on, the simulators they train in,
and how they report results. The field has largely converged on one recipe: reinforcement
learning on privileged state in a GPU simulator, distilled into a vision-only student for
transfer to hardware. Almost half the corpus trains that way, nearly all of it in one
simulator, whose own successors are barely used. A second and faster-growing branch learns from
teleoperated demonstration instead, through vision-language-action models, diffusion policies,
and systems whose contribution is the collection rig itself. Published work has tripled since
2022. Grasping remains the largest task family and bimanual coordination the fastest-rising,
and almost every experiment runs on one of four hands. What the survey adds comes from reading
released code and vendor specifications beside the papers. Eight hands that can be bought or
built from published designs appear in no method paper here. Where the depth a hand sinks into the object it holds is reported at all, it is not measured in the physics that produced the motion. Section VII proposes a protocol for the unreported quantities; the body
gives the counts behind every claim above.

\end{abstract}


\IEEEpeerreviewmaketitle

\section{Introduction}
\label{sec:intro}

\makeatletter
\@ifundefined{fnLoaded}{\IfFileExists{figs/finding_numbers.tex}{\providecommand{\fnLoaded}{}
\providecommand{\fnRows}{113}
\providecommand{\fnCode}{63}
\providecommand{\fnDisagree}{39}
\providecommand{\fnKept}{8}
\providecommand{\fnOtherClass}{31}
\providecommand{\fnParse}{14}
\providecommand{\fnAbsent}{8}
\providecommand{\fnSkew}{5}
\providecommand{\fnInternal}{4}
\providecommand{\fnPenSettled}{97}
\providecommand{\fnPenHandled}{12}
\providecommand{\fnPenClosed}{5}
\providecommand{\fnPenOther}{7}
\providecommand{\fnPenRollout}{0}
\providecommand{\fnTrials}{39}
\providecommand{\fnTrialsMedian}{15}
\providecommand{\fnTrialsMode}{10}
\providecommand{\fnTrialsModeN}{16}
\providecommand{\fnTotals}{24}
\providecommand{\fnTotalsMin}{12}
\providecommand{\fnTotalsMax}{750}
\providecommand{\fnUnclear}{7}
}{}}{}
\makeatother

\begin{figure*}[!t]
  \centering
  \IfFileExists{figs/fig_field.tex}{\begin{tikzpicture}[
  font=\scriptsize,
  bar/.style={draw=none,fill=black!62},
  barlight/.style={draw=none,fill=black!22},
  baracc/.style={draw=none,fill=accent},
  box/.style={draw=black!45,line width=0.3pt,inner sep=3pt,align=left},
  ghost/.style={draw=black!35,line width=0.3pt,dash pattern=on 1.4pt off 1.4pt,inner sep=3pt,align=left},
  lnk/.style={draw=black!38,line width=0.28pt},
  lbl/.style={font=\scriptsize},
  num/.style={font=\scriptsize\color{black!55}},
]

\draw[draw=black!42,line width=0.49pt] (2.64,-6.21) .. controls (3.19,-6.21) and (3.29,-6.14) .. (3.84,-6.14);
\draw[draw=black!42,line width=0.49pt] (6.48,-6.14) .. controls (7.03,-6.14) and (7.13,-1.59) .. (7.68,-1.59);
\draw[draw=black!42,line width=0.49pt] (10.32,-7.07) .. controls (10.87,-7.07) and (10.97,-6.62) .. (11.52,-6.62);
\draw[draw=black!44,line width=0.58pt] (6.48,-3.15) .. controls (7.03,-3.15) and (7.13,-7.07) .. (7.68,-7.07);
\draw[draw=black!45,line width=0.62pt] (2.64,-5.91) .. controls (3.19,-5.91) and (3.29,-5.10) .. (3.84,-5.10);
\draw[draw=black!45,line width=0.62pt] (6.48,-2.74) .. controls (7.03,-2.74) and (7.13,-6.26) .. (7.68,-6.26);
\draw[draw=black!47,line width=0.67pt] (10.32,-1.69) .. controls (10.87,-1.69) and (10.97,-6.26) .. (11.52,-6.26);
\draw[draw=black!47,line width=0.67pt] (10.32,-6.26) .. controls (10.87,-6.26) and (10.97,-1.39) .. (11.52,-1.39);
\draw[draw=black!48,line width=0.71pt] (6.48,-4.71) .. controls (7.03,-4.71) and (7.13,-4.50) .. (7.68,-4.50);
\draw[draw=black!48,line width=0.71pt] (6.48,-5.23) .. controls (7.03,-5.23) and (7.13,-5.35) .. (7.68,-5.35);
\draw[draw=black!51,line width=0.85pt] (6.48,-2.33) .. controls (7.03,-2.33) and (7.13,-4.12) .. (7.68,-4.12);
\draw[draw=black!55,line width=0.98pt] (10.32,-0.90) .. controls (10.87,-0.90) and (10.97,-2.77) .. (11.52,-2.77);
\draw[draw=black!55,line width=0.98pt] (14.16,-7.49) .. controls (14.71,-7.49) and (14.81,-4.29) .. (15.36,-4.29);
\draw[draw=black!56,line width=1.03pt] (14.16,-6.44) .. controls (14.71,-6.44) and (14.81,-3.89) .. (15.36,-3.89);
\draw[draw=black!57,line width=1.07pt] (10.32,-1.30) .. controls (10.87,-1.30) and (10.97,-5.34) .. (11.52,-5.34);
\draw[draw=black!57,line width=1.07pt] (10.32,-3.23) .. controls (10.87,-3.23) and (10.97,-4.12) .. (11.52,-4.12);
\draw[draw=black!58,line width=1.12pt] (14.16,-1.39) .. controls (14.71,-1.39) and (14.81,-5.52) .. (15.36,-5.52);
\draw[draw=black!60,line width=1.16pt] (14.16,-4.12) .. controls (14.71,-4.12) and (14.81,-3.11) .. (15.36,-3.11);
\draw[draw=black!61,line width=1.21pt] (14.16,-5.34) .. controls (14.71,-5.34) and (14.81,-3.50) .. (15.36,-3.50);
\draw[draw=black!64,line width=1.34pt] (6.48,-4.19) .. controls (7.03,-4.19) and (7.13,-1.00) .. (7.68,-1.00);
\draw[draw=black!66,line width=1.38pt] (6.48,-1.91) .. controls (7.03,-1.91) and (7.13,-2.93) .. (7.68,-2.93);
\draw[draw=black!67,line width=1.43pt] (2.64,-2.10) .. controls (3.19,-2.10) and (3.29,-2.12) .. (3.84,-2.12);
\draw[draw=black!70,line width=1.56pt] (14.16,-2.77) .. controls (14.71,-2.77) and (14.81,-2.71) .. (15.36,-2.71);
\draw[draw=black!75,line width=1.74pt] (2.64,-2.89) .. controls (3.19,-2.89) and (3.29,-4.32) .. (3.84,-4.32);
\draw[draw=black!77,line width=1.83pt] (2.64,-4.58) .. controls (3.19,-4.58) and (3.29,-2.95) .. (3.84,-2.95);
\draw[draw=black!77,line width=1.83pt] (10.32,-2.64) .. controls (10.87,-2.64) and (10.97,-0.60) .. (11.52,-0.60);
\draw[draw=black!85,line width=2.10pt] (14.16,-0.60) .. controls (14.71,-0.60) and (14.81,-2.32) .. (15.36,-2.32);
\node[font=\footnotesize\bfseries,anchor=south] at (1.32,0.48) {data source};
\node[font=\scriptsize\color{black!60},anchor=south] at (1.32,0.10) {Sec.~\ref{sec:train:human}};
\draw[draw=black!45,line width=0.4pt] (0.00,0.00) -- (2.64,0.00);
\draw[box,fill=white] (0.00,-1.50) rectangle (2.64,-3.49);
\node[lbl,anchor=west,align=left] at (0.10,-2.49) {human data\\named};
\node[num,anchor=east] at (2.54,-2.49) {54};
\draw[box,fill=white] (0.00,-3.65) rectangle (2.64,-5.52);
\node[lbl,anchor=west,align=left] at (0.10,-4.58) {no\\demonstrations};
\node[num,anchor=east] at (2.54,-4.58) {50};
\draw[ghost,fill=white] (0.00,-5.68) rectangle (2.64,-6.43);
\node[lbl,anchor=west,align=left] at (0.10,-6.06) {not stated};
\node[num,anchor=east] at (2.54,-6.06) {9};
\node[font=\footnotesize\bfseries,anchor=south] at (5.16,0.48) {embodiment};
\node[font=\scriptsize\color{black!60},anchor=south] at (5.16,0.10) {Sec.~\ref{sec:hands},~\ref{sec:bimanual}};
\draw[draw=black!45,line width=0.4pt] (3.84,0.00) -- (6.48,0.00);
\draw[box,fill=white] (3.84,-1.50) rectangle (6.48,-3.57);
\node[lbl,anchor=west,align=left] at (3.94,-2.53) {one hand};
\node[num,anchor=east] at (6.38,-2.53) {57};
\draw[box,fill=white] (3.84,-3.73) rectangle (6.48,-5.69);
\node[lbl,anchor=west,align=left] at (3.94,-4.71) {two hands};
\node[num,anchor=east] at (6.38,-4.71) {53};
\draw[ghost,fill=white] (3.84,-5.85) rectangle (6.48,-6.43);
\node[lbl,anchor=west,align=left] at (3.94,-6.14) {not stated};
\node[num,anchor=east] at (6.38,-6.14) {3};
\node[font=\footnotesize\bfseries,anchor=south] at (9.00,0.48) {simulator};
\node[font=\scriptsize\color{black!60},anchor=south] at (9.00,0.10) {Sec.~\ref{sec:simulators}};
\draw[draw=black!45,line width=0.4pt] (7.68,0.00) -- (10.32,0.00);
\draw[ghost,fill=white] (7.68,-0.55) rectangle (10.32,-2.04);
\node[lbl,anchor=west,align=left] at (7.78,-1.30) {not stated};
\node[num,anchor=east] at (10.22,-1.30) {36};
\draw[box,fill=white] (7.68,-2.20) rectangle (10.32,-3.67);
\node[lbl,anchor=west,align=left] at (7.78,-2.93) {Isaac Gym};
\node[num,anchor=east] at (10.22,-2.93) {35};
\draw[box,fill=white] (7.68,-3.83) rectangle (10.32,-4.79);
\node[lbl,anchor=west,align=left] at (7.78,-4.31) {MuJoCo};
\node[num,anchor=east] at (10.22,-4.31) {17};
\draw[box,fill=white] (7.68,-4.95) rectangle (10.32,-5.76);
\node[lbl,anchor=west,align=left] at (7.78,-5.35) {other};
\node[num,anchor=east] at (10.22,-5.35) {11};
\draw[box,fill=white] (7.68,-5.92) rectangle (10.32,-6.61);
\node[lbl,anchor=west,align=left] at (7.78,-6.26) {Isaac Lab or\\Sim};
\node[num,anchor=east] at (10.22,-6.26) {7};
\draw[box,fill=white] (7.68,-6.77) rectangle (10.32,-7.38);
\node[lbl,anchor=west,align=left] at (7.78,-7.07) {SAPIEN};
\node[num,anchor=east] at (10.22,-7.07) {4};
\node[font=\footnotesize\bfseries,anchor=south] at (12.84,0.48) {training paradigm};
\node[font=\scriptsize\color{black!60},anchor=south] at (12.84,0.10) {Sec.~\ref{sec:training}};
\draw[draw=black!45,line width=0.4pt] (11.52,0.00) -- (14.16,0.00);
\draw[box,fill=white] (11.52,-0.00) rectangle (14.16,-1.99);
\node[lbl,anchor=west,align=left] at (11.62,-0.99) {RL};
\node[num,anchor=east] at (14.06,-0.99) {54};
\draw[box,fill=white] (11.52,-2.15) rectangle (14.16,-3.39);
\node[lbl,anchor=west,align=left] at (11.62,-2.77) {BC};
\node[num,anchor=east] at (14.06,-2.77) {27};
\draw[box,fill=white] (11.52,-3.55) rectangle (14.16,-4.68);
\node[lbl,anchor=west,align=left] at (11.62,-4.12) {distillation};
\node[num,anchor=east] at (14.06,-4.12) {23};
\draw[box,fill=white] (11.52,-4.84) rectangle (14.16,-5.84);
\node[lbl,anchor=west,align=left] at (11.62,-5.34) {VLA};
\node[num,anchor=east] at (14.06,-5.34) {18};
\draw[box,fill=white] (11.52,-6.00) rectangle (14.16,-6.88);
\node[lbl,anchor=west,align=left] at (11.62,-6.44) {teleop-system};
\node[num,anchor=east] at (14.06,-6.44) {14};
\draw[box,fill=white] (11.52,-7.04) rectangle (14.16,-7.93);
\node[lbl,anchor=west,align=left] at (11.62,-7.49) {diffusion};
\node[num,anchor=east] at (14.06,-7.49) {14};
\node[font=\footnotesize\bfseries,anchor=south] at (16.68,0.48) {evaluation};
\node[font=\scriptsize\color{black!60},anchor=south] at (16.68,0.10) {Sec.~\ref{sec:evaluation}};
\draw[draw=black!45,line width=0.4pt] (15.36,0.00) -- (18.00,0.00);
\draw[box,fill=white] (15.36,-1.83) rectangle (18.00,-4.78);
\node[lbl,anchor=west,align=left] at (15.46,-3.30) {real robot};
\node[num,anchor=east] at (17.90,-3.30) {89};
\draw[box,fill=white] (15.36,-4.94) rectangle (18.00,-6.10);
\node[lbl,anchor=west,align=left] at (15.46,-5.52) {simulation only};
\node[num,anchor=east] at (17.90,-5.52) {24};
\end{tikzpicture}}{%
    \framebox[0.7\textwidth]{\rule{0pt}{2.2cm}\textit{figs/fig\_field.tex pending}}}
  \caption{The field on one page, and the map of this survey: nodes sized by corpus papers, edges
  drawn for the routes that carry at least two fifths of the smaller endpoint, so the dominant
  path reads as a path, each column head naming the section that covers it.}
  \label{fig:field}
\end{figure*}
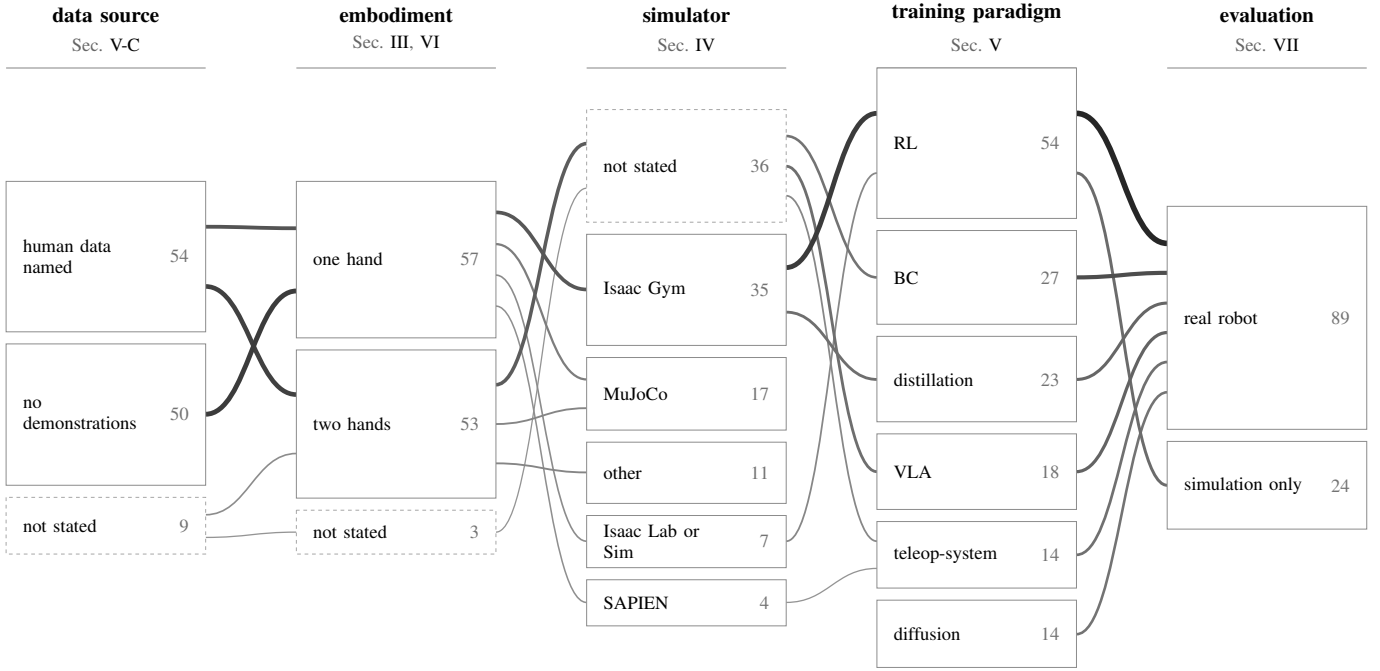

A hand is dexterous when it can change an object's pose without putting the object down. That
property, and not the finger count, is what separates a hand from a gripper. Bicchi
\cite{bicchi_grasping_chapter_2001} draws the same line in Sec.~1.1 between restraining an
object and manipulating it with the fingers: restraint is a static question about whether the
contacts prevent motion, while in-hand manipulation is a dynamic one about whether contacts can
be broken and remade while the object stays held. A parallel-jaw gripper can hold almost
anything and reorient almost nothing.

The classical theory of grasping is about holding, and it is good at it. Given a set of contact
points and an assumption about friction, it can decide whether an object is immobilized, and
what force each finger has to apply to keep it that way. What it cannot do is say what the
fingers should do next. Contact is what makes that hard: a finger is either touching or it is
not, and it can push but not pull, so the equations of motion change form every time a contact
is made or broken. The standard treatment says as much about its own results, that immobilizing
an object ``does not guarantee stability'', and that ``the nonsmooth nature of grasp dynamics,
because of the unilateral constraints on displacements and forces, has made a thorough analysis
very difficult'' (\cite[Sec.~1.5 and 1.8]{bicchi_grasping_chapter_2001}). The tradition
produced tests. You can check a candidate grasp against them; you cannot run them on a robot.

Learning changed what a practitioner has to produce. The old deliverable was a proof that a
configuration holds. The new one is a policy that scores well over many simulated attempts,
which means the simulator now stands in for the analysis and the score stands in for the proof.
That is why the field moved, and it is also why this survey is organized as it is. It shifts
two burdens onto the author. The simulator has to be faithful where fidelity matters. The
measurement has to mean something. Much of what follows asks how well this literature carries
those two.

One recipe now carries most of the field. Train a policy with reinforcement learning on
privileged state, the object pose, velocities and physical parameters the robot cannot sense, in
a GPU simulator running thousands of environments in parallel; distill that teacher into a
student that sees only what a camera and joint encoders provide; transfer the student.
Fifty-four of the 113 method papers here train this way, most of them in one simulator, and
about half of those add the distillation stage. In-hand cube reorientation is the recipe's
benchmark, inherited from \cite{openai_dexterity_2018} and still the task a new method is
expected to show.

A second branch is growing underneath it and inverts the dependency. Instead of a simulator and
a reward it takes human demonstrations through a teleoperation rig and fits a policy directly,
as a vision-language-action model, a diffusion policy, or a data-collection rig whose design is
itself the contribution. Section~\ref{sec:training} counts them, and
\figurename~\ref{fig:field} shows how fast they have arrived. It is the newer literature, most
of it from the last two years, and it trades the reward-design problem for a data-collection
problem. Which of the two branches ends up feeding the other is not settled here, but the growth
is the clearest trend in the corpus.

Both branches run on remarkably little hardware. Four hands carry almost every experiment in
this corpus, led by the Allegro at 35 of the 113 papers; \figurename~\ref{fig:hands} gives the
rest of the distribution. Eight further hands can be bought today or built from published designs
and appear in the experiments of none of them. The hardware exists and is purchasable, so this is
not a manufacturing lag, and Section~\ref{sec:hands} takes it up: a field whose results rest on
four platforms is more fragile than its publication volume suggests.

What this literature cannot currently do is compare its own results. Nearly every paper states
a success criterion, but they are not the same criterion, and only 70 of the 113 say how many
trials a reported rate rests on. The criteria run from never dropping the object to
reaching a pose within a tolerance each paper picks for itself. Two success rates from two
papers on nominally the same task are usually not measuring the same event, and none of the
surveys preceding this one supplies a protocol that would make them comparable.
Section~\ref{sec:evaluation} derives one, with the trial counts a stated confidence interval
actually requires.

Every number in this survey is traceable to a stated source. Appendix~\ref{app:method} says how
the papers were selected, read and coded, what the method cannot do, and why each coverage figure
here is a lower bound.

One part of that method is unusual enough to flag now. Where a paper published a repository, its
reward was read from the repository as well as from the paper, so the two can be compared.
Reading an implementation against its own
description is established practice outside robotics, in repeatability and reproducibility
studies \cite{collberg_repeatability_2016, raff_reproducibility_2019}, in audits of how
implementation detail moves reinforcement-learning results
\cite{engstrom_implementation_matters_2020}, in reward-design review
\cite{knox_reward_misdesign_2023}, in benchmark re-releases \cite{metaworld_plus_2025}, and
most directly in recent work on paper--code alignment \cite{scicoqa_2026, biocon_2026}. We are
aware of none that does it for a robotics field's released reward functions.
Section~\ref{sec:train:code} reports what this one found.

The scope is single-hand and bimanual multi-fingered manipulation under learned control.
Parallel-jaw work enters as a comparison and not as a subject. Two limits apply to every number
here. The corpus is of the learned era, which is a selection and not a verdict on the analytic
tradition; and where this survey failed to extract a value, the counts look the same as if the
source had never stated one.

\subsection{How to read this survey}
\label{subsec:howtoread}

\figurename~\ref{fig:field} puts the field on one page, and the sections work outward from the
task. Section~\ref{sec:taxonomy} sets out the task families, what makes each one hard, and the
handful of terms this survey uses precisely. Section~\ref{sec:hands} covers the hands and their
sensing, Section~\ref{sec:simulators} the simulators and what they expose about contact,
Section~\ref{sec:training} how policies are actually trained, Section~\ref{sec:bimanual} what
changes when there are two hands, and Section~\ref{sec:evaluation} how any of it is evaluated.
Section~\ref{sec:gaps} says what the corpus supports and what it does not. The appendices hold
the method and the full tabulation: \ref{app:method} the method and its limits,
\ref{app:tables} the hardware and engine tables, \ref{app:rewards} every paper-against-code
comparison, \ref{app:surveys} the surveys that precede this one, \ref{app:protocol} the
derivation behind the protocol.

If you are new to the field, read Section~\ref{sec:taxonomy} for the vocabulary, then the first
half of Section~\ref{sec:training} for the two ways a dexterous policy gets trained, then
Section~\ref{sec:hands} to see what hardware any of it runs on. Those three give the shape of the
field in about half an hour. Section~\ref{sec:evaluation} is the one to read before designing an
experiment, because it is about what the field's numbers do and do not mean.

If you already work here, the parts likely to tell you something you did not know are
Section~\ref{sec:sim:engines} on what simulators expose about contact,
Section~\ref{sec:train:code} on papers against their own released code,
Section~\ref{sec:bimanual} on bimanual work as its own problem, and
Section~\ref{subsec:protocol} on what a success rate at twenty trials can support.

If you want the data, the tables carry it: Tables~\ref{tab:hands-available}
and~\ref{tab:hands-announced} for the hands, Table~\ref{tab:simulators} for the simulators,
Table~\ref{tab:codegap} for the paper-against-code comparisons and Table~\ref{tab:protocol} for
the proposed protocol. The full per-paper tabulation is available from the author on request.

Three things to carry while reading. The corpus is of the learned era, which is a selection and
not a verdict on the analytic tradition that preceded it. Every coverage figure is a floor: when a
paper is silent and when this survey failed to extract a value, the counts look the same, and
Appendix~\ref{app:method} says which direction that bias runs. And a success rate in this
literature is rarely comparable across two papers, for reasons Section~\ref{sec:evaluation} sets
out, so prefer the sections that explain mechanisms to any ranking assembled from the numbers.
\section{A taxonomy of the problem}
\label{sec:taxonomy}

\subsection{Task families}
\label{sec:taxonomy:tasks}

Dexterous manipulation is not one task, and what separates its tasks is what makes some of them
hard: what is held, what moves, which way gravity points, and what counts as failure. Six
families cover most of the corpus, and the last paragraph here says what they miss. Grasping is
the largest, with 58 of the 113 papers carrying the label; the labels are not exclusive, so one
paper can sit in several families. Success is a lift that survives a hold, and the thresholds
differ by more than an order of magnitude: one paper requires the object ``held 10 cm above the
table for 20 s'' \cite{dexgraspvla_2025}, another ``held at least 0.5 s in simulation''
\cite{omnigrasp_2024}. What makes grasping hard at scale is the continuum of starting
configurations. UniDexGrasp++
\cite{unidexgrasp_pp_2023} states it in Sec.~4.3, that ``we are dealing with an infinite number
of tasks considering the initial object pose can change continuously''.

Functional and tool use is second with 45 papers. It is the family where a stable grasp can
still be the wrong answer. Dexterous Functional Grasping
\cite{dexterous_functional_grasping_2023} gives the case in Sec.~2.1, that ``grabbing a hammer
from the head or handle are both equally valid ways of using it'', and only one of the two lets
the tool be used. Success is defined against the tool's function, and the field has no shared
way to state it.

In-hand reorientation has 31 papers and the most settled criteria, because the community
inherited one number: OpenAI declared the goal achieved below 0.4 rad of orientation error
\cite{openai_dexterity_2018}, and its successors keep that tolerance or tighten it
\cite{dextreme_2022,eureka_2023}. A shared tolerance is not a shared protocol, because the stopping
rule differs. Visual Dexterity \cite{visual_dexterity_2022} measures error ``when the controller
predicts it has reached the goal and stops'', which makes the policy a judge of its own trial.

Tracking a human reference has 12 papers and inverts the problem. The target is a trajectory
rather than an endpoint, so success is a per-frame error band, and the bands are compound: one
method requires rotation, translation, joint and fingertip error all to stay inside separate
thresholds at once \cite{maniptrans_2025}. The difficulty is that the reference came from a human
hand and was never dynamically feasible for the robot. DexMachina \cite{dexmachina_2025} warns in
App.~B.4 that scoring by timesteps inside the thresholds makes ``the results highly sensitive to
the threshold values''.

Bimanual coordination has 43 papers and handover has 10. Handover is the smallest family and
the one whose failure has a single moment, because the giver must release only after the
receiver has the object. Dynamic Handover \cite{dynamic_handover_2023} reports a hit rate above
its success rate and blames the gap in Sec.~5.4 on ``occasional challenges encountered during
the grasping phase of the catcher''. Not every paper here learns both sides. In
\cite{dexterous_handover_2025} only the receiver is learned and the giver is a scripted arm.

Those six families do not cover the corpus. Twenty-four of the 113 papers carry a label from
outside them and eight carry no label from the six at all. Twenty fall in a catch-all class,
mostly generalist policies evaluated on a general task suite, among them
\cite{pi0_2024,pi05_2025,pistar06_2025,openvla_2024} and \cite{gemini_robotics_15_2025}. Three
adjacent families are named here instead of being folded into the six. Locomanipulation is
\cite{humanplus_2024,omnih2o_2024} and \cite{groot_n16_2025}, where the base is not fixed and
the gravity argument below changes character. Piano playing is \cite{robopianist_2023,rp1m_2024}
and \cite{pianomime_2024}, discussed in Section~\ref{sec:bimanual}, where success is a
per-timestep F1 against a MIDI score. Deformable manipulation is \cite{dexdeform_2023}, whose
object carries its own state and its own physics. Ferrari and Canny's grasp-quality measure
\cite{ferrari_canny_1992} carries no task label, because it is a measure rather than a task.
Table~\ref{tab:tasks}'s six families are the ones with enough papers to compare, and not a
partition of the corpus.

\subsection{Sources of difficulty}
\label{sec:taxonomy:hard}

Contact is non-smooth, and that is a property of the problem rather than of any solver. Bicchi
\cite{bicchi_grasping_chapter_2001} states in Sec.~1.2 that contact constraints are unilateral,
and that losing a contact ``involves an abrupt change of the structure of the model under
consideration''. Its Sec.~1.4 gives the sharper version, that a rod sliding on rough ground has
configurations with no consistent solution and configurations with more than one. Simulators
inherit the difficulty. The re-implementation study \cite{contact_models_comparison_2023}
concludes in Sec.~V that ``there is no fully satisfactory approach at the moment, as all existing
solutions compromise either accuracy, robustness, or efficiency''. Pang et al.
\cite{pang_global_planning_2022} treat non-smoothness as the thing to remove, curating every
contact pair in Sec.~III-D so that contact points and normals change smoothly.

The hand has more actuators than the object has degrees of freedom and is still short of authority
over it. The object moves only through contacts, and the contacts are what cannot be commanded.
Bicchi \cite{bicchi_grasping_chapter_2001} states in Sec.~1.4 that in multifingered grasps ``the
number of independent contact forces is much larger than the number of actuators. Thus, from a
controllability standpoint, not all the contact forces are controllable''.

Vision is occluded by the hand doing the work. Bai et al. \cite{bai_unified_manip_survey_2025}
name it in Sec.~4.3, that ``occlusion hampers object tracking''. DexPoint \cite{dexpoint_2022}
keeps vision and adds imagined hand points to the point cloud. Yin et al.
\cite{rotating_without_seeing_2023} remove vision entirely and rotate objects from 16 binary touch
sensors over the palm, links and fingertips.

Gravity direction changes the task. Visual Dexterity
\cite{visual_dexterity_2022} states that reorientation with the hand below the object ``is much
easier'', because ``with a downward-facing hand, the hand must manipulate the object while
simultaneously counteracting gravity''. AnyRotate \cite{anyrotate_2024} makes the direction a
randomised variable by ``randomly initializing hand orientations between episodes'', and its
Sec.~5.3 measures the cost, with performance dropping progressively from palm up and palm down,
through base up and base down, to thumb up and thumb down.

\label{sec:taxonomy:twohands}
Three counts describe
two hands and they measure different things. Fifty-three of the 113 papers record two
hands on the robot, which is all the corpus's two-hand flag claims. Forty-three carry the task
label for bimanual coordination, the narrower claim that coordinating the hands is the task.
Section~\ref{sec:bimanual} narrows again, to the 28 papers whose notes place a learned
closed-loop controller on two multi-fingered hands, and its opening paragraph names every
exclusion that takes the 53 down to the 28. That 28 is the denominator for every architecture
count in this survey. Every bimanual claim here names which of the three it uses.

Four things genuinely change when the second hand arrives.
Contact stays non-smooth, occlusion stays, and gravity stays the same problem.

Role asymmetry is one. AsymDex \cite{asymdex_2024} assigns a dominant hand with full finger and
wrist control and a facilitating hand with 6-DoF base pose only, so that ``the facilitating hand
repositions and reorients one object, while the dominant hand performs complex manipulations''.

Two hands on one object also close a kinematic chain through the object. In
\cite{bicchi_grasping_chapter_2001} Sec.~1.4 the hand and object dynamics are separate and are
``linked through the n rigid-body contact constraints''. A second hand adds a second constraint
set on the same object rather than a second independent problem. That is where physical
plausibility fails first. BimanGrasp \cite{bimangrasp_2024} is the one paper here that measures it,
rejecting a synthesised grasp when ``total penetrations exceeded 1.5 mm'' and reporting in
Sec.~IV-C that ``penetration remains the primary cause of grasp failure''.

Two-arm collision does not exist for one hand at all. DyDexHandover \cite{dydexhandover_2025}
resets an episode when ``any unintended arm contact with the environment or self-collisions''
occurs. Bunny-VisionPro \cite{bunny_visionpro_2024} pays for it in the controller, where adding a
sphere-approximated self-collision cost raises motion-control time from 0.74 ms to 7.85 ms. Bidex
\cite{bidex_teleop_2024} solves it in hardware, mounting the hands so the human arm and the
teacher arm ``are perpendicular to each other and do not collide''. DexMimicGen
\cite{dexmimicgen_2024} states plainly that it does not handle inter-arm collision at all.

The action space doubles as well. Bi-DexHands \cite{bidexhands_2022} runs 20 tasks on two Shadow
Hands and finds single-agent PPO beating multi-agent RL on most of them, offering in Sec.~5.2
that ``PPO algorithm is able to use all observations for training the policy, while MARL can
only use partial observations''. AsymDex \cite{asymdex_2024} attacks the dimensionality from the
other side, halving the observation and action dimension through its role split and a frame
relative to the facilitating hand's object.

\begin{table*}[!t]
  \caption{Task families against the properties that define success.}
  \label{tab:tasks}
  \centering\scriptsize
  \begin{tabular}{@{}L{0.085\textwidth}L{0.150\textwidth}L{0.150\textwidth}L{0.145\textwidth}L{0.130\textwidth}L{0.240\textwidth}@{}}
    \toprule
    \rowcolor{wash} \hdr{task family} & \hdr{what is held} & \hdr{what moves} & \hdr{what gravity does} & \hdr{what counts as failure} & \hdr{shortest success criterion as stated} \\
    \midrule
    In-hand reorientation &
    one rigid object, held by the fingers for the whole episode &
    the object's orientation in the palm frame, while contacts break and remake &
    a disturbance whose direction is set by hand orientation, and a load to counteract with the palm down, \cite{visual_dexterity_2022} &
    the object leaves the hand, e.g.\ the fall reset at goal distance 0.24 m in \cite{dextreme_2022} &
    orientation error below a stated tolerance under a stated stopping rule, 0.4 rad in \cite{openai_dexterity_2018} and \cite{dextreme_2022}, 0.1 rad in \cite{eureka_2023} \\
    \addlinespace[2.5pt]
    Grasping &
    nothing at the start, the object rests on a support &
    the hand closes, then the object is lifted clear &
    the test itself, since the grasp exists to resist it &
    no lift, or a drop inside the hold window &
    a stated lift height held for a stated duration, 10 cm for 20 s in \cite{dexgraspvla_2025}, 10 cm to episode end in \cite{graspxl_2024}, 0.5 s in \cite{omnigrasp_2024} \\
    \addlinespace[2.5pt]
    Functional and tool use &
    a tool, held in the pose its function requires rather than the most stable one &
    the tool's own joint, or the tool acting on a second object &
    secondary, the function dominates &
    a stable grasp in the wrong place, \cite{dexterous_functional_grasping_2023} &
    no agreed criterion. Criteria range from never dropping the object, \cite{dexterous_functional_grasping_2023}, to a four-point rubric, \cite{dexvla_2025}. The one mechanical threshold is holding the tool joint past a commanded angle for the episode, \cite{articulated_tools_inhand_2025} \\
    \addlinespace[2.5pt]
    Tracking a human reference &
    whatever the human held in the captured clip &
    hand and object follow a recorded trajectory under physics &
    full, and the reference never obeyed it, so the reference is not dynamically feasible &
    per-frame error leaves the band, or the object drops &
    all stated per-frame errors below threshold for the whole clip, 30 deg and 3 cm and 8 cm joint and 6 cm fingertip in \cite{maniptrans_2025}. Thresholds are not comparable across papers, and DexMachina \cite{dexmachina_2025} calls the results ``highly sensitive to the threshold values'' \\
    \addlinespace[2.5pt]
    Bimanual coordination &
    one object by two hands, or one object per hand &
    both hands, the object, and the relative pose of the hands &
    can be carried by one hand while the other acts, \cite{asymdex_2024} &
    the object drops, the arms collide, or the hands work against each other &
    no agreed criterion. Task-specific thresholds dominate, e.g.\ 0.035 m block-to-cup distance in \cite{asymdex_2024}. The one criterion with a physical-plausibility gate is 2.0 s of hold with total penetration under 1.5 mm, \cite{bimangrasp_2024} \\
    \addlinespace[2.5pt]
    Handover and in-hand transfer &
    by the giver at the start, by the receiver at the end &
    the object crosses between hands, released only after it is regained &
    unopposed during the exchange, and total in a thrown transfer, \cite{dynamic_handover_2023} &
    a drop at the moment of release, and the catcher's grasp phase, \cite{dynamic_handover_2023} Sec.~5.4 &
    the receiver holds the object and moves it clear, more than 10 cm from the first hand in \cite{hato_visuotactile_2024}, or holds it to episode end in \cite{dydexhandover_2025} \\
    \bottomrule
  \end{tabular}
\end{table*}

Two cells say the field has no agreed criterion, and both are in families where a second body is
involved. A criterion naming a distance, a duration and a trial count can be re-run by someone
else. A rubric cannot, and Section~\ref{sec:evaluation} takes up what follows from that.

\subsection{Terms used throughout}
\label{subsec:terms}

Six terms recur in every section that follows, and each is used here in one sense only.

\textbf{Method paper.} A paper whose own experiments train or evaluate a dexterous policy. It is
the unit every count is taken over: 113 of the corpus's 219 entries are method papers, the rest
being hands, simulators, datasets, benchmarks, surveys, tactile sensors and evaluation
protocols.

\textbf{Privileged state.} Information a simulator can supply and a robot cannot: object pose and
velocity, contact forces, mass, friction. A policy trained on it cannot be deployed as it stands,
which is what makes the next term necessary.

\textbf{Teacher--student distillation.} Training one policy on privileged state, then fitting a
second that sees only deployable observations, usually camera images and joint encoders, to
reproduce the first. Section~\ref{sec:training} finds 23 of the 113 methods do this, and it is the
second half of the recipe Section~\ref{sec:intro} describes.

\textbf{Reference and rollout.} A reference is a pose or a trajectory scored before anything
executes: a retargeted human motion, a synthesized grasp, a planned path. A rollout is what a
trained policy actually did. The distinction matters because a quantity can be measured carefully
at the reference and never at the rollout, which Section~\ref{sec:evaluation} finds is the case
for interpenetration.

\textbf{Interpenetration.} The distance by which two bodies that should be touching instead
overlap, a consequence of solving contact constraints approximately.
Section~\ref{sec:simulators} covers what each engine exposes; the depth is a property of the
compliance someone chose, not of the engine.

\textbf{Success criterion and per-cell trial count.} The criterion is the predicate a paper
uses to call an episode a success: a pose within a tolerance, an object never dropped, a task
completed. The per-cell count is how many trials one reported rate is computed from.
Section~\ref{sec:evaluation} finds 98 of them state a criterion and no two state the same one,
and that 70 state a count.

\subsection{What has changed since 2022}
\label{subsec:trends}

The corpus roughly tripled between 2022 and 2024, and its composition changed while it grew. Three
shifts are visible in the record, and none of them is small.

Sorting each method by whether it trains in simulation, through reinforcement learning or
distillation, or from human demonstration, through behavior cloning, a vision-language-action
model, a diffusion or flow policy, or a collection rig, the simulator branch led 13 papers to 4
in 2023 and trailed 15 to 18 in 2024. It has trailed since: demonstration-trained work is 22 of
the 34 papers from 2025 and 8 of the 13 so far from 2026. The recipe Section~\ref{sec:intro}
describes is this field's mature pipeline and not its growth area, and a reader who knows only
that recipe is reading the field as it was in 2023.

It was 8 of the 11
papers in 2022 and 3 of the 34 in 2025. The absolute count falls while the corpus triples,
which is not a shift in emphasis but a change in what the field treats as worth publishing.
Bimanual coordination moved the other way over the same span, from one paper to 16, and functional
and tool use from 2 to 15. The cube-rotation benchmark that Section~\ref{sec:training} shows
the reward literature organized itself around is the task the newer literature has left.

No paper earlier than 2025 uses Isaac Lab or Isaac Sim at all. Three do in 2025 and four in the
partial 2026, against nine and two respectively on Isaac Gym, so on small numbers the successors
already carry more of the most recent year than the original does. A survey assembled twelve
months earlier would have recorded a field that had not moved, and the concentration this survey
reports on one simulator is a fact about a corpus with a cutoff, not a permanent property of the
field.
\section{Hands and who makes them}
\label{sec:hands}

\subsection{Design axes}
\label{sec:hands:axes}

A policy is only as dexterous as the hand it runs on, and the hands available divide sharply: a
few a laboratory can buy, a few more it can build from published designs, and a growing number
that exist only in an announcement. What can be said about them is less than it looks. No
degree-of-freedom, force, weight or price figure in Table~\ref{tab:hands-available} or
Table~\ref{tab:hands-announced}, both set in Appendix~\ref{app:tables}, was measured by anyone
outside the maker. Six entries are the exception, from peer-reviewed papers with stated protocols:
ILDA, Pisa/IIT, ORCA, RUKA, LEAP and BiDexHand. Every other claim is dated at its source.
Table~\ref{tab:hands-announced} prints the date beside each announced hand, and the repository
tabulation carries a claim-date column for the hands that can be bought, running from Shadow's
specification of 4 December 2024 to vendor pages fetched on 17 September 2026.

The degree-of-freedom count is the first number a vendor states and the least comparable one,
because three different quantities get printed under that heading: joints, actuated joints and
actuators. Shadow's own specification separates them, reading ``20 actuated DOF and a further 4
under-actuated movements for a total of 24 joints'' \cite{shadow_dexterous_hand_2005}; Inspire's
page gives a DoF figure and a joint count that differ by a factor of two
\cite{inspire_rh56dfx_2023}; a tendon hand can list more actuators than it has kinematic degrees
of freedom \cite{daxo_muscle_v0_2025}. Table~\ref{tab:hands-available} therefore prints joints and
actuated DoF in separate columns, with actuators kept in a third column not printed here, and
the gap between them is the informative number.

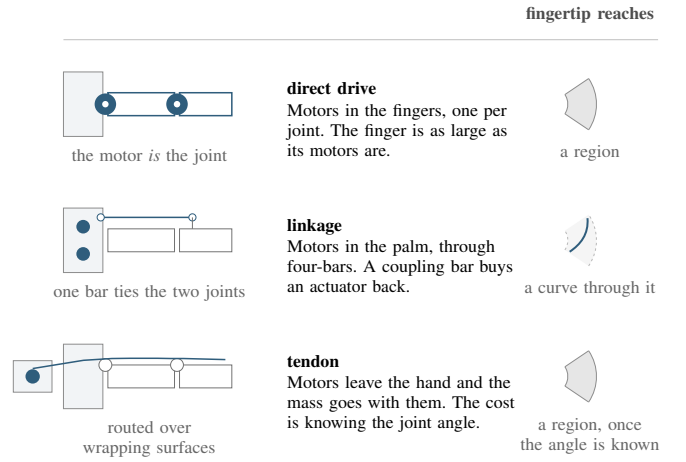
\begin{figure}[!t] \centering
\IfFileExists{figs/fig_transmission.tex}{\resizebox{\columnwidth}{!}{\begin{tikzpicture}[
  font=\footnotesize,
  palm/.style={draw=black!45,fill=wash,line width=0.3pt},
  link/.style={draw=black!55,fill=white,line width=0.35pt},
  linkacc/.style={draw=accent,fill=white,line width=0.6pt},
  bar/.style={draw=accent,line width=0.6pt},
  pin/.style={draw=accent,fill=white,line width=0.4pt},
  tend/.style={draw=accent,line width=0.6pt},
  wrap/.style={draw=black!55,fill=white,line width=0.35pt},
  hd/.style={font=\scriptsize\bfseries,anchor=north west},
  txt/.style={font=\scriptsize,anchor=north west,align=left,text width=32mm},
  capt/.style={font=\scriptsize\color{black!60},anchor=north,align=center,text width=30mm},
  glab/.style={font=\scriptsize\color{black!60},anchor=north,align=center,text width=21mm},
]

\begin{scope}[shift={(0,5.05)}]
  \node[palm,minimum width=5.5mm,minimum height=9mm,inner sep=0pt] at (0.275,0) {};
  \draw[linkacc] (0.62,-0.16) rectangle (1.55,0.16);
  \draw[linkacc] (1.62,-0.16) rectangle (2.35,0.16);
  \fill[accent] (0.585,0) circle (0.15);
  \fill[accent] (1.585,0) circle (0.15);
  \fill[white] (0.585,0) circle (0.045);
  \fill[white] (1.585,0) circle (0.045);
  \node[capt] at (1.20,-0.50) {the motor \emph{is} the joint};
\end{scope}

\begin{scope}[shift={(0,3.15)}]
  \node[palm,minimum width=5.5mm,minimum height=9mm,inner sep=0pt] at (0.275,0) {};
  \fill[accent] (0.275,0.19) circle (0.095);
  \fill[accent] (0.275,-0.19) circle (0.095);
  \draw[link] (0.62,-0.16) rectangle (1.55,0.16);
  \draw[link] (1.62,-0.16) rectangle (2.35,0.16);
  \draw[bar] (0.52,0.32) -- (1.80,0.32);
  \draw[pin] (0.52,0.32) circle (0.05);
  \draw[pin] (1.80,0.32) circle (0.05);
  \draw[black!55,line width=0.35pt] (1.80,0.32) -- (1.80,0.16);
  \node[capt] at (1.20,-0.50) {one bar ties the two joints};
\end{scope}

\begin{scope}[shift={(0,1.25)}]
  \draw[palm] (-0.70,-0.22) rectangle (-0.16,0.22);
  \fill[accent] (-0.43,0) circle (0.10);
  \node[palm,minimum width=5.5mm,minimum height=9mm,inner sep=0pt] at (0.275,0) {};
  \draw[link] (0.62,-0.16) rectangle (1.55,0.16);
  \draw[link] (1.62,-0.16) rectangle (2.35,0.16);
  \draw[wrap] (0.585,0.16) circle (0.09);
  \draw[wrap] (1.585,0.16) circle (0.09);
  \draw[tend] (-0.43,0.11) -- (0.28,0.225) -- (0.675,0.25)
              .. controls (1.05,0.255) and (1.24,0.255) .. (1.495,0.25)
              -- (1.675,0.25) -- (2.26,0.235);
  \node[capt] at (1.20,-0.50) {routed over wrapping surfaces};
\end{scope}

\node[hd]  at (3.00,5.48) {direct drive};
\node[txt] at (3.00,5.18) {Motors in the fingers, one per joint. The finger is as large as its
                           motors are.};

\node[hd]  at (3.00,3.58) {linkage};
\node[txt] at (3.00,3.28) {Motors in the palm, through four-bars. A coupling bar buys an actuator
                           back.};

\node[hd]  at (3.00,1.68) {tendon};
\node[txt] at (3.00,1.38) {Motors leave the hand and the mass goes with them. The cost is knowing
                           the joint angle.};

\node[font=\scriptsize\bfseries\color{black!70},anchor=south] at (7.35,6.06)
  {fingertip reaches};

\foreach \y in {5.05,1.25} {
  \begin{scope}[shift={(6.80,\y)}]
    \fill[black!10] (-34:0.24) arc (-34:34:0.24) -- (34:0.64) arc (34:-34:0.64) -- cycle;
    \draw[black!45,line width=0.3pt] (-34:0.24) arc (-34:34:0.24);
    \draw[black!45,line width=0.3pt] (-34:0.64) arc (-34:34:0.64);
    \draw[black!45,line width=0.3pt] (-34:0.24) -- (-34:0.64);
    \draw[black!45,line width=0.3pt] (34:0.24) -- (34:0.64);
  \end{scope}}
\node[glab] at (7.35,4.60) {a region};
\node[glab] at (7.35,0.80) {a region, once\\the angle is known};

\begin{scope}[shift={(6.80,3.15)}]
  \fill[black!4] (-34:0.24) arc (-34:34:0.24) -- (34:0.64) arc (34:-34:0.64) -- cycle;
  \draw[black!28,line width=0.25pt,dash pattern=on 1pt off 1.2pt] (-34:0.24) arc (-34:34:0.24);
  \draw[black!28,line width=0.25pt,dash pattern=on 1pt off 1.2pt] (-34:0.64) arc (-34:34:0.64);
  \draw[accent,line width=0.8pt] (-31:0.31) .. controls (-8:0.42) and (9:0.54) .. (31:0.60);
\end{scope}
\node[glab] at (7.35,2.70) {a curve through it};

\draw[draw=black!25,line width=0.28pt] (0,5.95) -- (8.30,5.95);
\draw[draw=black!25,line width=0.28pt] (0,-0.06) -- (8.30,-0.06);
\end{tikzpicture}}}{%
\framebox[0.95\columnwidth]{\rule{0pt}{5.2cm}\textit{figs/fig\_transmission.tex pending}}}
\caption{Three answers to where the motor goes, and what each one costs. A coupling bar saves
an actuator by tying two joints, so the fingertip reaches a curve through the reachable region
rather than the region itself.} \label{fig:transmission} \end{figure}

The actuation ratio is the real decision, and \figurename~\ref{fig:transmission} sets the three
answers side by side. The Pisa/IIT SoftHand drives nineteen joints from one motor, and its
authors state the cost plainly: ``No in-hand dexterous manipulation is required for this
prototype'' \cite{pisa_iit_softhand_2014}. Tendon drive moves the motors off the fingers and the
mass follows them, so a hand with every motor in its palm can weigh a quarter of a tendon hand
with a forearm \cite{shadow_dexterous_hand_2005,ilda_hand_2021}. What tendons cost is state
estimation. The Faive Hand estimated joint angles from tendon length through a Kalman filter and
its cube reorientation failed on hardware, which its authors blamed on ``poor joint angle
measurement from the EKFs, especially when there is contact'' \cite{faive_hand_2023}. Ruka-v2
adds detachable encoders to measure that problem without closing the loop on it, and they put
its open-loop joint-to-motor map at several degrees of error \cite{ruka_v2_2026}.

Direct drive trades torque density for transparency: LEAP's Dynamixel joints resist 19.5 N in a
pull-out test against the Allegro's 8.5 N \cite{leap_hand_2023}. Linkage drive is the argument
against the forearm, putting motors, drivers and ball screws inside the palm and still reaching
tens of newtons at the fingertip \cite{ilda_hand_2021}. Linkages also buy joint coupling for free,
as in BiDexHand's four-bar that drives each distal joint off the one behind it, though that hand's
paper and its own README disagree about how many joints and servos it has
\cite{bidexhand_2025}.

Fingertip force is the number a hand buyer reads first, and across these hands it is six
different measurements wearing one name. Pull-out resistance, a pinch between two fingers,
fingertip normal force under a standard indenter, a calibrated multi-trial measurement, a peak
vendor claim and two unlabelled columns on a spec sheet all appear as ``fingertip force''. One
vendor's figure turns out to be the newton equivalent of a payload at a fixed motor current, which
is a different quantity again \cite{orca_hand_2025}. Table~\ref{tab:hands-available} therefore
prints the protocol beside every force it reports and leaves the cell empty where the protocol
cannot be recovered. Payload is no more comparable: nine hands state one and no two state it under
the same test, so it is recorded with its protocol and this table does not print it at all.

Compliance is repairability seen twice over, and both are mechanical fuses: the Pisa/IIT
SoftHand's rolling-contact joints return to assembly after over-extension and ORCA's
``poppable'' pin joints dislocate under load \cite{pisa_iit_softhand_2014,orca_hand_2025}. ORCA
also reports tendon drive as maintenance: ``Prolonged use requires manual re-tensioning to
maintain performance''. What fails is rarely the motors. Its durability run found silicone skin
degrading on two fingertips after about 2,000 to 4,000 grasp cycles and sensor wires snapping on
three after about 4,500 to 7,000. The sensing wore out an order of magnitude sooner than the
hand.

Two properties matter before any of the above, and no vendor comparison carries them; both are
recorded here. The first is the rate at which a control loop can actually be
closed, which spans a factor of four across these hands and is not stated at all by some. One
hand's quoted rate turns out to be a serial query ceiling while its own transferred policy runs at
20 Hz, which answer different questions \cite{leap_hand_2023}. A reader deploying torque control
is bitten by that first. The second is whether a URDF or MJCF exists and who ships it, which is
the bridge to Section~\ref{sec:simulators} and to this survey's recommendation: prefer a hand a
simulator already carries. LEAP is the warning there, because its paper says the URDF is released
and the released API repository contains none \cite{leap_hand_2023}.

\subsection{Hands in use, and what they cost}
\label{sec:hands:used}

\begin{figure}[!t]
  \centering
  \resizebox{\columnwidth}{!}{\begin{tikzpicture}[
  font=\footnotesize,
  bar/.style={draw=none,fill=black!62},
  barlight/.style={draw=none,fill=black!22},
  baracc/.style={draw=none,fill=accent},
  box/.style={draw=black!45,line width=0.3pt,inner sep=3pt,align=left},
  ghost/.style={draw=black!35,line width=0.3pt,dash pattern=on 1.4pt off 1.4pt,inner sep=3pt,align=left},
  lnk/.style={draw=black!38,line width=0.28pt},
  lbl/.style={font=\scriptsize},
  num/.style={font=\scriptsize\color{black!55}},
]

\fill[bar] (0,0.00) rectangle (5.77,-0.34);
\fill[barlight] (5.77,0.00) rectangle (8.00,-0.34);
\node[font=\scriptsize\color{white},anchor=west] at (0.10,-0.17) {\textbf{75} of the 104 rows that name a hand name one of five};
\node[font=\scriptsize\color{black!60},anchor=east] at (7.90,-0.17) {29};
\node[font=\scriptsize\color{black!55},anchor=west] at (0,-0.64) {bars: method papers whose own experiments use the hand};
\draw[lnk] (0,-0.94) -- (8.00,-0.94);
\node[font=\scriptsize\bfseries,anchor=north west] at (0,-1.00) {sold};
\node[font=\scriptsize\color{black!55},anchor=north east] at (8.00,-1.00) {14 hands};
\node[lbl,anchor=east] at (2.32,-1.50) {Allegro};
\fill[bar] (2.42,-1.58) rectangle (7.30,-1.42);
\node[num,anchor=west] at (7.40,-1.50) {35};
\node[lbl,anchor=east] at (2.32,-1.77) {Shadow};
\fill[bar] (2.42,-1.85) rectangle (5.35,-1.68);
\node[num,anchor=west] at (5.45,-1.77) {21};
\node[lbl,anchor=east] at (2.32,-2.03) {Inspire RH56};
\fill[bar] (2.42,-2.12) rectangle (5.07,-1.95);
\node[num,anchor=west] at (5.17,-2.03) {19};
\node[lbl,anchor=east] at (2.32,-2.30) {XHand};
\fill[bar] (2.42,-2.38) rectangle (3.54,-2.21);
\node[num,anchor=west] at (3.64,-2.30) {8};
\node[lbl,anchor=east] at (2.32,-2.56) {Ability Hand};
\fill[bar] (2.42,-2.65) rectangle (3.40,-2.48);
\node[num,anchor=west] at (3.50,-2.56) {7};
\node[lbl,anchor=east] at (2.32,-2.83) {SharpaWave};
\fill[bar] (2.42,-2.91) rectangle (3.26,-2.74);
\node[num,anchor=west] at (3.36,-2.83) {6};
\node[lbl,anchor=east] at (2.32,-3.09) {Wuji};
\fill[bar] (2.42,-3.18) rectangle (2.70,-3.01);
\node[num,anchor=west] at (2.80,-3.09) {2};
\node[lbl,anchor=east] at (2.32,-3.36) {AgiBot OmniHand};
\fill[bar] (2.42,-3.44) rectangle (2.56,-3.27);
\node[num,anchor=west] at (2.66,-3.36) {1};
\node[lbl,anchor=east] at (2.32,-3.62) {BrainCo Revo 2};
\fill[bar] (2.42,-3.71) rectangle (2.56,-3.54);
\node[num,anchor=west] at (2.66,-3.62) {1};
\node[lbl,anchor=east] at (2.32,-3.89) {DEX-EE};
\fill[bar] (2.42,-3.97) rectangle (2.56,-3.80);
\node[num,anchor=west] at (2.66,-3.89) {1};
\node[lbl,anchor=east] at (2.32,-4.15) {Linker Hand};
\fill[bar] (2.42,-4.24) rectangle (2.56,-4.07);
\node[num,anchor=west] at (2.66,-4.15) {1};
\fill[wash] (0,-4.32) rectangle (8.00,-4.94);
\node[font=\scriptsize\color{black!60},anchor=north west,align=left] at (0.08,-4.36) {no method paper runs on: Proception ProHand, Tesollo DG-5F,\\Unitree Dex5};
\node[font=\normalsize\bfseries\color{accent},anchor=east] at (7.94,-4.63) {0};
\draw[lnk] (0,-5.04) -- (8.00,-5.04);
\node[font=\scriptsize\bfseries,anchor=north west] at (0,-5.10) {open hardware};
\node[font=\scriptsize\color{black!55},anchor=north east] at (8.00,-5.10) {6 hands};
\node[lbl,anchor=east] at (2.32,-5.60) {LEAP};
\fill[bar] (2.42,-5.69) rectangle (4.09,-5.52);
\node[num,anchor=west] at (4.19,-5.60) {12};
\fill[wash] (0,-5.77) rectangle (8.00,-6.39);
\node[font=\scriptsize\color{black!60},anchor=north west,align=left] at (0.08,-5.81) {no method paper runs on: BiDexHand, DexHand, ORCA, RUKA,\\Ruka-v2};
\node[font=\normalsize\bfseries\color{accent},anchor=east] at (7.94,-6.08) {0};
\draw[lnk] (0,-6.49) -- (8.00,-6.49);
\node[font=\scriptsize\bfseries,anchor=north west] at (0,-6.55) {research prototype, not purchasable};
\node[font=\scriptsize\color{black!55},anchor=north east] at (8.00,-6.55) {4 hands};
\node[lbl,anchor=east] at (2.32,-7.05) {LEAP Hand V2};
\fill[bar] (2.42,-7.14) rectangle (2.70,-6.97);
\node[num,anchor=west] at (2.80,-7.05) {2};
\node[lbl,anchor=east] at (2.32,-7.32) {Faive};
\fill[bar] (2.42,-7.40) rectangle (2.56,-7.23);
\node[num,anchor=west] at (2.66,-7.32) {1};
\fill[wash] (0,-7.49) rectangle (8.00,-7.88);
\node[font=\scriptsize\color{black!60},anchor=north west,align=left] at (0.08,-7.53) {no method paper runs on: ILDA, Pisa/IIT};
\node[font=\normalsize\bfseries\color{accent},anchor=east] at (7.94,-7.68) {0};
\draw[lnk] (0,-7.98) -- (8.00,-7.98);
\node[font=\scriptsize\bfseries,anchor=north west] at (0,-8.04) {announced by a company, not purchasable};
\node[font=\scriptsize\color{black!55},anchor=north east] at (8.00,-8.04) {9 hands};
\fill[wash] (0,-8.44) rectangle (8.00,-9.30);
\node[font=\scriptsize\color{black!60},anchor=north west,align=left] at (0.08,-8.48) {no method paper runs on: 1X NEO, Atlas hand, Clone, Daxo\\Muscle, Figure 03, PaXini DexH13, Sanctuary Phoenix, Tesla\\Optimus, Xiaomi CyberOne};
\node[font=\normalsize\bfseries\color{accent},anchor=east] at (7.94,-8.87) {0};
\draw[lnk] (0,-9.40) -- (8.00,-9.40);
\node[font=\scriptsize\bfseries,anchor=north west] at (0,-9.46) {platforms with no row in the hand tables};
\node[font=\scriptsize\color{black!55},anchor=north east] at (8.00,-9.46) {4 of them};
\node[lbl,anchor=east] at (2.32,-9.96) {parallel-jaw gripper};
\fill[bar] (2.42,-10.04) rectangle (4.09,-9.87);
\node[num,anchor=west] at (4.19,-9.96) {12};
\node[lbl,anchor=east] at (2.32,-10.22) {Adroit (Shadow in MuJoCo)};
\fill[bar] (2.42,-10.31) rectangle (2.98,-10.14);
\node[num,anchor=west] at (3.08,-10.22) {4};
\node[lbl,anchor=east] at (2.32,-10.49) {Fourier};
\fill[bar] (2.42,-10.57) rectangle (2.84,-10.40);
\node[num,anchor=west] at (2.94,-10.49) {3};
\node[lbl,anchor=east] at (2.32,-10.75) {D'Claw or TriFinger};
\fill[bar] (2.42,-10.84) rectangle (2.70,-10.67);
\node[num,anchor=west] at (2.80,-10.75) {2};
\draw[lnk] (0,-10.89) -- (8.00,-10.89);
\node[font=\scriptsize\color{black!55},anchor=north west] at (0,-10.97) {33 hands tabulated, 19 of them with no method row at all};
\end{tikzpicture}}
  \caption{Which hands the literature runs on, banded by how the hand can be got, with every tabulated hand shown and those no paper uses as named slots at zero. A paper naming two hands counts under both.}
  \label{fig:hands}
\end{figure}
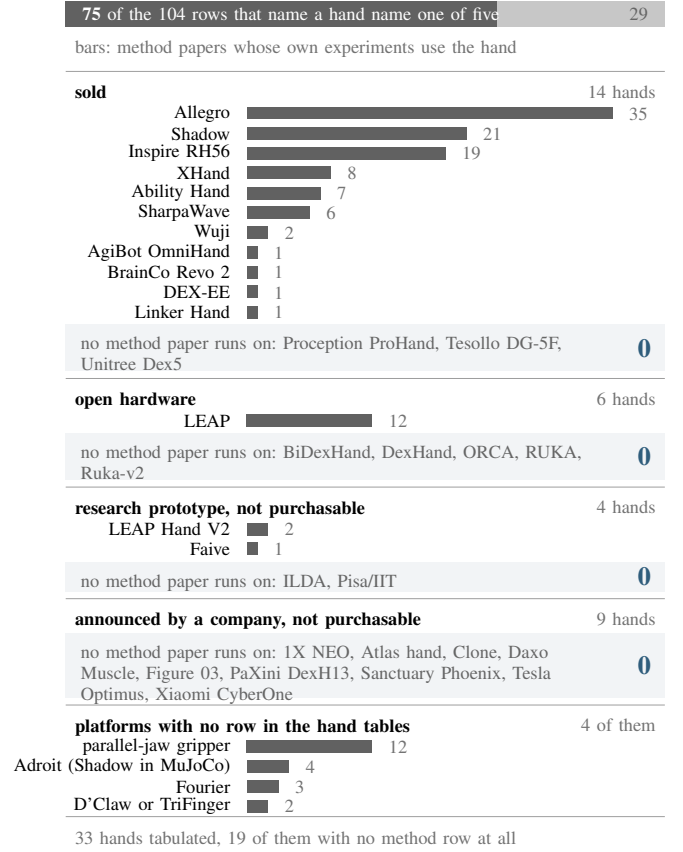

The two oldest designs in Table~\ref{tab:hands-available} carry 49 of the 104 papers that name a
hand at all, and seven papers use both. \figurename~\ref{fig:hands} counts, per hand, the papers
whose own experiments use it, and the shape of that figure is the finding: four hands take three
quarters of the corpus and everything else shares the remainder. What no paper states is how big
any of them is. Neither hand table carries a
dimension column, because no source behind them gives a hand's length or its width: the
specification a reader could hold against their own hand is the one these sources omit, while
the degree-of-freedom count, which all of them give, is compatible with a hand of any size at
all. Mass is the nearest thing on record and it is not the same quantity.

The Allegro's position is the uncomfortable part. Its product page at allegrohand.com/v4 returned
HTTP 404 while the Wonik wiki timed out \cite{allegro_hand_v4_2016}. Everything
Table~\ref{tab:hands-available} confirms about the most-used hand in the corpus comes from its ROS
driver: 16 joints, four fingers, a torque interface, a 333 Hz CAN clock, 12 V, no tactile sensing.
Its weight, joint torque, payload and price have no reachable source. The field's most common
platform cannot be specified from a page a reader can open.

Concentration would matter less if the hand did not move the result, and it does. On the same
simulated cube rotation LEAP turns the cube nearly three times faster than the Allegro
\cite{leap_hand_2023}, and on the same three-trial pinch test RUKA reports almost twice the
Allegro's force \cite{ruka_2025}. A method compared only on Allegro hardware is compared at one
point in a space where a single axis moves the headline number two or three times over. The three
newest vendor hands have arrived quickly and take a growing share, none of them before 2024, and
eight hands that can be bought or built take no papers at all.

\label{sec:hands:open}
Six hands in Table~\ref{tab:hands-available} are open hardware, and LEAP Hand V2 in
Table~\ref{tab:hands-announced} is a seventh. Five of the seven state a dollar cost: \$2,000 for
LEAP, \$3,000 for LEAP Hand V2, \$1,500 for Ruka-v2, \$1,300 for RUKA and \$300 for DexHand. ORCA
states a material cost below 2,000 CHF that the price column leaves unconverted, and BiDexHand
states none \cite{orca_hand_2025,bidexhand_2025}. The Faive Hand states neither a cost nor a
license that could be read, so it is not counted as open hardware here \cite{faive_hand_2023}. Two
of the stated bases need saying. DexHand's \$300 is ``additional total cost of components'',
excluding the printing and the wrist servos \cite{dexhand_open_source_2023}, and ORCA's own figure
sits against Ruka-v2's table listing ORCA at about \$3.5K \cite{ruka_v2_2026}.

The expensive end of that comparison is secondhand throughout: the six-figure prices for a Shadow
Hand come from other people's comparison tables \cite{ruka_2025,faive_hand_2023}, because the
vendor's own page asks buyers to discuss pricing. So the collapse is real at the cheap end and
inferred at the expensive one, and it has barely moved the literature. Twelve of the 104
hand-naming papers use an open-hardware hand, and all twelve of those are LEAP.

What the cheap hands give up is sensing. LEAP has none and names touch sensors as future work
\cite{leap_hand_2023}, RUKA states its design ``lacks tactile sensing'' \cite{ruka_2025}, and
BiDexHand's sources never mention touch \cite{bidexhand_2025}. ORCA is the exception, with binary
FSR fingertips whose threshold was measured as low as 0.05 N on a fresh fingertip, against the
sensor's rated 0.29 N, and 6.38 N on a degraded one \cite{orca_hand_2025}.

\subsection{Tactile sensing}
\label{sec:hands:tactile}

DIGIT set the cost floor, stating a manufacturing cost of about 15 USD per sensor in batches of a
thousand \cite{digit_2020}, which is where the price of touch sensing stopped being the obstacle.
Durability was as much of the contribution: its gel barely degraded over an abrasion test that
destroyed the two gels compared against it.

Digit 360 is the same lineage at a different operating point: millions of taxels, force
resolution around a millinewton, and on-device processing that cuts event-to-action latency by most
of an order of magnitude \cite{digit360_2024}. It is also not for sale, and no paper in this corpus
names it.

XELA's uSkin is the non-camera alternative, and the one that bolts onto hands the field already
uses: a curved fingertip kit for the Allegro and LEAP, with three-axis sensing points at
gram-force resolution running at a few hundred hertz \cite{xela_uskin_2020}.
Table~\ref{tab:hands-announced} gives the specification, including which figures belong to the
product family and not to the fingertip kit. One paper in this corpus uses it, and its
self-supervised encoders beat matched end-to-end models when both see a fraction of the labels,
though its own bead-maze policies never complete the maze on the real robot \cite{sparsh_2024}.

The usage number is the one to keep, and it needs its inclusion rule stated: a sensor physically
on the hand, read by the deployed policy. Eight papers meet it:
\cite{anyrotate_2024,articulated_tools_inhand_2025,dexteleop0_2026,dexumi_2025,hato_visuotactile_2024,robot_synesthesia_2023,rotateit_2023}
and \cite{rotating_without_seeing_2023}. The rule excludes \cite{penspin_2024}, whose 20 binary
contacts are simulated on an Allegro without tactile hardware and whose released configuration
disables tactile input; it also excludes \cite{dexndm_2025} and \cite{dexplore_2025} for the
same reason. Sixty-three of the 113 method papers use the word tactile somewhere in their parsed
text and 35 carry it in their structured note. Against the eight that meet the rule above,
either number is the gap worth quoting.

Almost none of the eight uses a high-resolution sensor. Yin et al.
\cite{rotating_without_seeing_2023} remove vision entirely and rotate objects from 16 binary
touch sensors over the palm, links and fingertips, deployed zero-shot to a real Allegro, and
Robot Synesthesia uses 16 force-sensing resistors read as binary \cite{robot_synesthesia_2023}.
Two papers exclude touch, HORA reporting rotation ``even without the usage of vision and tactile
sensing'' \cite{hora_2022}, and ORCA's authors dropping it from their reinforcement learning
``due to the additional complexity involved in accurately modeling them'' \cite{orca_hand_2025}.
Taxel counts have risen by three orders of magnitude while the policies consuming them have
stayed at binary contact.

\subsection{The catalog against the literature}
\label{sec:hands:announced}

Behind Table~\ref{tab:hands-announced}'s entries sit three
kinds of evidence, and conflating them is how a DoF figure with no source ends up in a survey.
Vendor prose carrying numbers is the strongest, as on 1X's page of 9 July 2026
\cite{onex_neo_hand_2026}. Video is second and carries none. Press or bibliography assertion is
third, as with Tesla's V3 hand, known here only through a paraphrase of patents because the
USPTO PDF parsed empty \cite{tesla_optimus_hand_2025}. Table~\ref{tab:hands-announced} blanks
the cells resting on the third class and footnotes who did the arithmetic.

A survey can go one step past recording that a claim is unverified, which is to say which claims
are implausible on their face. One vendor's actuator count and mass work out to about six grams
per actuator including structure, tendons, routing and skin \cite{daxo_muscle_v0_2025}.
Another's mass figure appears to exclude a half-kilowatt pump \cite{clone_robotics_hand_2024}. A
third's degree-of-freedom figure sits on the same page as a finger count exactly half of it, so
it is almost certainly the pair of hands \cite{figure_03_hand_2025}. A fourth, widely repeated
for the Optimus hand, is one article's arithmetic and not a vendor statement, and the previous
generation's own figure was half of it \cite{tesla_optimus_hand_2025}.

Skepticism belongs to the evidence class, not to which table an entry lands in. The
fully-actuated claims from the newest vendors are claims about unmeasured hardware, and they sit in
Table~\ref{tab:hands-available} beside the footnotes their own pages carry: ``Specifications may
vary between products'' on one, a spec that ``will continue to iterate'' on a beta product on
another \cite{sharpa_wave_2026,wuji_hand_2025}. One of the two has since published a teleoperated
dataset collected on the hand itself (\cite{sharpadex_2026}, Section~\ref{sec:bimanual}), which is
not a measurement of the specification either, but it is the first public evidence that the
hardware runs. Even the best-specified vendor page in the corpus
leaves its fingertip-force columns unlabelled.

\label{sec:hands:gap}
The band at zero in \figurename~\ref{fig:hands} carries the finding. None of the nine
company-announced hands in Table~\ref{tab:hands-announced} appears in a single paper whose
own experiments use it. Tesla, Figure, 1X, Sanctuary, Boston Dynamics, Xiaomi, Clone, Daxo and
PaXini account for zero of the 113 method papers' experiments. The nearest thing to a
counterexample is \cite{helix_2025}, a Figure blog post claiming a 35-DoF whole-upper-body action
space at 200 Hz that includes individual finger control. It never names the hand, gives no
per-hand DoF count, and reports no success rate or trial count for any task. It is the maker
describing its own unreleased hand, which is the evidence class the finding is about. The four
research prototypes in Table~\ref{tab:hands-announced} are in a different position, since the
Faive Hand and LEAP Hand v2 Advanced account for three papers between them
\cite{graspxl_2024,bidex_teleop_2024}.

Table~\ref{tab:hands-announced}'s emptiness is measurable and part of the same finding. Over the
specification columns it prints, from DoF to price, 58 percent of its cells are values no source
stated, against 42 percent for the hands that can be bought, and both captions state their own
count. The hands with the highest advertised DoF counts have the least specification behind
them.

The one hand that crosses the gap crosses it because its maker published a paper.
ByteDexter reaches a paper through a ByteDance technical report stating 21 DoF per hand, 16
piezoresistive fingertip channels in the action vector and real-robot success rates
\cite{gr_dexter_2025}, not through a product page. That mechanism is available to every vendor in
Table~\ref{tab:hands-announced} and none has used it.

The gap runs the other way too, and the last column of Table~\ref{tab:hands-available} shows it.
Eight hands that can be bought today or built from published designs take no papers at all: the
Unitree Dex5 \cite{unitree_dex5_2025}, the Tesollo DG-5F \cite{tesollo_dg5f_2024}, the Proception
ProHand \cite{proception_prohand_2026}, ORCA, RUKA, Ruka-v2, BiDexHand and DexHand. Cheapness is
established for only three of those eight, because the price cell is empty for the rest. A reader
choosing a hand on this corpus's evidence has four options with published precedent behind them,
and only one of the four, LEAP, has both a published design and a stated price. Below those four
the record thins fast, and 19 of the hands in Tables~\ref{tab:hands-available}
and~\ref{tab:hands-announced} sit there on their specifications alone, with no paper in this
corpus running on any of them.

Nineteen of the 33 hands in Table~\ref{tab:hands-available} and Table~\ref{tab:hands-announced}
appear in no paper, and 11 of those are neither sold nor open and appear in none for that
reason, so 33 is not the denominator for a software-lag claim: the eight hands of the paragraph
above are. The rule that decides used from unused is one pattern per hand matched against the
hand each paper names, so the partition is reproducible. The
lag runs through the tooling as well as the papers. Seven of the fifteen simulators ship a
first-party hand model at all, six of those hands are real, Brax's being a synthetic claw, and
every one of the six is an Allegro or a Shadow; only four of the 33 tabulated hands appear in
any engine. Bench2Dex \cite{bench2dex_2026} meanwhile compares 12 hands and DexVerse
\cite{dexverse_2026} six without stating a DoF count for any. A useful next step is a
conformance suite for hand models: one URDF or MJCF per hand, with fixed joint-limit, mass and
collision checks and a published pass or fail per engine.
\section{Simulators and the physics underneath}
\label{sec:simulators}

\subsection{Requirements of a dexterous simulation}
\label{sec:sim:requirements}

The clearest demonstration that hands are the hard case came from a benchmark that was not about
hands. Erez et al. built a 35-DOF arm modeled on the Shadow Hand, closed it around a capsule
with fixed spring-dampers, and asked four of the five engines for the largest timestep at which
the object was still in the hand. The answers spanned a factor of 500, from milliseconds down to
tens of microseconds, with MuJoCo at the slow-step end
\cite[Sec.~IV-D]{physics_engine_comparison_2015}. Three caveats travel with that spread. The
engines ran a restricted common model, hinge joints with sphere and capsule geometry and no
meshes; the authors call it ``an open question whether this model system can be tuned to work
better in the gaming engines''; and they wrote MuJoCo and say so.

A grasp is hard for nameable reasons. It is many contacts at once, all persistent, all near
stiction, on an object much lighter than the mechanism holding it. Persistence and multiplicity
make the problem hyperstatic, and Le Lidec et al. show that per-contact solvers of the projected
Gauss-Seidel family, and RaiSim's, then inject spurious jamming forces at stiction that vanish
only once the object slides. The mass ratio makes it ill-conditioned, and in their stacked-cube
test at a $10^{3}$ to $10^{-3}$ kg ratio those same methods fail to converge. Global methods with
proximal regularisation stay robust in both cases
\cite[Sec.~IV-A]{contact_models_comparison_2023}. What degrades a truncated solve is therefore
conditioning and redundancy, not stiffness, and a grasp supplies both by construction. Locomotion
is forgiving by comparison. On flat ground the contact model and solver choice ``hardly affects''
their quadruped's tracked base velocity, and only on rough, slippery terrain do RaiSim and CCP
deviate. A walking robot makes and breaks a few contacts against ground far heavier than itself. A
hand does neither.

Speed enters by the same door. MuJoCo Playground reports that contact time scales with the
number of possible contacts and not the active ones, because JAX requires static shapes, which
is why its tasks override the bound on contact points and the bound on geometry pairs by hand
\cite[Sec.~VI]{mujoco_playground_2025}. Every override printed in that paper is a locomotion
port. A hand's bound would have to be set the same way, for the worst case, but no hand
configuration is printed.

\begin{figure}[!t]
  \centering
  \IfFileExists{figs/fig_collision.tex}{\resizebox{\columnwidth}{!}{\begin{tikzpicture}[
  font=\footnotesize,
  vis/.style={draw=black!60,line width=0.5pt,fill=white},
  col/.style={draw=black!60,line width=0.5pt,fill=black!8},
  ghostmat/.style={draw=none,fill=accent,fill opacity=0.16},
  cut/.style={draw=black!45,line width=0.3pt,dash pattern=on 1.4pt off 1.2pt},
  tip/.style={draw=accent,line width=0.7pt,fill=white},
  tipghost/.style={draw=accent,line width=0.4pt,dash pattern=on 1.2pt off 1.2pt,fill=none},
  hd/.style={font=\footnotesize\bfseries,anchor=north},
  sub/.style={font=\scriptsize\color{black!60},anchor=north,align=center,text width=26mm},
  gap/.style={draw=accent,line width=0.5pt,<->},
  glab/.style={font=\scriptsize\color{accent},anchor=west},
]

\begin{scope}[shift={(0.30,0)}]
  \draw[vis] (0,0) -- (2.0,0) -- (2.0,1.5) -- (1.4,1.5) -- (1.4,0.75)
             -- (0.6,0.75) -- (0.6,1.5) -- (0,1.5) -- cycle;
  \draw[tip] (1.0,1.04) circle (0.28);
  \node[hd] at (1.0,-0.16) {visual mesh};
  \node[sub] at (1.0,-0.50) {the concavity is there, and the fingertip sits in it};
\end{scope}

\begin{scope}[shift={(3.05,0)}]
  \draw[col] (0,0) rectangle (2.0,1.5);
  \fill[ghostmat] (0.6,0.75) rectangle (1.4,1.5);
  \draw[cut] (0.6,1.5) -- (0.6,0.75) -- (1.4,0.75) -- (1.4,1.5);
  \draw[tip] (1.0,1.79) circle (0.28);
  \draw[tipghost] (1.0,1.04) circle (0.28);
  \draw[gap] (1.62,1.51) -- (1.62,1.78);
  \node[glab] at (1.70,1.65) {held out};
  \node[hd] at (1.0,-0.16) {its convex hull};
  \node[sub] at (1.0,-0.50) {the notch is filled with material only the solver has};
\end{scope}

\begin{scope}[shift={(5.80,0)}]
  \draw[col] (0,0) rectangle (0.6,1.5);
  \draw[col] (0,0) rectangle (2.0,0.75);
  \draw[col] (1.4,0) rectangle (2.0,1.5);
  \draw[cut] (0.6,0) -- (0.6,0.75);
  \draw[cut] (1.4,0) -- (1.4,0.75);
  \draw[tip] (1.0,1.04) circle (0.28);
  \node[hd] at (1.0,-0.16) {convex decomposition};
  \node[sub] at (1.0,-0.50) {three convex parts, and the contact is back};
\end{scope}

\draw[draw=black!25,line width=0.28pt] (0.30,2.28) -- (7.80,2.28);
\end{tikzpicture}}}{%
    \framebox[0.95\columnwidth]{\rule{0pt}{2.6cm}\textit{figs/fig\_collision.tex pending}}}
  \caption{The geometry the solver uses against the geometry the renderer draws. A single convex
  hull fills the concavity with material that exists only in the collision model, and holds the
  fingertip out of a contact the visual mesh allows.}
  \label{fig:collision}
\end{figure}
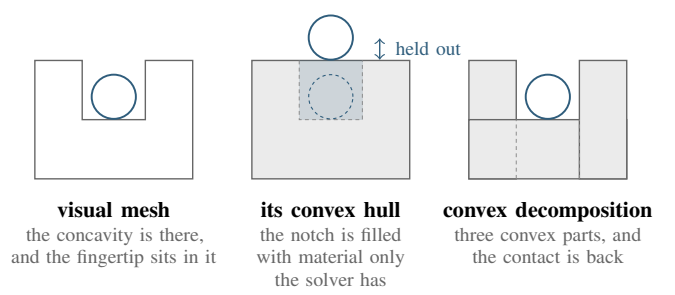

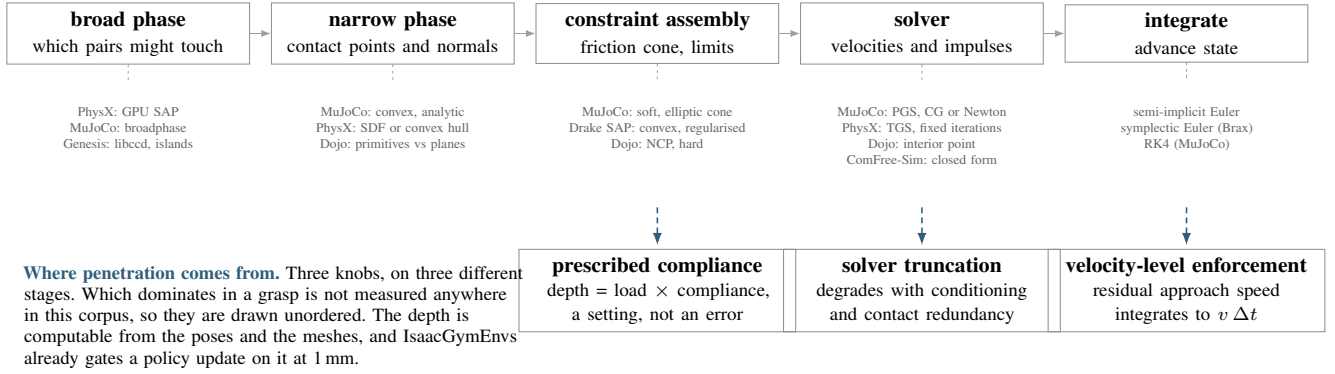
\begin{figure*}[!t]
  \centering
  \IfFileExists{figs/fig_simstep.tex}{\begin{tikzpicture}[
  font=\footnotesize,
  stage/.style={draw=black!45,line width=0.3pt,inner sep=3pt,align=center,text width=3.0cm},
  pnl/.style={draw=none,inner sep=0pt,align=left},
  lnk/.style={draw=black!38,line width=0.28pt},
  drop/.style={draw=black!38,line width=0.28pt,dash pattern=on 1pt off 1.4pt},
  src/.style={draw=accent,line width=0.6pt,dash pattern=on 2.2pt off 1.6pt,-latex},
  lbl/.style={font=\scriptsize},
  det/.style={font=\tiny\color{black!60},align=center,text width=3.45cm},
  note/.style={font=\scriptsize,align=left},
]

\node[stage] (s1) at (1.8,0)  {\textbf{broad phase}\\[1pt]{\scriptsize which pairs might touch}};
\node[stage] (s2) at (5.3,0)  {\textbf{narrow phase}\\[1pt]{\scriptsize contact points and normals}};
\node[stage] (s3) at (8.8,0)  {\textbf{constraint assembly}\\[1pt]{\scriptsize friction cone, limits}};
\node[stage] (s4) at (12.3,0) {\textbf{solver}\\[1pt]{\scriptsize velocities and impulses}};
\node[stage] (s5) at (15.8,0) {\textbf{integrate}\\[1pt]{\scriptsize advance state}};
\draw[lnk,-latex] (s1.east) -- (s2.west);
\draw[lnk,-latex] (s2.east) -- (s3.west);
\draw[lnk,-latex] (s3.east) -- (s4.west);
\draw[lnk,-latex] (s4.east) -- (s5.west);

\foreach \s in {s1,s2,s3,s4,s5} { \draw[drop] (\s.south) -- ++(0,-0.22); }
\node[det,anchor=north] at (1.8,-0.85)  {PhysX: GPU SAP\\ MuJoCo: broadphase\\ Genesis: libccd, islands};
\node[det,anchor=north] at (5.3,-0.85)  {MuJoCo: convex, analytic\\ PhysX: SDF or convex hull\\ Dojo: primitives vs planes};
\node[det,anchor=north] at (8.8,-0.85)  {MuJoCo: soft, elliptic cone\\ Drake SAP: convex, regularised\\ Dojo: NCP, hard};
\node[det,anchor=north] at (12.3,-0.85) {MuJoCo: PGS, CG or Newton\\ PhysX: TGS, fixed iterations\\ Dojo: interior point\\ ComFree-Sim: closed form};
\node[det,anchor=north] at (15.8,-0.85) {semi-implicit Euler\\ symplectic Euler (Brax)\\ RK4 (MuJoCo)};

\draw[src] (8.8,-2.30) -- (8.8,-2.80);
\draw[src] (12.3,-2.30) -- (12.3,-2.80);
\draw[src] (15.8,-2.30) -- (15.8,-2.80);

\node[stage,text width=3.45cm,anchor=north] at (8.8,-2.85)
  {\textbf{prescribed compliance}\\[1pt]{\scriptsize depth = load $\times$ compliance,\\ a setting, not an error}};
\node[stage,text width=3.45cm,anchor=north] at (12.3,-2.85)
  {\textbf{solver truncation}\\[1pt]{\scriptsize degrades with conditioning\\ and contact redundancy}};
\node[stage,text width=3.45cm,anchor=north] at (15.8,-2.85)
  {\textbf{velocity-level enforcement}\\[1pt]{\scriptsize residual approach speed\\ integrates to $v\,\Delta t$}};

\node[note,anchor=north west,text width=6.6cm] at (0.3,-2.95)
  {\scriptsize\textcolor{accent}{\textbf{Where penetration comes from.}} Three knobs, on three
   different stages. Which dominates in a grasp is not measured anywhere in this corpus, so they
   are drawn unordered. The depth is computable from the poses and the meshes, and IsaacGymEnvs
   already gates a policy update on it at 1\,mm.};
\end{tikzpicture}}{%
    \framebox[0.7\textwidth]{\rule{0pt}{2.2cm}\textit{figs/fig\_simstep.tex pending}}}
  \caption{One simulation step, and where engines differ: the stages every rigid-body engine runs,
  with the choices that separate the engines used for hands under each; dashed arrows mark the
  three stages that overlap.}
  \label{fig:simstep}
\end{figure*}

\figurename~\ref{fig:simstep} sets out the stages of one simulation step. Three of them make
overlap, and they do not answer to the same knob. In an engine that enforces non-penetration at
the velocity level, integration turns any residual approach velocity into overlap of order $v$
times $\Delta t$; an engine that enforces the gap at the next configuration carries no such
term, which is why Dojo's hard-contact NCP keeps its feet above the floor at every timestep it
was tested at. Constraint assembly fixes the compliance a loaded contact then rests at. A
truncated solver leaves a residual that grows with conditioning. Which of the three dominates
in a grasp is not measured anywhere in this corpus, and the three are not ordered here. A
fourth thing is not on the figure, because it is not a stage of the step and not a source of
overlap: a mismatch between the geometry the solver uses and the geometry the renderer draws,
which runs in both directions and is drawn in \figurename~\ref{fig:collision}.

\subsection{Contact models and solvers}
\label{sec:sim:engines}

Table~\ref{tab:simulators}, set in Appendix~\ref{app:tables}, is the engine-by-engine comparison,
one line per simulator, built only from what a note confirmed. Le Lidec et al. supply the taxonomy
that organises it, checking each formulation against the Signorini condition, Coulomb's law, and
the maximum dissipation principle. Linear complementarity, the family of Bullet, ODE, PhysX and
DART, satisfies Signorini alone, because linearising the friction cone to a pyramid biases
friction toward its corners. The cone complementarity problem satisfies the other two but relaxes
Signorini, so contact acts at a distance of size $\Delta t \cdot \mu \cdot \lVert c_T \rVert$. The
full nonlinear problem satisfies all three and is non-convex, by Table~II of
\cite{contact_models_comparison_2023}. A contact model is not the algorithm that solves
it. PhysX linearises the cone, which places its model in the LCP family, and it does not solve an
LCP. Its Temporal Gauss-Seidel scheme folds substepping into the Gauss-Seidel sweep and exposes
position and velocity iteration counts separately \cite[Sec.~3]{isaacgym_2021}.

MuJoCo is inside that taxonomy. Le Lidec et al. classify it as CCP-MuJoCo, a convex relaxation
solved by a Newton method on the primal QCQP, in the same family as CCP-Drake, the SAP-style
scheme, by Table~III of \cite{contact_models_comparison_2023}. The relaxation makes two errors
of opposite sign in one engine. A loaded contact carries a violation, and a sliding contact
carries force at a positive gap. Drake's SAP inherits the same pair, and its gliding effect at
distance $\varphi \approx \delta t \cdot \mu \cdot \lVert v_t \rVert$ ``unfortunately does not
go away as $\delta t \to 0$'' \cite[Sec.~V-A]{castro_sap_contact_2021}. Le Lidec et al. call
that compliance a ``numerical trick designed to circumvent the issues due to hyper-staticity or
ill-conditioning at the cost of impairing the simulation''.

The depth a loaded contact carries is a chosen number, not a property of the engine. In a
regularised formulation the steady-state violation at a contact is the normal load times the
compliance: zero at zero load, growing with what the contact carries. MuJoCo drives that
violation with a critically damped stabiliser whose steady-state depth, for an object resting
under gravity, has a closed form independent of the object's mass
\cite[Sec.~V]{mujoco_convex_contact_2014}. Whoever sets the stiffness sets the depth.

Published penetration figures do not compare with each other, and the reason matters before
anyone quotes one. Dojo solves a hard-contact problem and stays above the floor on an Atlas drop
test where MuJoCo overlaps by tens of millimetres \cite[Sec.~V-A]{dojo_2022}; Drake's figures
for the same quantity are orders of magnitude smaller, and they are analytical bounds for a
point mass at rest rather than measured depths \cite[Sec.~V-B]{castro_sap_contact_2021};
ComFree-Sim reports a depth directly, on small convex primitives dropped under their own weight
\cite[Sec.~IV-A]{comfree_sim_2026}. A point-mass bound, a humanoid drop transient and a dropped
primitive differ in load, inertia and regime, so the distances between those figures measure
nothing about each other. What the set does establish is the part that matters for a hand: in a
compliant formulation the depth follows from a stiffness somebody chose, and the GPU era moved
that knob without removing it. The oldest engine comparison here puts the consequence plainly,
reporting that engines pushed past their stable step ``effectively simulate a different physics
model which can no longer hold the object'' \cite[Sec.~IV-D]{physics_engine_comparison_2015}. No
paper in this corpus reports the depth for a fingertip loaded by a grasp, which is the one
configuration a dexterous result depends on.

Table~\ref{tab:simulators}'s most interesting column is the one that records whether an engine's
own paper reports a penetration depth, and it is nearly empty. Two engines report one, and both
are engines whose paper is about contact accuracy. The word ``penetration'' appears nowhere in
the Isaac Gym paper, and Isaac Lab's contact sensor reports force, duration and an average
contact point with no measure of contact quality at all. That is a documentation gap rather than
a capability gap: NVIDIA's own benchmark repository already computes an interpenetration depth
per environment for its assembly tasks and scales the reward down as the depth approaches a
millimetre. The quantity is a few lines of code away in the engine most of this literature
trains in.

The rest of Table~\ref{tab:simulators} is largely empty, and the emptiness is a result: 38
percent of the cells in its property columns hold a value no source stated, which is what its
caption counts. No engine paper states a default physics timestep, and the three that report one
report it for a named experiment, so the timestep column carries experiment settings and no
engine's default. Fewer than half the engines ship any dexterous hand. Two columns not printed
here are worth the trip: the solver iteration count, which four engines
state, and the license, which is the question a reader choosing an engine actually has and which
almost none of them answers.

\subsection{GPU-parallel engines and their throughput}
\label{sec:sim:gpu}

Isaac Gym set the pattern. Physics, observations, rewards and
actions stay on the GPU, and PhysX resolves contacts with the Temporal Gauss-Seidel sweep
described above. Its per-task timesteps are published, which is rare: the Shadow Hand runs a
1/120 s physics step under a 1/60 s control step, or 1/20 s in the OpenAI variant. The result
that reorganised the field is that reproducing OpenAI's Shadow Hand cube reorientation took
under an hour on one A100, against 30 hours on 6144 CPU cores and 8 V100s
\cite[Sec.~6.4.1]{isaacgym_2021}. Thirty-five of the papers examined run on it.

Orbit and Isaac Lab moved the stack to PhysX 5, and the dexterous offering is thinner than the
predecessor's: the first-party suite is lifting, grasping and reorienting with the KUKA Allegro
hand, while Shadow and Ability hands appear only through third-party grasp evaluation
\cite[Sec.~7.2.3]{isaaclab_2025}. MJX and MuJoCo Playground took the other route, putting MuJoCo's
own solver on the GPU through JAX at the cost described above. MuJoCo Warp ports the same solref
and solimp soft-constraint entries to NVIDIA Warp, and its README states that the legacy PGS solver
and the noslip pass are unsupported and that differentiability ``is not yet available''
\cite{mujoco_warp_2025}. GPU MuJoCo is not a differentiable MuJoCo. Genesis is the broadest,
pairing a rigid constraint solver with FEM, MPM, PBD and SPH solvers and a differentiable backward
path, and it ships a Shadow Hand. Its README quotes no throughput number at all, so the large FPS
figures that circulate for Genesis trace to nothing in \cite{genesis_2024}.

Newton changes how these names should be read. It is explicitly multi-solver: SolverMuJoCo wraps
MuJoCo Warp, SolverKamino is a proximal-ADMM and dual-variational-inequality contact solver, and
XPBD and VBD do position-based constraint projection \cite{newton_2025}. Its contact model is
solver-dependent, not a property of the engine, which is how Table~\ref{tab:simulators} records
it. Isaac Lab already offers a Newton--MuJoCo Warp backend in its hands demo, and its roadmap
announces Newton integration. An experiment reported as Isaac Lab may
be running PhysX 5 with TGS, or MuJoCo's soft-constraint entries under Warp, and those two make
different contact errors. The name also fails to fix the physics inside one engine. MuJoCo's
convex solver is a family, interior point or projected Newton, conjugate gradient or Gauss-Seidel
\cite[Sec.~II-D]{mujoco_2012}, and MuJoCo Warp supports neither PGS nor the noslip pass. PhysX 4
and PhysX 5 differ in whether non-convex rigid bodies get SDF collision, which is a
contact-geometry difference between Isaac Gym and Isaac Lab under one vendor name
\cite{orbit_2023,isaaclab_2025}. Naming a simulator no longer names its physics. Papers should
report the backend and the solver beside the framework.

\label{sec:sim:throughput}
Four headline figures measure four different quantities. Isaac Gym reports parallel environment
steps per second with physics, observations and rewards on device: 150,000 for the Shadow Hand at
16,384 environments on one A100. Isaac Lab reports frames per second in training, which includes
the learning update: over 900,000 for the DextrAH teacher task at 16,384 environments on eight RTX
Pro 6000 GPUs, with no single-GPU dexterous number anywhere in the paper. ManiSkill3 reports ``up
to 30,000+ FPS'' with RGBD and segmentation on one RTX 4090, measured over 1000 random actions
with reward and termination removed from the timing, and does not state the environment count
behind the figure \cite{maniskill3_2024}. Orbit reports a 125,000 FPS physics-only ceiling on an
RTX 3090, also with no environment count \cite[Sec.~VII]{orbit_2023}.

The one number reported cleanly enough to reuse is MuJoCo Playground's, in training steps per
second on one GPU over five seeds. A hand reorientation task runs about six times slower there than
a parallel-jaw pushing task on the same hardware in the same codebase. That factor is not a
property of the task alone: Playground tunes solver iterations, line-search iterations, timestep
and contact bounds per environment, in its own Table~III \cite{mujoco_playground_2025}, and does
not print the
two configurations side by side, which is what makes even those two figures incomparable. A
training-loop rate is also mostly not a measurement of physics; on Playground's own breakdown the
policy update dominates the frame and physics is a few percent of it.

Cross-framework comparisons add further free parameters. One widely cited table takes the best
result over a range of environment counts for each system \cite[Table~1a]{maniskill2_2023},
which tunes a free parameter per competitor, and the two systems being compared are not the same
kind of thing: one runs rigid-body physics on CPU workers behind a shared render server and the
other on the GPU. Because the measured loop includes rendering at a fixed control rate, what is
compared is a visual sample-collection loop and not two physics engines. Its authors say so
themselves, and Playground is equally explicit that its own cross-simulator plot borrows two of
its numbers from another paper. Several sources give no figure at all, pointing instead at an
external dashboard or omitting throughput from their documentation entirely.

The failure reaches past the engine papers into the methods. Of the 47 corpus method papers that
name a GPU-batched simulator, 19 state no environment count anywhere, and Isaac Lab's own paper
states no physics timestep, no decimation and no solver iteration count for any task. A throughput
figure is interpretable only with the environment count, the GPU, the timesteps, the solver
iteration budget, and a statement of whether rendering and the learning update sit inside the
measurement. Almost nobody reports all five.

\subsection{Tactile simulation and the sim-to-real gap}
\label{sec:sim:tactile}

The three tactile simulators in this corpus calibrate against three different things, and none of
them is a manipulation outcome. TACTO is a rendering layer over a host engine, by default
PyBullet's rigid contact model. It reads post-solve link poses and the engine's reported normal
force and maps that force to gel-mesh deformation at the rendering level, so it contributes no
contact physics of its own. Its only sim-to-real number is a tactile pose-estimation task, at
$1.66 \pm 0.16$ mm with color-jitter augmentation against $0.76 \pm 0.07$ mm for a model trained
on 128 real datapoints, by Table~II of \cite{tacto_2020}.

Taxim is example-based rather than simulated, calibrated from real indentations, and it beats
TACTO on every optical-similarity metric against real images. Its marker-motion model is calibrated
against a finite-element reference, and it tracks that reference about three times more tightly
than the reference tracks reality, so its residual is bounded below by its own calibration target.
The authors also state that they simulate quasi-static contact only, with slip left as future work
\cite[Sec.~V]{taxim_2021}. Slip is what a hand needs.

\begin{figure}[!t] \centering
\IfFileExists{figs/fig_tactiledepth.tex}{\resizebox{\columnwidth}{!}{\begin{tikzpicture}[
  font=\footnotesize,
  gel/.style={draw=black!55,line width=0.4pt,fill=wash},
  rigid/.style={draw=black!65,line width=0.5pt,fill=black!14},
  rest/.style={draw=black!45,line width=0.35pt,dash pattern=on 1.6pt off 1.4pt},
  over/.style={draw=none,fill=accent,fill opacity=0.22},
  dim/.style={draw=accent,line width=0.5pt,<->},
  outbox/.style={draw=black!45,line width=0.3pt,inner sep=3pt,align=center,
              font=\scriptsize,text width=24mm},
  arr/.style={draw=accent,line width=0.5pt,-latex},
  hd/.style={font=\scriptsize\bfseries,anchor=west},
  note/.style={font=\scriptsize\color{black!60},anchor=west,align=left},
  ax/.style={draw=black!55,line width=0.4pt},
  tk/.style={font=\scriptsize\color{black!60},anchor=north},
]

\begin{scope}[shift={(0,0)}]
  \fill[wash] (0,0) rectangle (3.30,0.80);
  \draw[gel] (0,0) rectangle (3.30,0.80);
  \draw[rest] (0,0.80) -- (3.30,0.80);
  \draw[rigid] (1.05,0.36) -- (2.25,0.36) -- (2.25,1.36) -- (1.05,1.36) -- cycle;
  \fill[over] (1.05,0.36) rectangle (2.25,0.80);
  \draw[accent,line width=0.5pt] (1.05,0.80) -- (1.05,0.36) -- (2.25,0.36) -- (2.25,0.80);
  \draw[gel] (0,0.80) -- (1.05,0.80);  \draw[gel] (2.25,0.80) -- (3.30,0.80);
  \node[font=\scriptsize\color{black!70}] at (1.65,1.00) {object};
  \node[font=\scriptsize\color{black!70},anchor=west] at (0.10,0.34) {sensor gel};
  \draw[dim] (2.52,0.36) -- (2.52,0.80);
  \node[font=\scriptsize\color{accent},anchor=west] at (2.60,0.58) {$d$};
  \node[note,anchor=north west,text width=76mm] at (0,-0.28)
    {$d$ is the depth by which the two solids overlap: the quantity every other engine in
     Sec.~IV works to remove};
\end{scope}

\node[outbox] (o1) at (5.35,1.14) {tactile image};
\node[outbox] (o2) at (5.35,0.30) {distributed force};
\draw[arr] (2.95,0.74) .. controls (3.70,0.98) and (3.85,1.14) .. (4.24,1.14);
\draw[arr] (2.95,0.50) .. controls (3.70,0.36) and (3.85,0.30) .. (4.24,0.30);

\begin{scope}[shift={(0,-2.55)}]
  \draw[ax] (0,0) -- (5.20,0);
  \foreach \x/\l in {0/0, 1.30/0.2, 3.25/0.5, 5.20/0.8} {
    \draw[ax] (\x,0) -- (\x,-0.10);
    \node[tk] at (\x,-0.12) {\l};}
  \node[tk,anchor=west] at (5.36,-0.06) {mm};
  \fill[black!12] (0,0.06) rectangle (1.30,0.34);
  \fill[black!22] (1.30,0.06) rectangle (3.25,0.34);
  \fill[accent] (3.25,0.06) rectangle (5.20,0.34);
  \node[font=\scriptsize\color{black!65}] at (0.65,0.20) {off};
  \node[font=\scriptsize\color{black!65}] at (2.28,0.20) {holds};
  \node[font=\scriptsize\color{white}] at (4.22,0.20) {on};
  \node[note,anchor=south west] at (0,0.54) {the same $d$, thresholded, is the contact flag:\\%
                                       on above 0.5\,mm, off below 0.2\,mm};
\end{scope}

\draw[draw=black!25,line width=0.28pt] (0,-3.42) -- (7.90,-3.42);
\node[anchor=north west,align=left,font=\scriptsize,text width=78mm] at (0,-3.58)
  {\textcolor{accent}{\textbf{The inversion.}} Everywhere else in this survey the overlap is the
   error. For a compliant tactile sensor it is the signal, so the quantity a policy elsewhere is
   penalised for producing is the one this sensor is calibrated to read.};
\end{tikzpicture}}}{%
\framebox[0.95\columnwidth]{\rule{0pt}{4.4cm}\textit{figs/fig\_tactiledepth.tex pending}}}
\caption{A compliant tactile sensor reads the overlap depth. The tactile image, the distributed
force and the binary contact flag are all functions of the one quantity every other engine in
this section works to drive to zero.} \label{fig:tactiledepth} \end{figure}

Tactile Genesis inverts the thread. It implements two penetration-depth backends on the
collision geometry, an analytic SDF query and a BVH raycast, and uses the depth as the sensor
signal rather than as an error, so the quantity every other engine treats as a defect sets this
one's operating point.

\label{sec:sim:sim2real}
Most of what is written about this gap is attribution without measurement. PenSpin asserts that
the pure physics gap ``cannot be bridged by extensive domain randomization alone'' while reporting
no experiment that isolates it \cite{penspin_2024}. MuJoCo Playground attributes its LEAP hand
failures to physical flex in low-cost hardware and names more accurate collision geometries as the
fix, without measuring either. DeXtreme lists four candidate causes for its shortfall, including a
malfunctioning Allegro thumb used in most trials, and disambiguates none by experiment
\cite{dextreme_2022}.

What has been measured is narrower, and Section~VII collects the transfer numbers. The one
result that links a quantified model error to transfer ranks system-identification runs by
dynamics-model error and pairs each with real success: the best-identified model grasps 8 of 10
attempts, the worst 0 of 10 \cite[Table~1]{humanoid_sim2real_recipe_2025}. Where a cause has been
pinned down elsewhere it is usually perception or actuation and not contact, with pose
estimators and camera trackers named as the unmodelled source in the real numbers
\cite[App.~C]{pddm_2019}.

The one paper that looks at the contact side is usually read backwards. DeXtreme's real-to-sim
replay interpenetrated because the replayed poses carried the pose estimator's error, not
because the contact model failed, and the paper says so: it could not calibrate physics against
the real system because the state it replayed was already wrong. The commonly cited reading,
that the contact model was at fault, is not what the sentence says.
\section{Training a dexterous policy}
\label{sec:training}

\subsection{Sources of supervision}
\label{sec:train:space}

A dexterous policy is defined by what supervises it, and four sources are in use across the
corpus's 113 papers: a reward function supervises reinforcement learning, a human
demonstration supervises imitation, a human reference trajectory supervises a physics-based
tracker, which is imitation with a simulator in the loop, and nothing supervises a model-based
planner, which is handed a cost and a model instead.

The labels do not partition the corpus. Reinforcement learning is the largest at 54 of the 113
papers, behavior cloning next at 27, distillation at 23, and ten further tags cover the rest.
They sum to far more than 113, because most methods published since 2024 sit on two branches at
once: a reinforcement-learning teacher and a behavior-cloned student are one paper, not two.
\figurename~\ref{fig:taxonomy} draws the tree and the cross-links. The repository carries the
paper-by-paper version of the same thing, one line per method, and that is what to scan when looking
for work comparable to your own.

What the deployed policy looks like once training is done matters more than the name a paper gives
itself, and on that axis the field has converged hard. Section~\ref{sec:train:hybrids} shows why.

\label{sec:train:master}
The emptiest columns of
the method tabulation are the ones a reader most needs. Only 70 of the 113 papers state how many
real trials are behind their headline number, and fewer still state an environment count or how
many unseen objects were tested; Section~\ref{subsec:reporting} takes up what that means for
comparison. A mostly empty line is not a weak method, but it is one that cannot be compared with
any other here. All 113 papers carry the same columns, and Appendix~\ref{app:method} says where
they are and how to read them.

\begin{figure*}[!t]
  \centering
  \IfFileExists{figs/fig_taxonomy.tex}{\begin{tikzpicture}[
  font=\footnotesize,
  bar/.style={draw=none,fill=black!62},
  baracc/.style={draw=none,fill=accent},
  box/.style={draw=black!45,line width=0.3pt,inner sep=3pt,align=left},
  tree/.style={draw=black!45,line width=0.35pt},
  lnk/.style={draw=black!38,line width=0.28pt},
  lbl/.style={font=\scriptsize,anchor=west},
  ann/.style={font=\scriptsize\itshape,anchor=west},
  key/.style={font=\scriptsize\ttfamily\color{black!58},anchor=west},
  num/.style={font=\scriptsize,anchor=west},
  bname/.style={font=\scriptsize\bfseries,anchor=north west,align=left,text width=2.7cm},
  xl/.style={draw=black!45,line width=0.3pt,dash pattern=on 2pt off 1.6pt,-latex},
  xt/.style={draw=accent,line width=0.4pt,dash pattern=on 3pt off 1.4pt on 1pt off 1.4pt,-latex},
  xw/.style={font=\tiny\itshape\color{black!58},fill=white,inner sep=0.6pt},
  ftr/.style={font=\scriptsize,anchor=north west,align=left,text width=17.6cm},
]
\node[box,anchor=west,text width=2.0cm] (root) at (0,-6.48) {\scriptsize\textbf{supervision for a dexterous policy}\\[1pt]{\tiny 112 method papers}};
\draw[box] (2.90,0.22) rectangle (6.40,-2.92);
\node[bname] at (3.02,0.16) {a reward function};
\node[num,anchor=east,font=\scriptsize\bfseries] at (6.28,0.03) {59};
\node[anchor=east,font=\tiny\color{black!55}] at (6.28,-0.23) {41 of 59 in a leaf};
\draw[tree] (root.east) .. controls (2.55,-6.48) and (2.65,-1.35) .. (2.90,-1.35);
\draw[tree] (6.55,0.00) -- (6.55,-2.70);
\draw[tree] (6.40,-1.35) -- (6.55,-1.35);
\draw[tree] (6.55,0.00) -- (6.69,0.00);
\draw[tree] (6.55,-1.08) -- (6.69,-1.08);
\draw[tree] (6.55,-2.16) -- (6.69,-2.16);
\draw[tree] (6.55,-2.70) -- (6.69,-2.70);
\draw[box] (2.90,-3.38) rectangle (6.40,-7.60);
\node[bname] at (3.02,-3.44) {a human demonstration};
\node[num,anchor=east,font=\scriptsize\bfseries] at (6.28,-3.57) {44};
\node[anchor=east,font=\tiny\color{black!55}] at (6.28,-3.83) {26 of 44 in a leaf};
\draw[tree] (root.east) .. controls (2.55,-6.48) and (2.65,-5.49) .. (2.90,-5.49);
\draw[tree] (6.55,-3.60) -- (6.55,-5.22);
\draw[tree] (6.40,-5.49) -- (6.55,-5.49);
\draw[tree] (6.55,-3.60) -- (6.69,-3.60);
\draw[tree] (6.55,-4.14) -- (6.69,-4.14);
\draw[tree] (6.55,-4.68) -- (6.69,-4.68);
\draw[tree] (6.55,-5.22) -- (6.69,-5.22);
\draw[box] (2.90,-8.06) rectangle (6.40,-11.20);
\node[bname] at (3.02,-8.12) {a human reference, tracked with physics};
\node[num,anchor=east,font=\scriptsize\bfseries] at (6.28,-8.25) {12};
\draw[tree] (root.east) .. controls (2.55,-6.48) and (2.65,-9.63) .. (2.90,-9.63);
\draw[tree] (6.55,-8.28) -- (6.55,-10.98);
\draw[tree] (6.40,-9.63) -- (6.55,-9.63);
\draw[tree] (6.55,-8.28) -- (6.69,-8.28);
\draw[tree] (6.55,-8.82) -- (6.69,-8.82);
\draw[tree] (6.55,-9.36) -- (6.69,-9.36);
\draw[tree] (6.55,-9.90) -- (6.69,-9.90);
\draw[tree] (6.55,-10.44) -- (6.69,-10.44);
\draw[tree] (6.55,-10.98) -- (6.69,-10.98);
\draw[box] (2.90,-11.66) rectangle (6.40,-13.18);
\node[bname] at (3.02,-11.72) {no learned policy};
\node[num,anchor=east,font=\scriptsize\bfseries] at (6.28,-11.85) {3};
\draw[tree] (root.east) .. controls (2.55,-6.48) and (2.65,-12.42) .. (2.90,-12.42);
\draw[tree] (6.55,-11.88) -- (6.55,-12.96);
\draw[tree] (6.40,-12.42) -- (6.55,-12.42);
\draw[tree] (6.55,-11.88) -- (6.69,-11.88);
\draw[tree] (6.55,-12.42) -- (6.69,-12.42);
\draw[tree] (6.55,-12.96) -- (6.69,-12.96);
\fill[bar] (6.75,-0.06) rectangle (7.75,0.06);
\node[num] at (7.85,0.00) {20};
\node[lbl] at (8.45,0.09) {PPO in a GPU-parallel simulator, no distillation stage};
\node[key] at (8.45,-0.15) {physhoi\_2023, force\_grasp\_sim2real\_2026, and 18 more};
\fill[baracc] (6.75,-0.60) rectangle (6.80,-0.49);
\node[num] at (7.85,-0.54) {1};
\node[ann] at (8.79,-0.45) {-- population-based training over hyperparameters};
\node[key] at (8.79,-0.69) {dexpbt\_2023};
\fill[bar] (6.75,-1.14) rectangle (7.55,-1.03);
\node[num] at (7.85,-1.08) {16};
\node[lbl] at (8.45,-0.99) {plus teacher-student distillation};
\node[key] at (8.45,-1.23) {unidexgrasp\_2023, dexplore\_2025, and 14 more};
\fill[bar] (6.75,-1.68) rectangle (7.00,-1.57);
\node[num] at (7.85,-1.62) {5};
\node[ann] at (8.79,-1.53) {-- of which the student is a vision policy};
\node[key] at (8.79,-1.77) {hora\_2022, visual\_dexterity\_2022, and 3 more};
\fill[baracc] (6.75,-2.22) rectangle (6.85,-2.10);
\node[num] at (7.85,-2.16) {2};
\node[lbl] at (8.45,-2.07) {reward written by a language model};
\node[key] at (8.45,-2.31) {eureka\_2023, dreureka\_2024};
\fill[bar] (6.75,-2.76) rectangle (7.05,-2.65);
\node[num] at (7.85,-2.70) {6};
\node[lbl] at (8.45,-2.61) {seeded by demonstrations (the RL+demo tag)};
\node[key] at (8.45,-2.85) {pianomime\_2024, dexmv\_2021, and 4 more};
\fill[bar] (6.75,-3.66) rectangle (7.45,-3.54);
\node[num] at (7.85,-3.60) {14};
\node[lbl] at (8.45,-3.51) {teleoperated on the target robot};
\node[key] at (8.45,-3.75) {dexteleop0\_2026, dexora\_2026, and 12 more};
\fill[bar] (6.75,-4.20) rectangle (6.95,-4.09);
\node[num] at (7.85,-4.14) {4};
\node[lbl] at (8.45,-4.05) {wearable or hand-held rig, no robot in the loop};
\node[key] at (8.45,-4.29) {dexcap\_2024, dexumi\_2025, and 2 more};
\fill[bar] (6.75,-4.74) rectangle (7.10,-4.63);
\node[num] at (7.85,-4.68) {7};
\node[lbl] at (8.45,-4.59) {egocentric video, no robot at all};
\node[key] at (8.45,-4.83) {dexmv\_2021, videodex\_2022, and 5 more};
\fill[baracc] (6.75,-5.28) rectangle (6.85,-5.17);
\node[num] at (7.85,-5.22) {2};
\node[lbl] at (8.45,-5.13) {synthetic demonstration generation};
\node[key] at (8.45,-5.37) {dexmimicgen\_2024, deximit\_2026};
\fill[baracc] (6.75,-5.82) rectangle (6.80,-5.71);
\node[num] at (7.85,-5.76) {1};
\node[ann] at (8.79,-5.67) {-- policy shape, an independent choice: action chunking};
\node[key] at (8.79,-5.91) {aloha\_act\_2023};
\fill[bar] (6.75,-6.36) rectangle (7.45,-6.25);
\node[num] at (7.85,-6.30) {14};
\node[ann] at (8.79,-6.21) {-- diffusion};
\node[key] at (8.79,-6.45) {diffusion\_policy\_2023, dp3\_2024, and 12 more};
\fill[bar] (6.75,-6.90) rectangle (7.15,-6.79);
\node[num] at (7.85,-6.84) {8};
\node[ann] at (8.79,-6.75) {-- flow matching};
\node[key] at (8.79,-6.99) {pi0\_2024, groot\_n1\_2025, and 6 more};
\fill[baracc] (6.75,-7.44) rectangle (6.85,-7.33);
\node[num] at (7.85,-7.38) {2};
\node[ann] at (8.79,-7.29) {-- autoregressive action tokens};
\node[key] at (8.79,-7.53) {openvla\_2024, metis\_2025};
\fill[bar] (6.75,-8.34) rectangle (6.95,-8.23);
\node[num] at (7.85,-8.28) {4};
\node[lbl] at (8.45,-8.19) {whole reference including fingers};
\node[key] at (8.45,-8.43) {dextrack\_2025, maniptrans\_2025, and 2 more};
\fill[baracc] (6.75,-8.88) rectangle (6.80,-8.77);
\node[num] at (7.85,-8.82) {1};
\node[lbl] at (8.45,-8.73) {wrist only, fingers learned};
\node[key] at (8.45,-8.97) {objdex\_2024};
\fill[baracc] (6.75,-9.41) rectangle (6.80,-9.30);
\node[num] at (7.85,-9.36) {1};
\node[lbl] at (8.45,-9.27) {object trajectory only};
\node[key] at (8.45,-9.51) {human2sim2robot\_2025};
\fill[baracc] (6.75,-9.95) rectangle (6.80,-9.84);
\node[num] at (7.85,-9.90) {1};
\node[lbl] at (8.45,-9.81) {reference as soft guidance};
\node[key] at (8.45,-10.05) {dexplore\_2025};
\fill[baracc] (6.75,-10.49) rectangle (6.85,-10.38);
\node[num] at (7.85,-10.44) {2};
\node[lbl] at (8.45,-10.35) {no retargeting, reference already on the embodiment};
\node[key] at (8.45,-10.59) {physhoi\_2023, omnigrasp\_2024};
\fill[baracc] (6.75,-11.03) rectangle (6.90,-10.92);
\node[num] at (7.85,-10.98) {3};
\node[lbl] at (8.45,-10.89) {whole-body humanoid, hands carried with the body};
\node[key] at (8.45,-11.13) {dexman\_2025, humanplus\_2024, omnih2o\_2024};
\fill[baracc] (6.75,-11.93) rectangle (6.80,-11.82);
\node[num] at (7.85,-11.88) {1};
\node[lbl] at (8.45,-11.79) {sampled plan, learned model};
\node[key] at (8.45,-12.03) {pddm\_2019};
\fill[baracc] (6.75,-12.47) rectangle (6.80,-12.36);
\node[num] at (7.85,-12.42) {1};
\node[lbl] at (8.45,-12.33) {sampled plan, given model};
\node[key] at (8.45,-12.57) {mjpc\_2022};
\fill[baracc] (6.75,-13.01) rectangle (6.80,-12.90);
\node[num] at (7.85,-12.96) {1};
\node[lbl] at (8.45,-12.87) {smoothed analytic contact model};
\node[key] at (8.45,-13.11) {pang\_global\_planning\_2022};
\draw[xl] (16.20,-1.08) .. controls (16.55,-1.08) and (16.55,-5.49) .. (16.20,-5.49);
\node[xw] at (16.57,-3.29) {distil};
\draw[xl] (16.20,0.00) .. controls (16.82,0.00) and (16.82,-5.22) .. (16.20,-5.22);
\node[xw] at (16.84,-2.61) {generate};
\draw[xl] (16.20,-4.68) .. controls (17.09,-4.68) and (17.09,-9.63) .. (16.20,-9.63);
\node[xw] at (17.11,-7.15) {retarget};
\draw[xl] (16.20,-9.63) .. controls (17.36,-9.63) and (17.36,-5.49) .. (16.20,-5.49);
\node[xw] at (17.38,-7.56) {distil};
\draw[xt] (16.20,-12.42) .. controls (17.63,-12.42) and (17.63,0.00) .. (16.20,0.00);
\node[xw] at (17.65,-6.21) {smooth};
\draw[xl] (16.20,-3.60) .. controls (17.90,-3.60) and (17.90,-2.70) .. (16.20,-2.70);
\node[xw] at (17.92,-3.15) {seed};
\node[ftr] at (0,-13.80) {Eight further rows carry a trajectory-optimisation or MPC tag inside a learned pipeline and are counted on the branch that learns. The double-dashed edge is the one overlap that is a theorem rather than a pipeline: \texttt{pang\_global\_planning\_2022} proves that a policy gradient and an analytic log-barrier relaxation compute the same local model of contact.};
\end{tikzpicture}}{%
    \framebox[0.7\textwidth]{\rule{0pt}{2.2cm}\textit{figs/fig\_taxonomy.tex pending}}}
  \caption{What supervises a dexterous policy, subdivided to where the 113 method papers actually
  differ; each leaf names two of its corpus keys and counts the rest, bars scaled by paper count,
  dashed curves marking overlap.}
  \label{fig:taxonomy}
\end{figure*}

\subsection{Reinforcement learning}
\label{sec:train:rl}

\label{sec:train:rl:recipe}
Sixty of the 113 papers learn from a reward. Forty-nine name PPO as the algorithm, and
the next most frequent, DAPG, appears four times. Forty-three run their own experiments in a
GPU-parallel simulator from the Isaac family or Genesis, and 36 do both. Only 32 papers state an
environment count, with a median of 8192 and a maximum of 64000. Twenty-three papers distill a
privileged teacher into a deployable student, and 16 combine all three of PPO, a GPU simulator and
distillation. That 16-paper intersection is the recipe as it is actually practised.

The hand is almost always an Allegro. It appears in 35 of them, and its hold on in-hand
reorientation is tighter still: more than half of those papers use it, against roughly a fifth
for the Shadow and an eighth for LEAP.

Privilege is handled two ways. OpenAI \cite{openai_dexterity_2018} gives the value network object
and target orientation, joint angles, joint velocities and object velocities the policy never
sees, with a footnote recording that current object orientation was left out of the policy inputs
by accident, and DeXtreme \cite{dextreme_2022} uses the same asymmetric critic with a 2048-unit
LSTM against the actor's 1024. Hora \cite{hora_2022} instead compresses nine privileged object
properties into an eight-dimensional vector regressed from 30 steps of proprioception, and Visual
Dexterity \cite{visual_dexterity_2022} distills twice, from a state teacher to a synthetic point
cloud and then to a rendered one, for a fivefold speedup.

Domain randomisation is the part of the recipe with the least discipline. The best practice in
the corpus makes it adaptive, widening each range when performance at its boundary is good enough
and narrowing it when it is not, and publishing the ranges that search discovered
\cite{openai_rubiks_cube_2019,dextreme_2022}. Below that standard, two shipped configurations
disagree with their own papers about what was randomised at all: one ships a joint-noise range
four times wider than the paper states, in a revision its own README says does not reproduce the
paper \cite{hora_2022}, and the other ships a disturbance force of zero where the paper describes
one \cite{penspin_2024}. Both are in Table~\ref{tab:codegap} with the file and the two values.

\label{sec:train:rl:reward}
The repository puts the 21 in-hand reorientation methods against nine recurring term families and
marks each cell by where the term was found: in the paper, in the code, in both, or in the code
with every shipped configuration setting its weight to zero.

The families are not equally popular. Almost every method has a goal or rotation tracking term,
which is the task itself. After that the common families are an effort penalty, a sparse success
bonus, an object-velocity penalty and a drop penalty, each in about half of them. Only four carry a
contact or force term at all.

Three marks read \emph{code (0)}, because in each the
term is in the released code with every shipped configuration setting its weight to zero: DeXtreme
\cite{dextreme_2022}'s timeout reward, DexPBT \cite{dexpbt_2023}'s fall penalty, and PenSpin
\cite{penspin_2024}'s action penalty. Marking them \emph{code} would tell a reader the code
optimises something the paper does not state, and leaving them unmarked would hide a term that is
in the file. The repository carries the config key and the zero beside each of the three marks.

Only three of those four contact terms are what they look like. AnyRotate \cite{anyrotate_2024}
scores good and bad fingertip contacts, Lin et al. \cite{poise_2026} reward a friction-cone
wrench margin, and Zhao et al. \cite{force_grasp_sim2real_2026} track a commanded grasp force.
The fourth \cite{visual_dexterity_2022} penalises the \emph{object} touching the table, which is
task shaping against using the table as a third finger and not a statement about hand--object
contact, and the flag that enables it defaults to off. So three of the 21 reorientation methods
put a hand--object contact quantity in the reward, and none puts interpenetration in it: the one
method that penalises interpenetration does so while building its references offline, not in the
policy's reward \cite{teledexter_2026}. The physical quality of the contact is not something
this literature optimises.

Term counts range from one to ten. DemoStart \cite{demostart_2024} gives a single binary terminal
reward of 1 and lets a curriculum over demonstration states do the shaping. AnyRotate
\cite{anyrotate_2024} and ViserDex \cite{viserdex_2026} each publish ten weighted terms. ViserDex
is the cleanest specification of the 21, with every term, weight and equation in one appendix
table, and a statement that the same weights are used for every object.

Weights are frequently unrecoverable. Three of the 21 papers give no numeric weight that survives
PDF conversion, because the equation blocks are images, which is a limit of this survey and is
recorded as such. Robot Synesthesia \cite{robot_synesthesia_2023} is a worse case. Its six
coefficients are described only as ``tuned hyper-parameters'' and no number is printed anywhere in
the paper, so the reward it reports cannot be reconstructed from the paper. DexReMoE
\cite{dexremoe_2025} names an angular-velocity penalty weight in prose with no entry in its
hyperparameter table, and DexTrack \cite{dextrack_2025} names an affinity reward that is missing
from its weight table.

\label{sec:train:rl:curricula}
Curricula in this literature relax physics or tighten tolerances. DexPBT \cite{dexpbt_2023}
tightens a success tolerance from 0.075 to 0.01 by a factor of 0.9 every 3000 environment steps
once three successes are logged, and ManipTrans \cite{maniptrans_2025} starts at zero gravity and
high friction and restores both while narrowing a fingertip threshold from 6 cm to 4 cm.
DexMachina \cite{dexmachina_2025} attaches virtual spring-damper controllers to the object, lets
them do the task at first, then decays their gains to zero. The two that ablate the curriculum
report it is load-bearing: RotateIt \cite{rotateit_2023} sets its rotation-penalty weight to zero
and ramps it to $-0.1$, because applying it from the start makes the policy learn only to hold the
object still, and ViserDex \cite{viserdex_2026}, whose performance-based curriculum runs over
regularisation weight, action latency and time between successes, fails completely without the
regularisation component.

Population-based training appears once. DexPBT \cite{dexpbt_2023} trains populations of up to
32 agents, mutating the middle third of each population and replacing the bottom with mutated
copies of the top, and reports 0.32 trillion environment steps for its largest run. The
technique is routine outside robotics and nothing else in this corpus uses it.

A language model can write the reward outright, which removes the hand-tuning and replaces it
with a search whose objective nobody has read. Eureka \cite{eureka_2023} has GPT-4 write the
reward function directly, constrained to return a total and a dictionary of named components,
and feeds per-component statistics back between iterations. Its appendix gives an example whose
reward correlates at $-0.26$ with the human-written one and still scores 1.45 on the
human-normalized metric, the strongest argument in the corpus that hand-tuned weights are not a
ceiling. DrEureka \cite{dreureka_2024} moves the safety requirement into the prompt instead of
a penalty term, asking in words for a cube rotating at about 0.25 radians per second with
fingers penalized for leaving their initial pose. Neither can be checked against a file.
Eureka's generated rewards are written at run time to a gitignored directory and never
committed, so the reward behind any reported number cannot be recovered from the repository,
though its README documents the paper's budget of five iterations and sixteen samples as the
default and marks the shipped one-iteration block a fast-test preset. DrEureka's repository
contains only the locomotion and globe-walking trees, so its LEAP-hand reward has no code at
all.

\label{sec:train:rl:stalls}
Reinforcement learning
stalls on objectives that are not learnable as stated. AnyRotate \cite{anyrotate_2024} found
the angular-velocity objective unlearnable in the multi-axis setting and replaced it with a
moving keypoint target. RotateIt \cite{rotateit_2023} reports that its multi-axis policy ``does
not converge when training with only reinforcement learning'' and adds an imitation loss
against single-axis oracles. Eureka \cite{eureka_2023} cannot spin a pen from scratch and needs
a pretrain-then-finetune split, with the from-scratch ablation failing to complete one cycle.
Section~\ref{sec:train:track}'s contact-graph term exists because of the general shape of the
problem: contact moves the object off the reference, so return goes down, so the policy learns
not to touch the object.

It stalls on geometry too. Hora \cite{hora_2022} fails on objects under 4 cm across because the
fingers collide with each other. What it does not stall on is publication: the real-trial
counts behind these headlines are small enough that Section~\ref{subsec:reporting} treats them
as the reporting problem they are.

\subsection{Learning from human data}
\label{sec:train:human}

\label{sec:train:human:teleop} Thirty papers in this corpus describe a system for getting human
motion onto a robot hand; the repository lists them with the retargeting objective compressed to
one clause each.

Retargeting splits four ways. The dominant family matches task-space vectors between the human
and the robot hand, weighting fingertip distances and adding a minimum separation between the
primary fingers, solved online; DexPilot \cite{dexpilot_2020} set that pattern and four later
systems use the same formulation
\cite{anyteleop_2023,open_television_2024,bunny_visionpro_2024,ace_teleop_2024}. The second copies
joint angles wherever the kinematics allow it and solves only the thumb by inverse kinematics
\cite{holo_dex_2022}. The third learns the map offline, which buys about an order of magnitude in
rate over online optimisation and raises configuration-space coverage from 38 to 90 percent
\cite{geometric_retargeting_2025}. The fourth removes retargeting from the loop altogether,
building an exoskeleton whose fingertip workspace is designed to match the target hand and
regressing encoder values straight to motor values \cite{dexumi_2025}.

What that tabulation does not carry is the point. Fewer than half of the teleoperation papers
state a latency at all, and only a handful state a rig cost. Where costs are stated they are low
and falling, from a few hundred dollars for a headset or a glove \cite{holo_dex_2022,doglove_2025}
to a few thousand for a full mocap backpack \cite{dexcap_2024,bidex_teleop_2024}. Latency is not
one quantity either: some figures are end to end \cite{dexpilot_2020,holo_dex_2022}, others are a
single retargeting stage \cite{geometric_retargeting_2025,anyteleop_2023}, and several cells hold a
control rate instead, so the column's spread is not a range. Fewer than half report how many
trajectories they collected, and fewer still report hours of data.

Where throughput is measured, human collection beats teleoperation by a wide margin. Qiu et al.
\cite{humanoid_policy_human_policy_2025} time a grasp demonstration at 3.79 seconds for a bare
human and 19.72 seconds through a teleoperated humanoid, and a pour at 4.81 against 37.31
seconds. It attributes the gap to retargeting latency and to the workspace of a 7-DoF arm, not
to the hand.

\label{sec:train:human:arch}
Four architectures cover the
corpus. Action chunking came from \cite{aloha_act_2023}, which predicts a chunk of joint
targets with a CVAE and combines overlapping chunks by exponentially weighted ensembling,
reaching 80 to 90 percent on fine bimanual tasks from about 50 demonstrations each. Diffusion
came from Diffusion Policy \cite{diffusion_policy_2023}, which denoises an action sequence and
reports a 46.9 percent average success improvement over LSTM-GMM, IBC and BET across 15 tasks.
3D Diffusion Policy \cite{dp3_2024} swaps the image encoder for a small point-cloud encoder and
reports 74.4 percent against 59.8 percent across 72 simulated tasks, and 85.0 against 35.0
percent on four real tasks from 40 demonstrations each. Flow matching is the 2024-onward
default for large models and carries eight papers. Discrete action tokens are the fourth, used
by \cite{openvla_2024} and \cite{metis_2025}, and are the only one of the four that needs no
continuous head at all.

Data generation is a separate lever and is underrated. DexMimicGen \cite{dexmimicgen_2024} turns
60 human demonstrations into 21,000 simulated ones, and a real two-handed humanoid trained on the
generated set succeeds where the same policy trained on the source demonstrations never does. Dex1B
\cite{dex1b_2025} iterates optimisation, a proposal model and a simulator filter into roughly a
billion synthetic grasps, and beats the prior best by a wide margin on its own benchmark.

\label{sec:train:human:video}
Seven papers train a dexterous-hand policy from video with no teleoperation at any stage. DexMV
\cite{dexmv_2021} retargets 700 self-recorded demonstrations by matching palm-to-fingertip
task-space vectors and estimating actions by inverse dynamics, VideoDex \cite{videodex_2022} mines
965 trajectories from EpicKitchens to pretrain a neural dynamic policy, and DexVIP
\cite{dexvip_2022} goes further into the reward, clustering 715 curated HowTo100M frames into one
consensus grasp pose per object category and adding it as a reward term. OKAMI \cite{okami_2024},
Lum et al. \cite{human2sim2robot_2025} and Guzey et al. \cite{hudor_2024} work from a single video
each, and Goswami et al. \cite{wm_dex_human_videos_2025} pretrain a world model on 829 hours of
EgoDex.

\label{sec:train:human:gap} Three papers measure the gap directly. OKAMI \cite{okami_2024} runs
the same reference-plan and warping pipeline with and without human body and hand retargeting,
and the version without it loses 75.0, 42.1, 82.0 and 74.0 points across four task and setting
pairs. That is the largest clean embodiment-gap number in the corpus.

Qiu et al. \cite{humanoid_policy_human_policy_2025} ablate two mitigations separately on the
same vertical-grasp task. Removing the interpolation that slows human trajectories down cuts
success to a quarter of the full method's, and removing the unified state space collapses it to
zero, because the policy is handed a shortcut for telling the two embodiments apart. EgoZero
\cite{egozero_2025} reports the sharpest version of the same effect: removing its 3D augmentation
drops every one of its seven tasks to no successes at all, and substituting monocular depth for
triangulated depth does the same, because the best metric depth models it tested carry more than 5
cm of error.

Chen et al. \cite{objdex_2024} supply the counterexample. Their method retargets only the wrist
and lets reinforcement learning discover finger motion, and their ablation that adds a
fingertip-matching reward ``does not yield benefits and even leads to lower performance''. Richer
human correspondence is not uniformly better, and this is the one result in the corpus that says
so with an ablation.

\label{sec:train:track}
Twelve papers sit in the human-reference tracking family. Eight track a human hand on an
object and are covered here; the other four sit at the edges, Lum et al.
\cite{human2sim2robot_2025} tracking only the object's trajectory, DexMan \cite{dexman_2025}
retargeting bimanual video onto a full humanoid, and HumanPlus \cite{humanplus_2024} and OmniH2O
\cite{omnih2o_2024} tracking whole-body motion. A reference is retargeted onto the robot and a
policy trained to make the simulated body follow it: the reference supplies the shaping reward
engineering would otherwise have to invent, and the simulator the physical consistency pure
imitation lacks.

The multiplicative reward that most human-reference methods now use has a single source.
PhysHOI \cite{physhoi_2023} is the origin of the form. It multiplies a body term, an object
term, an interaction-graph term and a contact-graph term, and reaches 95.4 percent success on
GRAB against 27.0 percent for a DeepMimic baseline. The contact-graph term exists to stop the
policy learning not to touch the object. Omnigrasp \cite{omnigrasp_2024} replaces the action
space with a pretrained latent motion prior and reports 94.6 percent grasp success and 84.8
percent trajectory success on GRAB, while stating outright that it omits penetration metrics.

The next step was to stop treating demonstrations as fixed and let the method generate its own.
DexTrack \cite{dextrack_2025} adds an imitation loss to PPO and mines its own demonstrations
through a homotopy search over easier neighbouring trajectories, beating the best baseline by
about eight points at either threshold. It has a penetration-depth formula and applies it only
to its input references, never to its own rollouts, and presents tolerance of ``severe
hand-object penetrations'' as robustness. ManipTrans \cite{maniptrans_2025} freezes a generalist
hand imitator and trains a per-task residual on top, with single-hand success about half again
its bimanual figure. Its stance on contact is to raise the friction coefficient above the real
value instead of modelling skin deformation, and its real deployment is open-loop replay.

Two papers in this corpus ran the same curriculum and disagreed about whether it helps.
DexMachina \cite{dexmachina_2025} moves to Genesis with 12,000 environments and six hand URDFs,
and drives the object with virtual controllers that decay to zero. Its re-implementation of
\cite{maniptrans_2025}'s curriculum does not beat no curriculum on its own setup, which is a
direct disagreement between two papers worth quoting in both directions. Dexplore
\cite{dexplore_2025} drops the explicit retargeting stage and treats the reference as soft
guidance, with termination thresholds derived from the failure rate, reaching 87.7 percent on
GRAB with an Inspire hand.

\begin{figure}[!t]
  \centering
  \IfFileExists{figs/fig_retarget.tex}{\resizebox{\columnwidth}{!}{\begin{tikzpicture}[
  font=\footnotesize,
  palm/.style={draw=black!45,fill=wash,line width=0.3pt},
  ref/.style={draw=black!50,line width=0.5pt,dash pattern=on 1.6pt off 1.4pt},
  reftip/.style={draw=black!50,line width=0.5pt,dash pattern=on 1.6pt off 1.4pt,fill=none},
  rob/.style={draw=accent,line width=0.7pt},
  robtip/.style={draw=accent,line width=0.7pt,fill=white},
  obj/.style={draw=black!65,line width=0.5pt,fill=black!12},
  over/.style={draw=none,fill=accent,fill opacity=0.22},
  dim/.style={draw=accent,line width=0.5pt,<->},
  lead/.style={draw=black!45,line width=0.28pt},
  ax/.style={draw=black!55,line width=0.4pt},
  tol/.style={draw=black!45,line width=0.35pt,dash pattern=on 1.4pt off 1.2pt},
  bar/.style={draw=black!45,fill=black!22,line width=0.3pt},
  hd/.style={font=\scriptsize\bfseries,anchor=north,align=center},
  sub/.style={font=\scriptsize\color{black!60},anchor=north,align=center,text width=34mm},
  glab/.style={font=\scriptsize\color{black!60},anchor=west},
  alab/.style={font=\scriptsize\color{accent},anchor=west},
  tk/.style={font=\scriptsize\color{black!60},anchor=north},
  note/.style={font=\scriptsize,anchor=north west,align=left},
]

\node[hd] at (1.80,3.12) {what the joint-angle error sees};

\draw[palm] (0.10,2.20) rectangle (0.45,2.70);
\draw[ref] (0.45,2.45) -- ++(-4:0.70) coordinate (t1)
                       -- ++(-22:0.55) coordinate (t2)
                       -- ++(-46:0.42) coordinate (t3);
\draw[reftip] (t3) circle (0.15);
\draw[rob]    (0.45,2.45) -- ++(-10:0.70) coordinate (r1)
                          -- ++(-31:0.55) coordinate (r2)
                          -- ++(-58:0.42) coordinate (r3);
\draw[robtip] (r3) circle (0.15);
\fill[accent] (0.45,2.45) circle (0.05);
\fill[accent] (r1) circle (0.05);
\fill[accent] (r2) circle (0.05);

\draw[lead] (t3) -- (2.28,2.02);
\node[glab] at (2.30,2.02) {reference};
\draw[lead] (r3) -- (2.28,1.62);
\node[alab] at (2.30,1.62) {retargeted};

\draw[tol] (0.45,1.30) -- (3.05,1.30);
\node[font=\scriptsize\color{black!55},anchor=south west] at (0.47,1.33) {tolerance};
\draw[ax] (0.45,0.80) -- (3.05,0.80);
\foreach \x/\h/\l in {0.80/0.18/MCP, 1.60/0.27/PIP, 2.40/0.36/DIP} {
  \draw[bar] (\x,0.80) rectangle ++(0.34,\h);
  \node[tk] at (\x+0.17,0.78) {\l};}
\node[sub,text width=38mm] at (1.78,0.50) {per-joint angle error: a few degrees each, all inside
                                           tolerance};

\draw[draw=black!20,line width=0.28pt] (3.88,-0.22) -- (3.88,3.02);

\node[hd] at (6.10,3.12) {what happens at the surface};

\draw[obj] (4.35,2.05) rectangle (6.10,2.19);
\draw[lead] (4.35,2.12) -- (4.18,2.12);
\node[font=\scriptsize\color{black!60},anchor=east] at (4.15,2.12) {$t$};

\draw[reftip] (5.15,2.49) circle (0.30);
\begin{scope}
  \clip (4.35,2.05) rectangle (6.10,2.19);
  \fill[over] (5.15,2.07) circle (0.30);
\end{scope}
\draw[robtip,fill=none] (5.15,2.07) circle (0.30);

\draw[dim] (6.42,2.49) -- (6.42,2.07);
\node[alab] at (6.50,2.28) {$\sum_i \ell_i \varepsilon_i$};

\draw[lead] (4.62,2.05) -- (4.48,1.74);
\node[glab,font=\scriptsize\color{black!60},anchor=north west] at (4.18,1.72)
     {a thin object, in section};
\node[sub,text width=40mm] at (6.15,1.26) {the composed displacement is larger than the object is
                                           thick, so the fingertip is through it and the depth is
                                           positive};

\draw[draw=black!25,line width=0.28pt] (0,-0.34) -- (8.20,-0.34);
\node[note,text width=79mm] at (0,-0.50)
  {\textcolor{accent}{\textbf{The gap.}} Joint-angle error is a sum over joints and carries no term
   for the object, so it cannot see the right-hand panel at all. Errors small enough to pass at
   every joint compose along the finger into a fingertip displacement larger than a thin object is
   thick, and the pose is scored as good with the object inside the finger. Sec.~V's one method
   that measures penetration measures it on the reference, before the policy runs.};
\end{tikzpicture}}}{%
    \framebox[0.95\columnwidth]{\rule{0pt}{4.4cm}\textit{figs/fig\_retarget.tex pending}}}
  \caption{A retargeted pose can be scored good and still be inside the object. The joint-angle
  error is a sum over joints with no term for the object; a few degrees at each joint compose
  along the finger into a fingertip displacement larger than a thin object is thick.}
  \label{fig:retarget}
\end{figure}

TopoRetarget \cite{toporetarget_2026} is the only corpus method that quantifies penetration as an
evaluation quantity, reporting maximum depth and the fraction of frames past 2 mm, and
constraining it during retargeting with a 1 mm soft tolerance and a 30 mm hard bound. It reaches
87.5 percent on its own pen-spinning set against 46.9 for the best baseline. It does not
re-measure penetration after the tracking policy runs, so the property it constrains is a property
of the reference and not of the behavior. \figurename~\ref{fig:retarget} draws why the usual
retargeting objective cannot see the failure at all: joint-angle error is summed over joints and
carries no term for the object, so errors small enough to pass at every joint compose along the
finger into a fingertip displacement larger than a thin object is thick, and the pose is scored as
good with the object inside the finger.

The pattern holds across all eight, and it is the reference-versus-rollout split in its sharpest
form. Penetration is handled at the reference, if at all, and never at the rollout. Chen et al.
\cite{objdex_2024} complete the set with the only real-robot numbers among them, from 100 percent
on a microwave and a laptop down to 41.2 percent on a ketchup bottle over 20 trials each.

Human data does not port across hands, and the map is usually
left unstated. Fifty-four papers use human data and name 45 distinct hand strings between
them, and twenty of those appear among the 30 teleoperation systems with a stated retargeting
objective. The other 33 never say how the human motion reached the hand, and the stated
objectives do not converge, from a fingertip keypoint-vector energy in \cite{anyteleop_2023} to
a per-joint regression from exoskeleton encoders in \cite{dexumi_2025} to no retargeting at all
in \cite{egozero_2025}. A shared benchmark could make these mappings comparable. TopoRetarget's
format \cite{toporetarget_2026} is a useful starting point: it reports contact precision,
contact alignment, maximum penetration and the share of frames past 2 mm, per hand and per
dataset.

\subsection{Generalist and vision-language-action policies}
\label{sec:train:vla}

The finding is the size of the hand. Eighteen papers carry the generalist tag. Fourteen of them
settle whether the reported evaluation ran on a multi-fingered hand. Eight of those fourteen did,
six report no hand result at all, and the remaining four never say, GR00T N1.6
\cite{groot_n16_2025} naming no end effector anywhere on its page.

Five of the eight state a hand size, and that comparison has to be made in actuated degrees of
freedom rather than joints, for the reason Section~III opens with. Four of the five use a
six-actuator hand and one a twelve, so their median is six. The reinforcement-learning literature
of Section~\ref{sec:train:rl} works an order of magnitude above that: its median hand actuates 16
joints. No generalist policy here that settles the question evaluates on a hand in that band, and
the two that would reach it on paper are the two that never say whether the hand was in the
evaluation at all.

The six with no hand result are parallel-jaw throughout ($\pi_0$ \cite{pi0_2024}, whose released
code encodes each ALOHA gripper as one scalar, $\pi_{0.5}$ \cite{pi05_2025}, $\pi^{*}_{0.6}$
\cite{pistar06_2025}, OpenVLA \cite{openvla_2024}, Helix \cite{helix_2025}, which claims
individual-finger control of a 35-DoF upper body with no success rate or trial count for any task,
and Gemini Robotics \cite{gemini_robotics_2025}, whose Apollo hand appears in a figure caption
with no number attached). Its successor \cite{gemini_robotics_15_2025} closed that gap with Apollo
progress scores, not success rates, of 0.74 down to 0.62 across five generalisation axes. DexVLA
\cite{dexvla_2025} is the borderline case the other way: a hand in one real-robot task family,
grippers in the other five and in all 91 pretraining tasks.

The released code also drifts. GR00T N1 \cite{groot_n1_2025}'s repository is a later N1.7
generation with a different backbone, a different action horizon and embodiments the paper does
not describe, and the $\pi_{0.5}$ repository \cite{pi05_2025} supports only the flow-matching
head, not the hybrid recipe the paper describes.

The gap is one of size, not absence. The
generalist policies that do touch a hand run it at roughly a third of the actuation the
reinforcement-learning literature assumes, and the papers that never settle the question are the
smaller half of the same gap. The standard
generalist suite, with the hand's actuated degrees of freedom and its vendor printed beside the
number, should include at least one dexterous-hand task.

\subsection{Model-based baselines and hybrids}
\label{sec:train:mbc}

Three corpus methods solve dexterous tasks without learning a policy, and they are the control
group the rest of this section lacks.

Not every method here learns a policy. A smaller line plans through a model instead, and reaches
dexterous tasks on a fraction of the data. PDDM \cite{pddm_2019} learns an ensemble of dynamics
models and plans through it with an MPPI-style optimiser, replanning every step, on a few hours of
data in simulation and a few more on a real Shadow Hand. Predictive Sampling \cite{mjpc_2022}
removes the learned model too, sampling rollouts through MuJoCo itself at millisecond latency, and
reorients a cube in real time from scratch. It also reports no success criterion, no trial count
and no real-robot result.

The work of Pang et al. \cite{pang_global_planning_2022} is the most substantial of the three.
It proves that the randomised smoothing implicit in reinforcement learning and an analytic
log-barrier relaxation compute the same local linear model of contact, then uses the analytic
version inside a trajectory optimiser and an RRT. Allegro in-hand rotation takes 19.59 seconds
to optimize and its plate-pickup task 117.16 seconds of planning on a 16-core CPU, against the
GPU-days of Section~\ref{sec:train:rl}, and it is the only method here that imposes
non-penetration as a hard constraint. Its own limitation is stated: its 3D systems transfer to
hardware far worse than its 2D ones, because the quasi-dynamic assumption breaks and planned
grasps miss contacts under a second-order solver.

\label{sec:train:hybrids}
The recurring
shape is reinforcement learning in simulation distilled into a policy that looks like an
imitation policy, and it appears in four variants. The first distills a privileged teacher into
a vision student inside one paper, which is \cite{hora_2022}, \cite{visual_dexterity_2022},
\cite{rotateit_2023}, \cite{robot_synesthesia_2023} and \cite{viserdex_2026}. The second uses
reinforcement learning as a demonstration factory. DexTrack \cite{dextrack_2025} mines
demonstrations with per-trajectory RL and trains a generalist on them, and ManipTrans
\cite{maniptrans_2025} does the same to build a 3.3K-episode dataset. DexterityGen
\cite{dexteritygen_2025} goes further, training a diffusion controller on ten billion simulated
grasp-to-grasp transitions and projecting a human teleoperator's coarse command onto it, taking
four reorientation tasks from 0 out of 20 under raw teleoperation to 12, 13, 10 and 9 out of
20. The third generates data without reinforcement learning at all, which is
\cite{dexmimicgen_2024}, \cite{dex1b_2025} and \cite{deximit_2026}; Dex1B's entry is classed as a
dataset, so \figurename~\ref{fig:taxonomy}'s leaf, which counts papers, holds the other
two. The fourth wraps an RL policy in a residual, which is \cite{resdex_2024} at 88.8 percent
over 3,200 objects in 12 GPU-hours.

In the variants that dominate 2025 and 2026 work, the reward is no longer the objective of the
deployed policy. It is the objective of the process that produced the deployed policy's training
data. Reward engineering has moved upstream into data curation, and every reward-shaping
pathology in Section~\ref{sec:train:rl} now reaches the shipped policy through a dataset and not
a gradient.

\subsection{Paper against released code}
\label{sec:train:code}

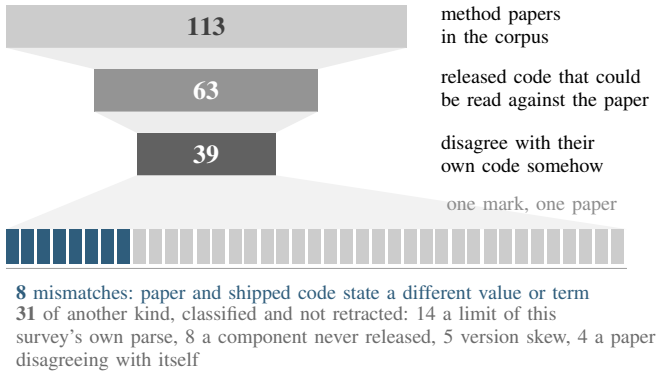
\begin{figure}[!t]
  \centering
  \IfFileExists{figs/fig_codegap.tex}{\resizebox{\columnwidth}{!}{\begin{tikzpicture}[
  font=\footnotesize,
  bar/.style={draw=none,fill=black!62},
  barmid/.style={draw=none,fill=black!42},
  barlight/.style={draw=none,fill=black!20},
  baracc/.style={draw=none,fill=accent},
  seg/.style={draw=white,line width=0.5pt},
  lnk/.style={draw=black!38,line width=0.28pt},
  ghost/.style={draw=black!45,line width=0.3pt,dash pattern=on 1.2pt off 1.2pt},
  lbl/.style={font=\scriptsize},
  small/.style={font=\scriptsize\color{black!55}},
  num/.style={font=\scriptsize\color{black!55}},
]

\fill[barlight] (0.00,0.00) rectangle (4.90,-0.52);
\node[font=\small\bfseries\color{black!75}] at (2.45,-0.26) {113};
\node[lbl,anchor=west,align=left] at (5.20,-0.26) {method papers\\in the corpus};
\fill[black!7] (0.00,-0.52) -- (4.90,-0.52) -- (3.82,-0.78) -- (1.08,-0.78) -- cycle;
\fill[barmid] (1.08,-0.78) rectangle (3.82,-1.30);
\node[font=\small\bfseries\color{white}] at (2.45,-1.04) {63};
\node[lbl,anchor=west,align=left] at (5.20,-1.04) {released code that could\\be read against the paper};
\fill[black!7] (1.08,-1.30) -- (3.82,-1.30) -- (3.30,-1.56) -- (1.60,-1.56) -- cycle;
\fill[bar] (1.60,-1.56) rectangle (3.30,-2.08);
\node[font=\small\bfseries\color{white}] at (2.45,-1.82) {39};
\node[lbl,anchor=west,align=left] at (5.20,-1.82) {disagree with their\\own code somehow};
\fill[black!5] (1.60,-2.08) -- (3.30,-2.08) -- (7.60,-2.74) -- (0,-2.74) -- cycle;
\node[anchor=south east,font=\scriptsize\color{black!45}] at (7.60,-2.68) {one mark, one paper};
\fill[baracc] (0.00,-2.74) rectangle (0.15,-3.16);
\fill[baracc] (0.19,-2.74) rectangle (0.34,-3.16);
\fill[baracc] (0.39,-2.74) rectangle (0.54,-3.16);
\fill[baracc] (0.58,-2.74) rectangle (0.73,-3.16);
\fill[baracc] (0.78,-2.74) rectangle (0.93,-3.16);
\fill[baracc] (0.97,-2.74) rectangle (1.12,-3.16);
\fill[baracc] (1.17,-2.74) rectangle (1.32,-3.16);
\fill[baracc] (1.36,-2.74) rectangle (1.51,-3.16);
\fill[barlight] (1.56,-2.74) rectangle (1.71,-3.16);
\fill[barlight] (1.75,-2.74) rectangle (1.90,-3.16);
\fill[barlight] (1.95,-2.74) rectangle (2.10,-3.16);
\fill[barlight] (2.14,-2.74) rectangle (2.29,-3.16);
\fill[barlight] (2.34,-2.74) rectangle (2.49,-3.16);
\fill[barlight] (2.53,-2.74) rectangle (2.68,-3.16);
\fill[barlight] (2.73,-2.74) rectangle (2.88,-3.16);
\fill[barlight] (2.92,-2.74) rectangle (3.07,-3.16);
\fill[barlight] (3.12,-2.74) rectangle (3.27,-3.16);
\fill[barlight] (3.31,-2.74) rectangle (3.46,-3.16);
\fill[barlight] (3.51,-2.74) rectangle (3.66,-3.16);
\fill[barlight] (3.70,-2.74) rectangle (3.85,-3.16);
\fill[barlight] (3.90,-2.74) rectangle (4.05,-3.16);
\fill[barlight] (4.09,-2.74) rectangle (4.24,-3.16);
\fill[barlight] (4.29,-2.74) rectangle (4.44,-3.16);
\fill[barlight] (4.48,-2.74) rectangle (4.63,-3.16);
\fill[barlight] (4.68,-2.74) rectangle (4.83,-3.16);
\fill[barlight] (4.87,-2.74) rectangle (5.02,-3.16);
\fill[barlight] (5.07,-2.74) rectangle (5.22,-3.16);
\fill[barlight] (5.26,-2.74) rectangle (5.41,-3.16);
\fill[barlight] (5.46,-2.74) rectangle (5.61,-3.16);
\fill[barlight] (5.65,-2.74) rectangle (5.80,-3.16);
\fill[barlight] (5.85,-2.74) rectangle (6.00,-3.16);
\fill[barlight] (6.04,-2.74) rectangle (6.19,-3.16);
\fill[barlight] (6.24,-2.74) rectangle (6.39,-3.16);
\fill[barlight] (6.43,-2.74) rectangle (6.58,-3.16);
\fill[barlight] (6.63,-2.74) rectangle (6.78,-3.16);
\fill[barlight] (6.82,-2.74) rectangle (6.97,-3.16);
\fill[barlight] (7.02,-2.74) rectangle (7.17,-3.16);
\fill[barlight] (7.21,-2.74) rectangle (7.36,-3.16);
\fill[barlight] (7.41,-2.74) rectangle (7.56,-3.16);
\draw[lnk] (0,-3.22) -- (7.60,-3.22);
\node[anchor=north west,align=left,font=\scriptsize\color{accent}] at (0,-3.30) {\textbf{8} mismatches: paper and shipped code state a different value or term};
\node[anchor=north west,align=left,font=\scriptsize\color{black!58}] at (0,-3.56) {\textbf{31} of another kind, classified and not retracted: 14 a limit of this\\survey's own parse, 8 a component never released, 5 version skew, 4 a paper\\disagreeing with itself};
\end{tikzpicture}}}{%
    \framebox[0.95\columnwidth]{\rule{0pt}{2.2cm}\textit{figs/fig\_codegap.tex pending}}}
  \caption{Papers whose released code was readable, and the \fnKept{} where a value in it differs from the value the paper prints. Table~\ref{tab:codegap} names them, with the file and both values.}
  \label{fig:codegap}
\end{figure}

\begin{table*}[!t]
\centering
\caption{The 8 papers whose repository states a different value from the paper. Each file links to the revision read.}
\label{tab:codegap}
\scriptsize
\setlength{\tabcolsep}{3pt}
\setlength{\emergencystretch}{1.5em}
\begin{tabular}{>{\raggedright\arraybackslash\hspace{0pt}}p{56pt}>{\raggedright\arraybackslash\hspace{0pt}}p{30pt}>{\raggedright\arraybackslash\hspace{0pt}}p{84pt}>{\raggedright\arraybackslash\hspace{0pt}}p{94pt}>{\raggedright\arraybackslash\hspace{0pt}}p{106pt}>{\raggedright\arraybackslash\hspace{0pt}}p{106pt}}
\toprule
\rowcolor{wash} \hdr{method} & \hdr{conf.} & \hdr{repository} & \hdr{file, linked at the revision read} & \hdr{what the paper prints} & \hdr{what that file contains} \\
\midrule
PDDM~\cite{pddm_2019} & high & \texttt{google-research/}\allowbreak{}\texttt{pddm} & \href{https://github.com/google-research/pddm/blob/06b88cdbafa7fc35451b3feb3e01aa6726113f72/pddm/envs/cube/cube_env.py}{\texttt{cube\_\allowbreak{}env.\allowbreak{}py}}\newline \texttt{\_\allowbreak{}get\_\allowbreak{}obs} & observation dimension 46 for in-hand reorientation, Table 2 & the six fields \texttt{\_\allowbreak{}get\_\allowbreak{}obs} concatenates sum to 39 on the code's own inline comments \\
DeXtreme~\cite{dextreme_2022} & high & \texttt{isaac-sim/}\allowbreak{}\texttt{IsaacGymEnvs} & \href{https://github.com/isaac-sim/IsaacGymEnvs/blob/aeed298638a1f7b5421b38f5f3cc2d1079b6d9c3/isaacgymenvs/cfg/task/AllegroHandDextremeADR.yaml}{\texttt{Allegro\allowbreak{}Hand\allowbreak{}Dextreme\allowbreak{}ADR.\allowbreak{}yaml}}\newline \texttt{action\allowbreak{}Delta\allowbreak{}Penalty\allowbreak{}Scale} & action-delta penalty weight -0.25, Table 2 & \texttt{action\allowbreak{}Delta\allowbreak{}Penalty\allowbreak{}Scale: -0.\allowbreak{}2}, and -0.01 in the ManualDR yaml beside it \\
Visual Dexterity~\cite{visual_dexterity_2022} & high & \texttt{Improbable-AI/}\allowbreak{}\texttt{dexenv} & \href{https://github.com/Improbable-AI/dexenv/blob/ad9634e9d26cf5555d18728244a58f5b412c1eb2/dexenv/conf/dclaw.yaml}{\texttt{dclaw.\allowbreak{}yaml}}\newline \texttt{alg.\allowbreak{}num\_\allowbreak{}envs} & 32,000 teacher training environments, Table S1 & \texttt{num\_\allowbreak{}envs: 8000}, and 16384 in the parent \texttt{hand\_\allowbreak{}default.\allowbreak{}yaml} \\
DexPBT~\cite{dexpbt_2023} & high & \texttt{NVIDIA-Omniverse/}\allowbreak{}\texttt{IsaacGymEnvs} & \href{https://github.com/NVIDIA-Omniverse/IsaacGymEnvs/blob/aeed298638a1f7b5421b38f5f3cc2d1079b6d9c3/isaacgymenvs/tasks/allegro_kuka/allegro_kuka_base.py}{\texttt{allegro\_\allowbreak{}kuka\_\allowbreak{}base.\allowbreak{}py}}\newline \texttt{compute\_\allowbreak{}kuka\_\allowbreak{}reward} & four staged reward terms, Sec. III-C & eight summed components, \texttt{hand\_\allowbreak{}delta\_\allowbreak{}penalty} multiplied by 0 with the comment \texttt{currently disabled} \\
PhysHOI~\cite{physhoi_2023} & high & \texttt{wyhuai/}\allowbreak{}\texttt{PhysHOI} & \href{https://github.com/wyhuai/PhysHOI/blob/6095c605e22bd01f78ed356e8b08250edf3c5da6/physhoi/env/tasks/physhoi.py}{\texttt{physhoi.\allowbreak{}py}}\newline \texttt{compute\_\allowbreak{}humanoid\_\allowbreak{}reward} & object-rotation weights 0.1 and 0.01 for GRAB, Table 4 & \texttt{eor} and \texttt{eorv} set to \texttt{torch.\allowbreak{}zeros\_\allowbreak{}like(ep)}, the computation commented out on the same lines \\
UniDexGrasp~\cite{unidexgrasp_2023} & high & \texttt{PKU-EPIC/}\allowbreak{}\texttt{UniDexGrasp} & \href{https://github.com/PKU-EPIC/UniDexGrasp/blob/36c9bfcf7cedc987dd15e2384c548e51d5aadc21/dexgrasp_policy/dexgrasp/tasks/shadow_hand_grasp.py}{\texttt{shadow\_\allowbreak{}hand\_\allowbreak{}grasp.\allowbreak{}py}}\newline \texttt{compute\_\allowbreak{}hand\_\allowbreak{}reward, goal\_\allowbreak{}cond branch} & seven named weights from 0.1 to 10, Table 7 & a \texttt{torch.\allowbreak{}where} cascade on literals -0.5, -1.0, 0.9, 0.1, 0.2, none of the seven present \\
PenSpin~\cite{penspin_2024} & medium & \texttt{HaozhiQi/}\allowbreak{}\texttt{penspin} & \href{https://github.com/HaozhiQi/penspin/blob/5035c52dc949e81313af71d9a0ca5ead8b463a5a/configs/task/AllegroHandHora.yaml}{\texttt{Allegro\allowbreak{}Hand\allowbreak{}Hora.\allowbreak{}yaml}}\newline \texttt{force\allowbreak{}Scale} & a disturbance force of 0.2 times object mass at probability 0.25, Table 8 & \texttt{force\allowbreak{}Scale: 0.\allowbreak{}0} and \texttt{random\allowbreak{}Force\allowbreak{}Prob\allowbreak{}Scalar: 0.\allowbreak{}0}, the only shipped config naming either \\
PianoMime~\cite{pianomime_2024} & high & \texttt{sNiper-Qian/}\allowbreak{}\texttt{pianomime} & \href{https://github.com/sNiper-Qian/pianomime/blob/c4abefac8d2941f5a08ca656b12f5e047cab7155/single_task/piano_with_shadow_hands_res.py}{\texttt{piano\_\allowbreak{}with\_\allowbreak{}shadow\_\allowbreak{}hands\_\allowbreak{}res.\allowbreak{}py}}\newline \texttt{\_\allowbreak{}set\_\allowbreak{}rewards} & two weighted terms, 2/3 and 1/3, Table 3 & five summed terms; \texttt{\_\allowbreak{}compute\_\allowbreak{}energy\_\allowbreak{}reward} ends \texttt{return 0} and \texttt{\_\allowbreak{}compute\_\allowbreak{}fingering\_\allowbreak{}reward} \texttt{return 0.\allowbreak{}0} \\
\bottomrule
\end{tabular}
\end{table*}

A paper that states its reward and also ships the code that trained it can be read twice, and
the two readings sometimes disagree. In eight of the papers here they state different things.
Table~\ref{tab:codegap} sets each one out the same way: a repository, a file linked at the
revision this survey read, the value the paper prints and the value that file contains.

Most of the usual explanations for such a difference are ordinary ones. A repository is
refactored after a paper is published, cleaned up for release, or carried forward into the next
project, and the version a reader clones months later is not the version a result came from.
A public default need not be the setting an experiment used: a configuration committed for
others to run is often the general case, with the paper's run specified elsewhere or not at all.
A value can also be a typo in a table rather than a different objective, and one entry here is
exactly that. So a row in this table is a note for a reader who intends to use that code, telling
them which of two documents to believe about what it computes. It is not a statement about whether
the paper's reported results are right, and nothing here is a statement about intent or about
work that released nothing to read.

When an author has told us which document is authoritative, the answer has been the code. The
first author of PianoMime \cite{pianomime_2024} confirmed that the released implementation is
the one behind that paper's results and that its description of the reward is the imprecise side
of the pair (personal communication, C. Qian, Oct 2026). That is the most useful thing a reply
can settle, and it is the direction a reader using the code needs.

The sharpest difference in the table is PhysHOI \cite{physhoi_2023}: its paper assigns the
object's rotation error a nonzero weight on the GRAB dataset, and in the released reward function
that error is set to zero with the computation that would produce it commented out on the same
line, so no weight on it could move the reward. Its reported success criterion uses position
alone, so a run of that file would not expose the difference either. Every line of
Table~\ref{tab:codegap} can be confirmed or refuted in a browser, without this survey and without
anyone's agreement.

The authors of every paper named in the table were written to on 19 September 2026, before this
survey was posted, with the claim, its evidence and the sentences the survey would print, and
were given until 10 October 2026 to answer. One reply changed a comparison and removed it from
the count. A second, quoted above, settled one in the code's favour. The rest are recorded as
they came, and silence is recorded as silence rather than as agreement. A disputed comparison is
corrected at the address in the author block, and Appendix~\ref{app:rewards} prints every
comparison beside the note that revised it.

The practical lesson for a reader of this literature is simple: read the reward function before
the reward table, and treat a printed weight as a hypothesis about the code. Future papers can
remove the ambiguity altogether by generating the reward table from the configuration used for
training and citing the revision it came from.
\section{Bimanual dexterous manipulation}
\label{sec:bimanual}

\makeatletter
\@ifundefined{plateGraspPenetration}{\IfFileExists{figs/plates.tex}{
\newcommand{\plateGraspPenetration}{%
  \begin{figure}[t]
    \centering
    \includegraphics[width=0.86\columnwidth]{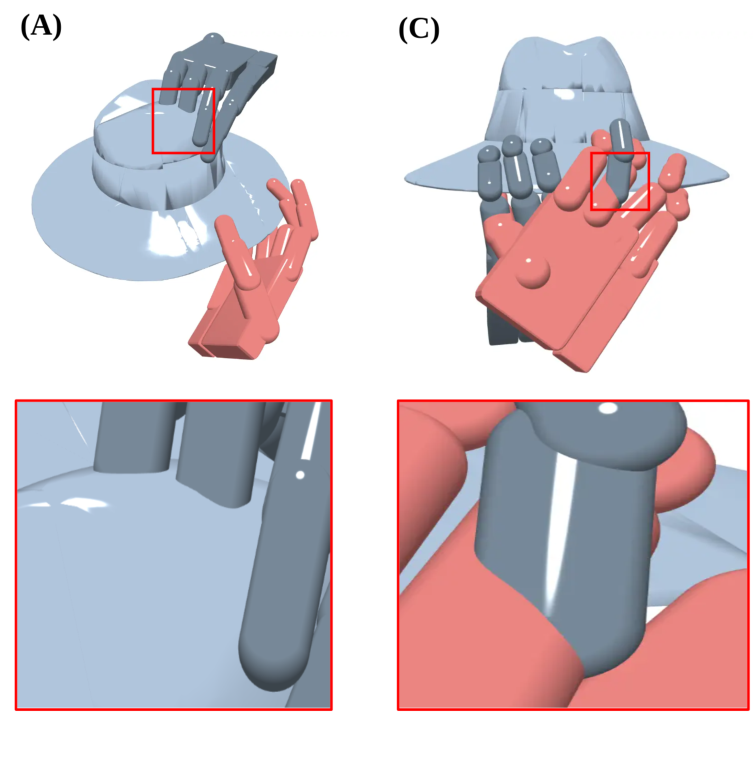}
    \caption{Two failures of synthesised two-hand grasps, each enlarged at the
      contact: (A) a finger inside the object, (C) a finger inside the other hand.
      Both look plausible whole~\cite{bimangrasp_2024}.}
    \label{fig:bimanual-penetration}
    \licensed[panels (A) and (C) of Fig.~9]{Fig.~9 of Shao and Xiao, arXiv:2411.15903}{CC BY 4.0}{https://creativecommons.org/licenses/by/4.0/}
  \end{figure}}
}{}}{}
\makeatother

Two hands are not two instances of the same problem. They share one object, they can collide
with each other, and the action space doubles, so neither the hardware result nor the policy
carries over from one hand. Counting the work is the first difficulty, because the widest count,
the 53 papers whose robot has two end effectors, is not a count of bimanual dexterous
manipulation at all. Thirteen of those put no dexterous hand on the robot: five are parallel-jaw
throughout \cite{aloha_act_2023,rdt1b_2024,egomimic_2024,h_rdt_2025,umi_2024}, four name no hand
and run grippers on every embodiment
\cite{pi0_2024,pi05_2025,pistar06_2025,diffusion_policy_2023}, two report every bimanual number
on grippers and show five-fingered hands only qualitatively
\cite{gemini_robotics_2025,gemini_robotics_15_2025}, and two name no end effector anywhere
\cite{helix_2025,groot_n16_2025}. Three more run a gripper on one embodiment and a hand on
another \cite{ace_teleop_2024,dp3_2024,open_television_2024}. Appendix~\ref{app:method} lists
every exclusion, so the denominator can be audited.

The denominator for this section is 28: the papers whose notes place a learned closed-loop
controller on two multi-fingered hands. It is the 53 less those 16; less two static
grasp-synthesis methods, BimanGrasp \cite{bimangrasp_2024} and BiDexGrasp
\cite{bidexgrasp_2026}; less three systems with no learned policy, Castro et al. and DexTeleop-0
\cite{castro_sap_contact_2021,dexteleop0_2026} and Pang et al. \cite{pang_global_planning_2022};
and less four papers where no learned policy holds both hands: OmniH2O \cite{omnih2o_2024},
whose fingers are mapped open-loop from the Vision Pro and sit outside its 19-DoF policy; OKAMI
\cite{okami_2024}, whose headline pipeline is open-loop retargeting with a learned policy only
in a side experiment; DexDeform \cite{dexdeform_2023}, a skill model refined by trajectory
optimization and not a closed-loop controller; and Omnigrasp \cite{omnigrasp_2024}, a simulated
human body with no bimanual task. Frau-Alfaro et al. \cite{dexterous_handover_2025} never enter,
because their entry records one hand. Benchmarks and datasets are outside the 28 by class,
Bi-DexHands and Bench2Dex \cite{bidexhands_2022,bench2dex_2026} and RoboPianist
\cite{robopianist_2023} being benchmarks and RP1M \cite{rp1m_2024} and HumanoidGen
\cite{humanoidgen_2025} datasets; they are quoted here as evidence and never counted. Every
paper below runs two multi-fingered hands unless said otherwise.

\subsection{Coupling, roles and collision}
\label{subsec:two_not_twice}

When both hands hold the same object, the object closes a kinematic loop between them. Neither
hand can move without changing what the other must do. BiDexGrasp \cite{bidexgrasp_2026} reports
that the coupled objective ``often yields imbalanced solutions, where one hand dominates stability
while the other contributes marginally.'' ArtiGrasp \cite{artigrasp_2023} reports simulation speed
scaling ``roughly quadratically with the number of contacts,'' and trains each hand alone before
pairing them.

Role asymmetry is an assumption most of these systems make silently. BiDexHD \cite{bidexhd_2024}
states it outright: ``we assume the robot to be right-handed by default, i.e., the left hand
handles the target object and the right hand handles the tool,'' and Lin et al.
\cite{twisting_lids_2024} bake the same split into their reward, putting reference contact
keypoints on the bottle base for the left fingertips and on the lid for the right. The bias is
in the data first: TACO \cite{taco_2024} recruited 14 right-handed subjects and measures the
right hand moving consistently faster.

The arms collide, and the corpus handles this by construction, not by control. RoboPianist
\cite{robopianist_2023} ships a forearm-forearm collision term that its own reward table never
lists. Two papers name arm collision as unsolved. DexMimicGen \cite{dexmimicgen_2024} ``does not
explicitly handle collisions,'' and DexMan \cite{dexman_2025} says its control parameterisation
``overlooks full arm posture,'' which it needs to avoid them.

The action space doubles and the observation grows faster: a two-hand benchmark carries action
vectors in the tens and observations in the hundreds of dimensions \cite{bidexhands_2022}, and a
recent generalist emits a chunk of arm joints, end-effector poses, hand joints and fingertips
together \cite{gr_dexter_2025}. What the larger vector does not carry is any statement of who does
what. The usual answer is one scalar reward summed over both sides, and the cost is a compounding
failure rate: where success requires both hands to succeed, ManipTrans
\cite{maniptrans_2025}'s rate falls by about a third from single-hand sequences to bimanual
ones.

\subsection{Coordination architectures}
\label{subsec:coordination}

\begin{figure*}[!t]
\centering
\IfFileExists{figs/fig_bimanual.tex}{\begin{tikzpicture}[
  font=\footnotesize,
  panel/.style={draw=rule,line width=0.4pt,inner sep=0pt},
  panelacc/.style={draw=accent,line width=0.8pt,inner sep=0pt},
  box/.style={draw=black!45,fill=wash,line width=0.3pt,inner sep=0pt,align=center},
  hand/.style={draw=black!55,fill=wash,line width=0.3pt,circle,inner sep=0pt,minimum size=5.0mm,
               font=\tiny},
  handbig/.style={hand,draw=accent,line width=0.5pt,minimum size=6.6mm},
  lnk/.style={draw=black!50,line width=0.35pt,-latex},
  soft/.style={draw=black!45,line width=0.35pt,dash pattern=on 1.4pt off 1.4pt},
  hd/.style={font=\footnotesize\bfseries,align=center,anchor=north},
  cnt/.style={font=\scriptsize\bfseries\color{accent},anchor=north},
  dsc/.style={font=\tiny\color{black!60},align=center,anchor=north},
  ky/.style={font=\tiny\ttfamily\color{black!75},align=center,anchor=north},
]

\begin{scope}[shift={(0,0)}]
  \node[panelacc,minimum width=42mm,minimum height=40mm] at (2.1,2.00) {};
  \node[hd] at (2.1,3.86) {one policy, both hands};
  \node[cnt] at (2.1,3.56) {21 of 28};
  \node[dsc] at (2.1,3.32) {observation of both hands and object,\\one network, joint action vector};
  \node[box,minimum width=13mm,minimum height=4.0mm,font=\tiny] (p1) at (2.1,2.50) {policy};
  \node[hand] (l1) at (1.15,1.95) {L};
  \node[hand] (r1) at (3.05,1.95) {R};
  \draw[lnk] (p1.south west) -- (l1.north east);
  \draw[lnk] (p1.south east) -- (r1.north west);
  \node[ky] at (2.1,1.50) {twisting\_lids\_2024\\dexmachina\_2025\\maniptrans\_2025\\gr\_dexter\_2025\\groot\_n1\_2025\\and 16 more};
\end{scope}

\begin{scope}[shift={(4.38,0)}]
  \node[panel,minimum width=42mm,minimum height=40mm] at (2.1,2.00) {};
  \node[hd] at (2.1,3.86) {a network per hand};
  \node[cnt] at (2.1,3.56) {4 of 28};
  \node[dsc] at (2.1,3.32) {each hand its own network, a centralised\\critic or its own observation only};
  \node[box,minimum width=11mm,minimum height=4.0mm,font=\tiny] (a2) at (1.15,2.50) {$\pi_{\mathrm{L}}$};
  \node[box,minimum width=11mm,minimum height=4.0mm,font=\tiny] (b2) at (3.05,2.50) {$\pi_{\mathrm{R}}$};
  \node[hand] (l2) at (1.15,1.95) {L};
  \node[hand] (r2) at (3.05,1.95) {R};
  \draw[soft] (a2.east) -- (b2.west);
  \draw[lnk] (a2.south) -- (l2.north);
  \draw[lnk] (b2.south) -- (r2.north);
  \node[ky] at (2.1,1.50) {bidexhd\_2024\\artigrasp\_2023\\dynamic\_handover\_2023\\dydexhandover\_2025};
\end{scope}

\begin{scope}[shift={(8.76,0)}]
  \node[panel,minimum width=42mm,minimum height=40mm] at (2.1,2.00) {};
  \node[hd] at (2.1,3.86) {leader and follower};
  \node[cnt] at (2.1,3.56) {1 of 28};
  \node[dsc] at (2.1,3.32) {one hand assigned the dominant role,\\the other reacts to it};
  \node[handbig] (l3) at (1.15,2.25) {L};
  \node[hand] (r3) at (3.05,2.25) {R};
  \draw[lnk] (l3.east) -- node[above,font=\tiny\color{black!60},inner sep=1.5pt] {reacts} (r3.west);
  \node[ky] at (2.1,1.85) {asymdex\_2024};
  \node[dsc] at (2.1,1.50) {the giver in\\dexterous\_handover\_2025 is a\\scripted arm, so that row is\\not counted};
\end{scope}

\begin{scope}[shift={(13.14,0)}]
  \node[panel,minimum width=42mm,minimum height=40mm] at (2.1,2.00) {};
  \node[hd] at (2.1,3.86) {relative frame};
  \node[cnt] at (2.1,3.56) {1 of 28};
  \node[dsc] at (2.1,3.32) {action or observation expressed between\\the hands, or in the held object's frame};
  \node[hand] (l4) at (1.15,2.25) {L};
  \node[hand] (r4) at (3.05,2.25) {R};
  \draw[soft,latex-latex] (l4.east) -- node[above,font=\tiny\color{black!60},inner sep=1.5pt] {$\Delta$} (r4.west);
  \node[ky] at (2.1,1.85) {asymdex\_2024};
  \node[dsc] at (2.1,1.50) {dexmimicgen\_2024 preserves it\\offline when generating data,\\not in the policy's\\observation};
\end{scope}

\draw[draw=black!38,line width=0.28pt] (0,-0.12) -- (17.34,-0.12);
\end{tikzpicture}}{%
  \framebox[0.7\textwidth]{\rule{0pt}{2.2cm}\textit{figs/fig\_bimanual.tex pending}}}
\caption{Four mechanisms for controlling two dexterous hands, counted over the 28 papers in this corpus with
a learned closed-loop controller on two multi-fingered hands, each read from its note. The panels
are not a partition: the fourth is the same single paper as the third, and the two papers whose notes
never say which architecture they use appear in neither.}
\label{fig:bimanual}
\end{figure*}

\figurename~\ref{fig:bimanual} sets the four mechanisms side by side. Every one of the 28 carries
exactly one architecture, read from its note: one policy over both hands, a network per hand,
leader and follower, or a note that does not say. The fourth panel is not a fourth architecture but
the relative frame, and it is the same single paper as the third.

Twenty-one of the 28 put one policy over both hands, which is three quarters of the set: the
trackers and tracking-adjacent methods
\cite{dexmachina_2025,dexman_2025,maniptrans_2025,objdex_2024} and \cite{humanplus_2024}; the
reinforcement-learning tasks \cite{dexpbt_2023,twisting_lids_2024,eureka_2023,pianomime_2024} and
\cite{humanoid_sim2real_recipe_2025}; the demonstration pipelines
\cite{bidex_teleop_2024,dexcap_2024,dexwild_2025,dexmimicgen_2024,hato_visuotactile_2024} and
\cite{humanoid_policy_human_policy_2025}; and the generalist policies
\cite{gr_dexter_2025,groot_n1_2025,dexora_2026,metis_2025} and \cite{egoscale_2026}. In every one
of them a single network takes a concatenated two-hand observation and emits a two-hand action.
Two papers do not say which they are, Bunny-VisionPro \cite{bunny_visionpro_2024} and DexImit
\cite{deximit_2026}, whose notes describe the rig and the data pipeline but never the policy's own
decomposition. The concentration is not the outcome of a comparison that was won.

Four of the 28 give each hand its own network, and the two papers that compare the choice
disagree. Bi-DexHands \cite{bidexhands_2022}, a benchmark entry and so outside the 28, ships the
MARL baselines and finds PPO over the full observation beats HAPPO and MAPPO ``in most cases,''
with the gap narrowing on tasks that need both hands, because PPO ``can use all observations''
where MARL sees only part. BiDexHD \cite{bidexhd_2024} concludes the opposite. Its independent PPO
teachers score 74.59 percent stage-two tracking rate on trained tasks against 53.88 for a
centralised policy over both observations. Both are simulation only, on different tasks and hands,
so neither settles it. ArtiGrasp \cite{artigrasp_2023} trains one PPO policy per hand, and Dynamic
Handover \cite{dynamic_handover_2023} and DyDexHandover \cite{dydexhandover_2025} use MAPPO with
one agent per arm-hand system.

One paper assigns explicit leader and follower roles, and it is the same one that expresses its
policy in a relative frame. AsymDex \cite{asymdex_2024} gives the dominant hand full finger and
wrist control, restricts the facilitating hand to a base pose, and writes both the dominant hand
and the object in a frame attached to the object the other hand holds, which halves the
observation and the action. It is also the only paper here that ablates the two mechanisms
separately, and neither works without the other: relative frames without asymmetry and asymmetry
without relative frames both collapse to near zero where the full method succeeds most of the
time. Its real rig pairs an Allegro with a cheaper hand because only one Allegro was available,
so the roles are confounded with the hardware. Two other papers use a relative quantity without
giving either hand a role, one appending the inter-hand pose to its observation
\cite{dexwild_2025} and one applying a shared transform to both arms while generating data
\cite{dexmimicgen_2024}, which is the mechanism offline and not in a policy.

One architecture carries three quarters of the field, and the choice behind it has been compared
twice, with opposite outcomes, and ablated once. A decisive comparison would run one handover
task on three architectures with the same hand and the same seeds, reporting separate giver and
receiver returns.

\subsection{Benchmarks, datasets and the shared axes}
\label{subsec:bimanual_benchmarks}

Two purpose-built bimanual dexterous suites exist in the corpus, four years apart.

The two bimanual benchmarks in the corpus trade coverage against measurement, in opposite
directions. Bi-DexHands \cite{bidexhands_2022} is 20 tasks on two Shadow Hands in Isaac Gym,
ordered by the infant age at which humans acquire the skill, at 2048 environments and a
reported 30,000-plus FPS. Its measurement discipline is weaker than its coverage. It reports
reward and normalized score, never a success rate. Its only success flag in code tests the
object-to-goal distance against 3 cm, which ignores orientation and exists only in the four
catching tasks. Any success rate later work attributes to Bi-DexHands comes from that flag or
its own definition.

Bench2Dex \cite{bench2dex_2026} is the more instrumented of the two, running 26 long-horizon
tasks across 12 arm-and-hand embodiments with thousands of teleoperated demonstrations in eight
modalities, including a ray-cast tactile image that maps every hand geometry into one
representation. Its headline result is the gap between conditions and not the leaderboard: the
matched leader takes about half the rollouts and falls to a fifth under combined perturbation,
which compresses the field until the ranking no longer holds. Tasks are not crossed with
embodiments, so no contrast isolates the hand, and the authors state that their hand and tactile
models are not calibrated against matching physical hardware.

Eleven papers in this corpus are bimanual datasets, five of them two-hand interaction capture,
and their annotations are where the contact problem first appears. ARCTIC \cite{arctic_2022}
labels over-shooting vertices as in contact by design, so its ground truth records
interpenetration as contact, and \figurename~\ref{fig:contactlabel} is the chain that produces
it. TACO \cite{taco_2024} fits its sequences under an attraction loss and a penetration loss
that the paper says trade against each other. OakInk-V2 \cite{oakink2_2024} is the only one
publishing a penetration figure for its own annotations, at a mean depth of a few millimetres.
The two largest egocentric sets \cite{hot3d_2024,gigahands_2024} publish none at all. Those five
feed most of the physics-based tracking work in this section
\cite{dexmachina_2025,artigrasp_2023,bidexhd_2024,maniptrans_2025,dexman_2025}, so an
interpenetrating reference is where several of these pipelines start.

One of the eleven is not human capture at all. SharpaDex \cite{sharpadex_2026} is a robot
teleoperation release, 59 tasks and about 300 hours of two-handed demonstration, and it is the
largest such collection here by an order of magnitude: the next largest teleoperated set in this
corpus is tens of hours, and the bigger figures elsewhere are egocentric human video. It is also
the only dataset here that ships a tactile stream beside the video, which is why it is here.
Section~\ref{sec:hands} records that tactile hardware has been cheap for years and barely
appears in the experiments; a teleoperated collection at this scale with tactile in it is the
first thing that could change that. What it does not carry is a result. Its card reports no
baseline policy and no success criterion, so nothing in Section~\ref{sec:evaluation} applies to
it, and its licence does not permit commercial use, which is the distinction this survey keeps
wherever it calls a dataset public.

\begin{figure}[!t]
  \centering
  \IfFileExists{figs/fig_contactlabel.tex}{\resizebox{\columnwidth}{!}{\begin{tikzpicture}[
  font=\footnotesize,
  stage/.style={draw=black!45,line width=0.3pt,inner sep=3.5pt,align=center,
                font=\scriptsize,text width=21mm,minimum height=9mm},
  lnk/.style={draw=black!45,line width=0.35pt,-latex},
  det/.style={font=\scriptsize\color{black!60},anchor=north,align=center,text width=24mm},
  surf/.style={draw=black!60,line width=0.5pt},
  fing/.style={draw=accent,line width=0.6pt,fill=white},
  over/.style={draw=none,fill=accent,fill opacity=0.20},
  call/.style={draw=black!45,line width=0.28pt,dash pattern=on 1.2pt off 1.2pt},
  note/.style={font=\scriptsize,anchor=north west,align=left},
]

\node[stage] (s1) at (1.20,0) {multi-view video\\and mocap};
\node[stage] (s2) at (4.05,0) {hand and object\\meshes fitted to it};
\node[stage] (s3) at (6.90,0) {contact read off\\the fitted meshes};
\draw[lnk] (s1.east) -- (s2.west);
\draw[lnk] (s2.east) -- (s3.west);
\node[det,anchor=north] at (1.20,-0.62) {measured};
\node[det,anchor=north] at (4.05,-0.62) {estimated, with a\\residual the dataset\\reports};
\node[det,anchor=north] at (6.90,-0.62) {\textcolor{accent}{inferred from the
                                          estimate, never measured}};

\begin{scope}[shift={(0.55,-3.05)}]
  \fill[black!8] (0,0) rectangle (2.30,-0.52);
  \draw[surf] (0,0) -- (2.30,0);
  \draw[fing] (0.92,-0.24) -- (1.52,-0.24) -- (1.52,0.76) -- (0.92,0.76) -- cycle;
  \fill[over] (0.92,0) rectangle (1.52,-0.24);
  \draw[accent,line width=0.6pt] (0.92,0) -- (0.92,-0.24) -- (1.52,-0.24) -- (1.52,0);
  \draw[surf] (0,0) -- (0.92,0);
  \draw[surf] (1.52,0) -- (2.30,0);
  \node[font=\scriptsize\color{black!70},anchor=west] at (0.06,-0.36) {object};
  \node[font=\scriptsize\color{accent},anchor=west] at (1.62,0.30) {finger};
  \draw[call] (1.52,-0.24) -- (2.05,-0.62);
  \node[font=\scriptsize\color{black!65},anchor=west] at (2.05,-0.76)
       {vertex past the surface};
  \node[font=\scriptsize\color{black!65},anchor=west] at (2.05,-1.14)
       {label written: \textcolor{accent}{\textbf{in contact}}};
\end{scope}

\draw[draw=black!25,line width=0.28pt] (0,-4.80) -- (8.20,-4.80);
\node[note,text width=81mm] at (0,-4.96)
  {The annotation is as good as the fit and no better, and a penetration is recorded as a
   contact rather than as an error. Of the five bimanual capture sets in Sec.~VI, one publishes a
   penetration number for its own annotations.};
\end{tikzpicture}}}{%
    \framebox[0.95\columnwidth]{\rule{0pt}{5.0cm}\textit{figs/fig\_contactlabel.tex pending}}}
  \caption{Where a capture set's contact annotation comes from. Nothing in the chain is measured
  at a surface, and a vertex that has gone through the surface is written down as a contact rather
  than as an error.}
  \label{fig:contactlabel}
\end{figure}

\label{subsec:bimanual_axes}
The method tabulation carries the bimanual papers on the same columns as every other method here,
and three of those columns are worth reading together.

Real trials are few and are not counted the same way. Where a count is given it runs from ten
per condition to a few hundred across an object set
\cite{asymdex_2024,hato_visuotactile_2024,dexmimicgen_2024,bidexgrasp_2026}, and one paper
reaches real hands by open-loop replay with no trial count and no success rate at all
\cite{maniptrans_2025}. Seven of the 28 have no real robot
\cite{artigrasp_2023,bidexhd_2024,dexmachina_2025,dexman_2025,dexpbt_2023,dydexhandover_2025,pianomime_2024},
and the two piano papers outside the denominator are simulation-only as well
\cite{robopianist_2023,rp1m_2024}.

\providecommand{\plateGraspPenetration}{}\plateGraspPenetration

Penetration, and \figurename~\ref{fig:bimanual-penetration} is why it has to be reported beside
a success rate and not instead of one. Most bimanual papers read ``not addressed.'' BimanGrasp
\cite{bimangrasp_2024} gates on it, failing any grasp whose total penetration across
hand-object, self and inter-hand exceeds 1.5 mm, and reports that ``penetration remains the
primary cause of grasp failure.'' BiDexGrasp \cite{bidexgrasp_2026} reports it as a number, 0.15
to 0.20 cm maximum depth for its own grasps. DexImit \cite{deximit_2026} carries a named
hand-hand penetration term in its grasp-synthesis objective. DexMachina \cite{dexmachina_2025}
resolves it once, replaying retargeted joints against a fixed object ``to eliminate object
penetrations'' and measuring nothing during rollout. ArtiGrasp \cite{artigrasp_2023} declines
the metric outright, because its baselines ``include a physics simulation which exhibits no
interpenetration.''

Comparability fails even on identical grasps. BimanGrasp \cite{bimangrasp_2024} reports 54
percent success in Isaac Gym; BiDexGrasp \cite{bidexgrasp_2026} re-runs the same grasps in MuJoCo
at a lower friction coefficient and gets half that, with more than a centimetre of penetration
depth. Neither number transfers, and Section~\ref{sec:train:track}'s re-implementation result is
the same lesson on the training side.

\subsection{Handover, and what transfers from one hand}
\label{subsec:handover}
Handover is the one bimanual task where the hands are unambiguously asymmetric, because one gives
and one receives. All three corpus handover papers use a single shared reward across giver and
receiver. None defines separate objectives for the two roles.

Whether the two hands optimize the same objective is usually left implicit. Dynamic Handover
\cite{dynamic_handover_2023} states one formula, $r = r_{\mathrm{dis}} + r_{\mathrm{linvel}} +
r_{\mathrm{torque}}$, and never says whether thrower and catcher receive different
decompositions of it. DyDexHandover \cite{dydexhandover_2025} is explicit that they do not:
``Both hands share aligned objectives, forming a fully cooperative relationship,'' with one
weighted sum feeding both agents. Only the thrower gets an extra KL regulariser toward a policy
pretrained on human throws, which constrains style and not role.

The third case is weaker still. In \cite{dexterous_handover_2025} only the receiver is learned.
The giver is a UR5e with an Allegro hand that ``holds the object without moving during the whole
episode.'' There is one agent, one reward and no second policy, its entry accordingly records
one hand, and it is one of the papers the 28 excludes. The 94 percent often attached to this paper
needs its conditions. It is Total Success, which counts ``Indetermination'' cases where the
simulator failed to resolve collisions and the object clipped through the giver's hand, on the
short prism, in simulation, over 100 episodes, with no real robot in the paper.

Nothing in the handover literature measures contact quality. Contact appears only as a positive
signal, a boolean per-phalange touch in \cite{dexterous_handover_2025} and a boolean contact
reward in \cite{dydexhandover_2025}. The two systems that do handover well sidestep the problem.
HATO \cite{hato_visuotactile_2024} reaches 10 of 10 on a real slippery handover by imitating
teleoperation with fingertip touch sensing and no physics model. Lin et al.
\cite{humanoid_sim2real_recipe_2025} stage their reward with a discrete variable switching which
hand's fingertips are scored, and still calls handover its hardest task at 52.5 percent real
success.

\label{subsec:bimanual_transfer}
The training recipe transfers intact. PPO with thousands of parallel environments, an asymmetric
critic on privileged state, domain randomisation and distillation into a vision policy is the same
in \cite{twisting_lids_2024} and \cite{humanoid_sim2real_recipe_2025} as in
Section~\ref{sec:train:rl}. Reward forms transfer literally, and the contact-goal reward
$r_{\mathrm{contact}} = \sum 1/(1+\alpha d)$ appears in both.

Multi-task learning does not transfer, and the bimanual benchmarks show it clearest. Bi-DexHands
\cite{bidexhands_2022} finds multi-task PPO and ProMP fail on MT4, MT20, ML4 and ML20. RoboPianist
\cite{robopianist_2023} finds F1 on the shared training song drops ``from roughly 0.7 F1 for 1
song to almost 0 F1 for 16 songs,'' with no positive transfer ``regardless of the size of the
pre-training tasks.''

Single-hand evaluation does not transfer either. Success criteria compound, because ManipTrans
\cite{maniptrans_2025} fails a trajectory ``if either hand fails to meet these conditions.'' And
object-centric metrics cannot say which hand caused the failure. BiDexHD \cite{bidexhd_2024}
scores the steps where both objects track their references, and DexMachina \cite{dexmachina_2025}
averages ADD across object parts before an AUC. Both say whether the task worked and nothing about
which hand lost it.

One failure mode has no single-hand counterpart. Two hands can penetrate each other, and the
corpus almost never looks. Three papers carry an inter-hand penetration term, and all three are
static grasp synthesis: \cite{bimangrasp_2024,bidexgrasp_2026} and \cite{deximit_2026}. Not one
learned bimanual controller in the corpus measures or penalizes hand-hand penetration during a
rollout. PianoMime \cite{pianomime_2024} ships a flag that turns inter-hand collision off in the
physics.

\section{Evaluation, and a protocol for comparison}
\label{sec:evaluation}

\makeatletter
\@ifundefined{fnLoaded}{\IfFileExists{figs/finding_numbers.tex}{\providecommand{\fnLoaded}{}
\providecommand{\fnRows}{113}
\providecommand{\fnCode}{63}
\providecommand{\fnDisagree}{39}
\providecommand{\fnKept}{8}
\providecommand{\fnOtherClass}{31}
\providecommand{\fnParse}{14}
\providecommand{\fnAbsent}{8}
\providecommand{\fnSkew}{5}
\providecommand{\fnInternal}{4}
\providecommand{\fnPenSettled}{97}
\providecommand{\fnPenHandled}{12}
\providecommand{\fnPenClosed}{5}
\providecommand{\fnPenOther}{7}
\providecommand{\fnPenRollout}{0}
\providecommand{\fnTrials}{39}
\providecommand{\fnTrialsMedian}{15}
\providecommand{\fnTrialsMode}{10}
\providecommand{\fnTrialsModeN}{16}
\providecommand{\fnTotals}{24}
\providecommand{\fnTotalsMin}{12}
\providecommand{\fnTotalsMax}{750}
\providecommand{\fnUnclear}{7}
}{}}{}
\makeatother

\subsection{Reporting practice}
\label{subsec:reporting}

\begin{figure}[!t]
\centering
\IfFileExists{figs/fig_reporting.tex}{\resizebox{\columnwidth}{!}{\begin{tikzpicture}[
  font=\footnotesize,
  bar/.style={draw=none,fill=black!62},
  barmid/.style={draw=none,fill=black!42},
  barlight/.style={draw=none,fill=black!20},
  baracc/.style={draw=none,fill=accent},
  seg/.style={draw=white,line width=0.5pt},
  lnk/.style={draw=black!38,line width=0.28pt},
  ghost/.style={draw=black!45,line width=0.3pt,dash pattern=on 1.2pt off 1.2pt},
  lbl/.style={font=\scriptsize},
  small/.style={font=\scriptsize\color{black!55}},
  num/.style={font=\scriptsize\color{black!55}},
]

\node[lbl,anchor=east] at (0,-0.00) {success criterion stated};
\fill[barlight] (0.12,-0.10) rectangle (5.32,0.10);
\fill[bar] (0.12,-0.10) rectangle (4.68,0.10);
\node[num,anchor=west] at (5.44,-0.00) {99 of 113 (88\%)};
\node[lbl,anchor=east] at (0,-0.44) {real-robot experiment};
\fill[barlight] (0.12,-0.54) rectangle (5.27,-0.34);
\fill[black!8] (5.27,-0.54) rectangle (5.32,-0.34);
\fill[bar] (0.12,-0.54) rectangle (4.22,-0.34);
\node[num,anchor=west] at (5.44,-0.44) {89 of 112 (79\%)};
\node[lbl,anchor=east] at (0,-0.88) {real trial count stated};
\fill[barlight] (0.12,-0.98) rectangle (4.22,-0.78);
\fill[bar] (0.12,-0.98) rectangle (3.34,-0.78);
\node[num,anchor=west] at (5.44,-0.88) {70 of 89 (79\%)};
\node[lbl,anchor=east] at (0,-1.32) {code released};
\fill[barlight] (0.12,-1.42) rectangle (5.14,-1.22);
\fill[black!8] (5.14,-1.42) rectangle (5.32,-1.22);
\fill[bar] (0.12,-1.42) rectangle (3.02,-1.22);
\node[num,anchor=west] at (5.44,-1.32) {63 of 109 (58\%)};
\node[lbl,text=accent,anchor=east] at (0,-1.76) {unseen-object count stated};
\fill[barlight] (0.12,-1.86) rectangle (5.32,-1.66);
\fill[baracc] (0.12,-1.86) rectangle (1.96,-1.66);
\node[font=\scriptsize\color{accent},anchor=west] at (5.44,-1.76) {40 of 113 (35\%)};
\node[lbl,text=accent,anchor=east] at (0,-2.20) {contact or penetration handled};
\fill[barlight] (0.12,-2.30) rectangle (4.58,-2.10);
\fill[black!8] (4.58,-2.30) rectangle (5.32,-2.10);
\fill[baracc] (0.12,-2.30) rectangle (0.67,-2.10);
\node[font=\scriptsize\color{accent},anchor=west] at (5.44,-2.20) {12 of 97 (12\%)};
\draw[line width=0.4pt,draw=accent] (7.40,-1.60) -- (7.50,-1.60) -- (7.50,-2.36) -- (7.40,-2.36);
\draw[lnk] (0.12,-2.48) -- (5.32,-2.48);
\node[anchor=north west,align=left,text width=5.2cm,font=\scriptsize\color{accent}] at (0.12,-2.60) {the two rows in this colour are the pair a reader needs in order to compare any two methods};
\end{tikzpicture}}}{%
  \framebox[0.95\columnwidth]{\rule{0pt}{2.2cm}\textit{figs/fig\_reporting.tex pending}}}
\caption{What gets reported, and what does not. Each bar is a count of the 113 corpus method
papers; the pale track behind it is the denominator that statistic is quoted against, printed beside
the count, because four of the six are not 113. A faint tail past a track is the papers whose notes
left that question unsettled. The two rarest reporting practices are marked in color.}
\label{fig:reporting}
\end{figure}
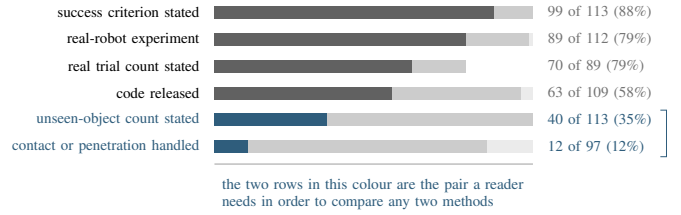

Most of this literature does test on hardware, which is the check that matters and the one a
simulator cannot substitute for: 89 of the 113 papers report a real-robot experiment. What they
report about those experiments is much less even. \figurename~\ref{fig:reporting} draws six
reporting axes, each against the denominator that belongs to it, and two of the six are the ones
a reader most needs. Without a count of unseen test objects there is no denominator for a claim
about generalisation; without some statement about contact there is no way to tell whether the
hand passed through the object. Those two are the axes the field states least often. Every share
in the figure is a lower bound, for reasons Appendix~\ref{app:method} gives.

The plainest gap is the trial count. Nineteen of the papers that ran a real robot never say how
many times they ran it, and a percentage with no denominator cannot be given an interval, so it
cannot be compared with anything.

\begin{figure}[!t]
  \centering
  \IfFileExists{figs/fig_trials.tex}{\resizebox{\columnwidth}{!}{\begin{tikzpicture}[
  font=\footnotesize,
  bar/.style={draw=none,fill=black!62},
  barmid/.style={draw=none,fill=black!42},
  barlight/.style={draw=none,fill=black!20},
  baracc/.style={draw=none,fill=accent},
  seg/.style={draw=white,line width=0.5pt},
  lnk/.style={draw=black!38,line width=0.28pt},
  ghost/.style={draw=black!45,line width=0.3pt,dash pattern=on 1.2pt off 1.2pt},
  lbl/.style={font=\scriptsize},
  small/.style={font=\scriptsize\color{black!55}},
  num/.style={font=\scriptsize\color{black!55}},
]

\fill[bar] (0.00,0) rectangle (0.56,0.13);
\node[num,anchor=south] at (0.28,0.15) {1};
\draw[lnk] (0.56,0) -- (0.56,-0.06);
\node[lbl,anchor=north] at (0.56,-0.08) {5};
\fill[bar] (0.70,0) rectangle (1.26,2.10);
\node[num,anchor=south] at (0.98,2.12) {16};
\draw[lnk] (1.26,0) -- (1.26,-0.06);
\node[lbl,anchor=north] at (1.26,-0.08) {10};
\fill[bar] (1.40,0) rectangle (1.96,0.53);
\node[num,anchor=south] at (1.68,0.55) {4};
\draw[lnk] (1.96,0) -- (1.96,-0.06);
\node[lbl,anchor=north] at (1.96,-0.08) {15};
\fill[bar] (2.10,0) rectangle (2.66,1.58);
\node[num,anchor=south] at (2.38,1.60) {12};
\draw[lnk] (2.66,0) -- (2.66,-0.06);
\node[lbl,anchor=north] at (2.66,-0.08) {20};
\fill[bar] (2.80,0) rectangle (3.36,0.26);
\node[num,anchor=south] at (3.08,0.28) {2};
\draw[lnk] (3.36,0) -- (3.36,-0.06);
\node[lbl,anchor=north] at (3.36,-0.08) {25};
\draw[lnk] (4.06,0) -- (4.06,-0.06);
\node[lbl,anchor=north] at (4.06,-0.08) {30};
\fill[bar] (4.20,0) rectangle (4.76,0.13);
\node[num,anchor=south] at (4.48,0.15) {1};
\draw[lnk] (4.76,0) -- (4.76,-0.06);
\node[lbl,anchor=north] at (4.76,-0.08) {35};
\fill[bar] (4.90,0) rectangle (5.46,0.13);
\node[num,anchor=south] at (5.18,0.15) {1};
\draw[lnk] (5.46,0) -- (5.46,-0.06);
\node[lbl,anchor=north] at (5.46,-0.08) {40};
\draw[lnk] (6.16,0) -- (6.16,-0.06);
\node[lbl,anchor=north] at (6.16,-0.08) {45};
\fill[bar] (6.30,0) rectangle (6.86,0.13);
\node[num,anchor=south] at (6.58,0.15) {1};
\draw[lnk] (6.86,0) -- (6.86,-0.06);
\node[lbl,anchor=north] at (6.86,-0.08) {50};
\fill[bar] (7.34,0) rectangle (7.90,0.13);
\node[num,anchor=south] at (7.62,0.15) {1};
\draw[lnk] (7.62,0) -- (7.62,-0.06);
\node[lbl,anchor=north] at (7.62,-0.08) {$>$50};
\draw[lnk] (-0.06,0) -- (7.96,0);
\node[lbl,anchor=north] at (3.95,-0.42) {real trials per cell, in bins of five};
\draw[draw=accent,line width=0.4pt,dash pattern=on 1.4pt off 1.2pt] (1.96,-0.02) -- (1.96,2.44);
\node[anchor=west,font=\scriptsize\color{accent}] at (2.04,2.34) {median 15};
\node[small,anchor=north west,align=left] at (-0.06,-0.66) {39 rows state a per-cell count; the mode is 10, in 16 of them\\24 state a grand total instead, 12 to 750, and are not pooled into this\\7 more state a count on a basis that could not be read};
\end{tikzpicture}}}{%
    \framebox[0.95\columnwidth]{\rule{0pt}{2.2cm}\textit{figs/fig\_trials.tex pending}}}
  \caption{Where a denominator is stated it is small: the \fnTrials{} per-cell real-trial counts,
  binned by fives, median \fnTrialsMedian{} marked. The \fnTotals{} grand totals are a separate
  quantity, counted beneath the axis and not pooled in.}
  \label{fig:trials}
\end{figure}
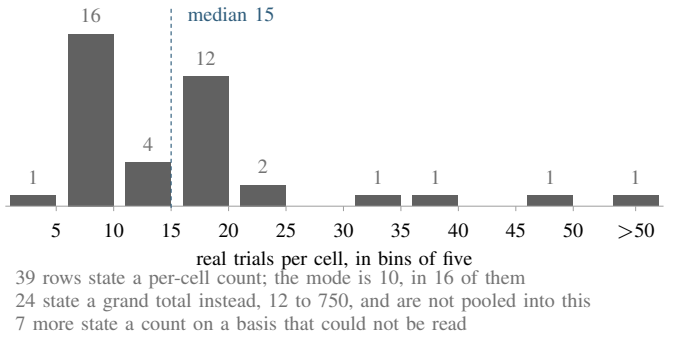

Where a count is stated it is small, and it is not always the same quantity. Some papers report a
per-cell count, a fixed number of trials on each task or each condition; others report a grand
total over every cell; a few cannot be assigned to either basis. The two distributions are kept
apart here and in \figurename~\ref{fig:trials}, because pooling them produces a figure that
describes nothing: Hora \cite{hora_2022}'s 240 trials and Visual Dexterity
\cite{visual_dexterity_2022}'s 20 are experiments of comparable size recorded on different bases.

The per-cell count is the one a confidence interval attaches to, and at the sizes this field
uses the interval is wider than the differences being claimed. Writing $\hat{p} = s/n$ for $s$
successes in $n$ trials and $z = z_{1-\alpha/2}$ for the normal quantile at the chosen level,
the Wilson bounds are \begin{equation} \frac{2n\hat{p} + z^{2} \;\pm\;
z\sqrt{4n\hat{p}\,(1-\hat{p}) + z^{2}}}{2\,(n + z^{2})}, \label{eq:wilson} \end{equation} which,
unlike the normal approximation, stays inside $[0,1]$ and does not collapse to zero width at
$\hat{p} = 0$ or $\hat{p} = 1$. At twenty trials per cell, two methods twenty points apart in
success rate are not statistically distinguishable: the two-proportion test on that comparison
returns nothing, and Appendix~\ref{app:protocol} gives the intervals and the test. The modal
cell in this corpus is smaller still. This is the single most consequential fact about
evaluation in dexterous manipulation, and it is why Section~\ref{subsec:protocol} proposes a
protocol and not a league table.

Reporting has improved a little and the experiments have not. About two thirds of papers from
2025 onward state a trial count, against three fifths before, but the counts themselves have
not moved: the median has sat near twenty trials per cell every year since 2024, and earlier
years rest on too few observations to read as a trend. A field can get better at disclosing a
number without the number itself getting any better.

The denominators also sit on different hardware. The papers that name their own hand describe it
in 79 different ways, unnormalised, so that figure counts strings and not designs;
\figurename~\ref{fig:hands} gives the distribution. A success rate on an Allegro and a success
rate on an Inspire hand are not measurements of the same thing, because the two differ in actuated
joints by nearly a factor of three.

Nor are the criteria the same. DexVerse \cite{dexverse_2026} counts a pick a success when the
cube is ``lifted at least 0.20 m above its resetting height''. Bench2Dex \cite{bench2dex_2026}
requires its terminal predicate to hold for a continuous dwell time, to reject transient contacts.
The Colosseum \cite{colosseum_2024} counts an episode successful ``if the model completes the task
fully''. DexTrack \cite{dextrack_2025} reports every success rate as a pair under two threshold
sets, which can move the same rollouts by nearly twenty points. Among the papers that state no
criterion at all, most have no success predicate to state: they report radians rotated or
time-to-fall and never define a success, which is a fact about the paper rather than a gap in this
survey. Where a criterion is stated it is usually a rubric, a staged partial credit or a deferral
to a benchmark's own definition rather than a threshold, and a rubric is reproducible inside a lab
and not across two.

\label{subsec:axes} Seven quantities dissociate in the published data, which is why they have to
be reported separately.

Task success is the first, and binary success discards the difference between a near-miss and
inaction. Snyder et al. \cite{beyond_binary_success_2026} put it plainly: ``a policy that completes
90\% of the task is clearly better than a policy that is frozen the whole time, yet their success
rates would be identically 0\%''. In \cite{kress_gazit_policy_eval_2024} one policy scores in the
teens overall on a pancake task while picking up the spatula and flipping the pancake in every
attempt.

Robustness to perturbation has been measured, though not yet on a hand. The Colosseum
\cite{colosseum_2024} reports a success drop of a third to a half under single perturbation
factors and more than three quarters under all of them together. In \cite{bench2dex_2026} the
leader of the matched condition is not the leader under combined shift. The ranking at the anchor
is not the ranking under shift.

Generalisation to unseen objects is stated by 40 papers, and half of those test on ten objects or
fewer.

Physical plausibility of the contact is the least reported of the six, at 12 percent of the papers
whose notes settle the question. Section~\ref{subsec:plausibility} takes it up.

Sample and wall-clock cost is reported by about a third of the corpus as an environment count and
by fewer as an episode count. RoboPianist \cite{robopianist_2023} is the exception that shows what
disclosure looks like, giving samples per song and hours per run on named hardware.

Real-robot transfer is the one that undoes the rest, because simulated rank order is not real rank
order. AutoEval \cite{autoeval_2025} scores one policy at 6 of 50 in simulation and 47 of 50 on
the real arm for the same task. Badithela et al. \cite{suresim_2025} state the limit directly:
``the simulation-to-real gap precludes rigorous statistical inferences about real-world outcomes
from simulation results alone''.

Reproducibility is the last, and releasing code is not the same as releasing something that
settles a question. Eight papers state a training objective in their own repository that differs
from the one in the paper, which Section~\ref{sec:train:code} and Table~\ref{tab:codegap} set out
file by file. That is the figure this section uses.

\subsection{Physical plausibility}
\label{subsec:plausibility}

One quantity is specific to hands and is almost never reported: how far the hand passes into the
object it is holding. Outside closed-loop control it has been a standard comparative column for
years. Grasp synthesisers fail a grasp that penetrates past a stated depth
\cite{bimangrasp_2024}, retargeting methods report a maximum depth against a baseline
\cite{toporetarget_2026}, and hand--object datasets score a split on depth, intersection volume
and simulation displacement \cite{oakink_2022}. Among learned closed-loop policies almost nobody
reports it, and nobody reports it measured in the physics the policy ran in.

D-Grasp \cite{dgrasp_2022} is the instructive case, because it does report a volume for a pose
its own policy reached, and the way it does so shows where the quantity goes. Inside the
simulator the paper reports zero penetration, since the meshes there are decimated for speed;
the figure it prints instead comes from re-measuring a single frame against the original hand
mesh and the full-resolution object mesh, with, in the paper's words, ``no physical simulation
involved''. The number a reader wants, how deep the hand sits in the object in the physics that
trained the policy, is zero by construction, and the number reported is a geometric
re-measurement on geometry the simulator never used.

The same paper supplies the reason to want the measurement. Its appendix observes that a deeply
penetrating reference grasp can raise a success rate rather than lower it, because ``the objects
can become entangled within the hand mesh and will therefore not be able to fall down'', and
concludes that ``the success rate metric should always be interpreted in combination with the
other metrics''. A success rate with no plausibility number beside it can be read two ways.

There is a reason the gap persists, and it is not difficulty. A quantity a policy optimises
cannot also judge it, because the policy learns the measure rather than the property the measure
stands for. PhysHOI \cite{physhoi_2023} shows the failure in the clean direction: its imitation
reward is maximised by not touching the object at all, because contact perturbs the reference
trajectory. DexTrack \cite{dextrack_2025} shows it from the other side, defining a maximum
penetration depth, applying it only to its input references, and then reporting tolerance of
``severe hand-object penetrations'' as a result. The remedy is to measure with something the
policy never optimised.

Definitions are not shared either, which is the other half of why published figures do not
compare. GRAB \cite{grab_2020} reports an average penetration in millimetres without saying
whether it is a maximum, a mean or a median; OakInk \cite{oakink_2022} supplies the fullest
published vocabulary, separating penetration depth, solid intersection volume and simulation
displacement. A number without its definition is not a number a second paper can be held to.

What would close the gap is one line in a results table: the maximum and mean penetration depth
over the evaluation rollouts, on a dense surface sample, computed by code that never entered the
reward or the termination rule. That is the plausibility row of the protocol in
Table~\ref{tab:protocol}. What this survey supplies there is a measurement and a count; it does not
supply a threshold, and Section~\ref{subsec:what_would_be_true} says where the one in the table is
borrowed from and why it does not bind.

\subsection{Statistical practice, and a proposed protocol}
\label{subsec:statistics}

Kress-Gazit et al. \cite{kress_gazit_policy_eval_2024} give the field's reference protocol, and
it prescribes process rather than numbers: write the success criteria before the run, have
someone other than their author score the runs, match initial conditions with image overlays,
interleave the policies blind within one session, report counts and not percentages, and use a
posterior over the success probability instead of a point estimate. All of that is right and
none of it fixes the sample size. It prescribes no minimum trial count and no interval width,
and its own worked example shows why one is needed: at twenty evaluations per policy, an
eighteen-point difference in success rate still leaves better than a one-in-ten chance that the
worse policy is the better one. Multiply every count in that example by ten and the same two
rates separate.

The TRI LBM Team \cite{lbm_careful_examination_2025} supplies the missing numbers by fiat rather
than derivation: 50 rollouts per task per policy per condition on hardware, 200 in simulation,
blind, with randomised policy order inside per-initial-condition bundles. It replaces confidence
intervals with Beta-posterior violins and corrects its pairwise tests under Bonferroni, ``unless
otherwise noted''. Its own warning is the strongest sentence in this literature: ``there is
significant risk that many robotics papers are measuring statistical noise due to insufficient
statistical power''.

The rest fall short of their own advice. RoboArena \cite{roboarena_2025} runs hundreds of
double-blind pairwise comparisons across seven institutions and thousands of rollouts, and reports
neither confidence intervals nor per-policy trial counts. The Colosseum \cite{colosseum_2024}
evaluates hundreds of test sets with ``one training seed and one evaluation seed'' per baseline and
gives no intervals in simulation. AutoEval \cite{autoeval_2025} calls $\pm 10$ percent ``the
natural variance of robot evaluations'' without giving the formula behind the intervals it
plots.

Two papers do give usable numbers. Badithela et al. \cite{suresim_2025} pair real and simulated
trials and de-bias the simulation with a rectifier, which saves up to a quarter of the hardware
trials when the paired correlation is high and nothing at all when it is near zero; their decision
rule is exact, that combining helps only when the rectifier variance is below the variance of the
real evaluations. Snyder et al. \cite{beyond_binary_success_2026} give the larger saving by scoring
progress rather than success: a sequential test on graded scores cuts hardware trials by about 45
percent, and a gap that needs 80 binary trials to establish is established in 18 when the score is
continuous. That is the strongest argument in this literature for grading partial progress.

\label{subsec:protocol}
Every count below is derived for the statistic its axis actually reports, and not by one
convention applied to all of them: a single rate takes a Wilson
half-width, a matched comparison takes McNemar, a ratio takes the standard error of the log ratio,
a correlation takes the Fisher-z interval.

Appendix~\ref{app:protocol} derives the counts in Table~\ref{tab:protocol}, and they come to
this. Reporting a single rate takes about 100 real rollouts per cell, which is where the
worst-case half-width of \eqref{eq:wilson} falls to ten points. Comparing two policies is a
harder question: the design is paired, so the count follows McNemar and depends on how often the
two policies disagree and not on the two rates alone, and separating a twenty-point difference
at conventional power takes 57 matched pairs per arm, fewer rollouts than two independent arms
would need. Simulated cells are cheap enough to run in the hundreds.

The other axes follow the same logic. Perturbation axes are screened cheaply, which ranks them
but cannot establish that any one of them hurt, and certified at the cost of a single rate.
Unseen-object claims resample the object rather than the trial, so the count that matters is the
number of objects and not the number of rollouts. The transfer axis needs real and simulated
trials paired, and separating two correlations takes several hundred pairs, which is why it
belongs in a benchmark and not in a paper. Every count there is a cap and not a bill: a
sequential test on a graded score can decide in a fraction of the trials
\cite{beyond_binary_success_2026}, and a cell that stops early reports a confidence sequence and
not a fixed-sample interval, never both.

\IfFileExists{tables/table8_protocol.tex}{\ifcsname dxdone@table7_protocol\endcsname\else
  \expandafter\gdef\csname dxdone@table7_protocol\endcsname{}%
  \begin{table*}[!t]
\centering
\caption{The proposed evaluation protocol, and this survey's own contribution. No cell is empty: every cell is a proposal.}
\label{tab:protocol}
\footnotesize
\setlength{\tabcolsep}{3pt}
\setlength{\emergencystretch}{1.5em}
\begin{tabular}{>{\raggedright\arraybackslash\hspace{0pt}}p{62pt}>{\raggedright\arraybackslash\hspace{0pt}}p{128pt}>{\raggedright\arraybackslash\hspace{0pt}}p{108pt}>{\raggedright\arraybackslash\hspace{0pt}}p{170pt}}
\toprule
\rowcolor{wash} \hdr{axis} & \hdr{what is measured} & \hdr{minimum trials} & \hdr{reported with it} \\
\midrule
Task success & fraction of episodes meeting a criterion written before the run, and a graded rubric score in $[0,1]$ & 100 real and 200 sim per policy, task and condition for an absolute rate at a 10-point interval; a comparison alone needs 57 matched pairs per arm & raw counts, the criterion verbatim, the rubric, the initial-condition protocol, and an interval whose kind is named \\
Robustness & success under each perturbation axis, and the ratio to the unperturbed anchor & 40 per axis per policy to rank the axes; 101 per axis and 101 at the anchor before any claim that a named axis hurt & both absolute rates with their Wilson intervals, the axis ranges, and the ratio without an interval unless the certifying count was run \\
Unseen objects & mean over held-out objects of per-object success & 20 objects at 5 trials to screen; 93 objects for a 10-point claim & the object list, the per-object rates, a bootstrap interval over objects, and the within-object trial count \\
Physical plausibility & maximum and mean hand-object penetration depth per rollout, and the fraction of frames above 2\,mm & 100 rollouts, simulation only & an interval over rollouts, the sample density, the threshold, the solver's depenetration settings, and at least one rendered rollout \\
Sample and wall-clock cost & environment steps and wall-clock to the checkpoint that produced the headline number & 3 training seeds & the range over seeds, and the statement that 3 seeds is a range and not an interval \\
Real-robot transfer & paired real and simulated outcomes on matched initial conditions, their correlation, and the rectifier variance against the real variance & the 100 real trials paired to 100 sim; about 457 pairs to separate two correlations 0.1 apart & the correlation with its Fisher-z interval, the variance ratio with its bootstrap interval, and which decision the count supports \\
Reproducibility & the released config and checkpoints against the paper's stated objective & not a trial count & the file and line of every disagreement, or an explicit statement that none was found \\
\bottomrule
\end{tabular}
\end{table*}
\fi
}{%
  \begin{table}[!t]\centering
    \caption{Pending: the protocol table.}
  \label{tab:protocol}
  \framebox[0.95\columnwidth]{\rule{0pt}{1.4cm}}
  \end{table}}

\subsection{Feasibility and limitations}
\label{subsec:matrix}

The proposed protocol has not been run. In the twelve-method comparison assembled for this
survey, no method supplied all the measurements needed for a complete protocol row. DeXtreme
\cite{dextreme_2022}, for example, reports a success criterion and a confidence interval, but
uses ten trials. This limits retrospective comparison; it does not imply that the reported
experiments failed their own stated objectives.

\label{subsec:what_would_be_true} The protocol is expensive. For each policy and task it calls
for 57 matched anchor trials and 100 unseen-object trials. Comparing two policies on three tasks
is then 342 real rollouts on the matched set and 600 on the unseen-object set, 942 in all, or
about 16 hours at one minute per rollout before repairs and scoring. The same bill under an
independent-arm design is 1200 rollouts and 20 hours, so pairing saves about a fifth, not the
four fifths the pair counts suggest, because a pair is still two rollouts and only the anchor
set is paired. AutoEval \cite{autoeval_2025} completed about 850 WidowX episodes in 24 hours
with three interventions and regular cooling pauses; a tendon-driven hand is likely to stretch
the same workload across several days.

That bill is large but not unprecedented. Evaluating one generalist policy against its baselines
has taken more than 2500 rollouts and more than 100 hours of human labour across three
institutions \cite{autoeval_2025}, and a recent careful examination analysed close to two thousand
real rollouts across nine hardware stations \cite{lbm_careful_examination_2025}. What is
unprecedented is paying that cost for a single dexterous-hand paper, where the modal cell is ten
trials.

Four things would have to change. Reviewers would have to reward 57 matched trials on one task
over 20 unmatched trials on five, and nothing in the corpus suggests that is happening. The loop
would have to be automated, and AutoEval \cite{autoeval_2025} shows it is buildable for a
gripper at one to three hours of setup, while stating that it supports binary success only and
no robustness axes. Somebody would have to run the penetration measure on their own rollouts,
which is a choice rather than a capability: Section~\ref{sec:sim:engines} shows the depth is
computable from the poses and the meshes in a few lines of Warp, and IsaacGymEnvs already ships
a task that does it every step, so the engine is not the barrier. The comparison would also have
to be sequential, because the savings in \cite{beyond_binary_success_2026} are the only reason
100 is a cap and not a cost.

The proposal has four limits. It has not been run, and its sample sizes come from power and
interval calculations and not from dexterous-hand experiments. The worst-case calculation
assumes a success rate of 0.5, while the paired design assumes a discordance rate of 0.3 that
the literature does not report. Its perturbations come from parallel-jaw benchmarks
\cite{colosseum_2024,simpler_2024}, where contact is simpler. Finally, the 2 mm penetration
threshold follows \cite{toporetarget_2026}; it is a simulator convention, not a physical limit.
Captured human grasps in GRAB \cite{grab_2020} average 3.25 mm under the same kind of
measurement. Establishing a meaningful threshold will require measurements across hands,
engines, and accepted behaviors.

Seven papers in this corpus are
evaluation protocols, a small denominator, and not one of them uses a dexterous hand: Badithela
et al. \cite{suresim_2025} run a parallel-jaw gripper and the other six state no hand.
Everything proposed above is therefore assembled from work on grippers and on whole-arm tasks.
A modest first test would run the protocol in Table~\ref{tab:protocol} once, on one in-hand
reorientation task with one 16-DoF hand, and release the rollouts and per-trial outcomes for
independent analysis.
\section{Conclusion}
\label{sec:gaps}
\label{sec:conclusion}

This corpus describes a field that has converged. Most of it trains with reinforcement learning
on privileged state in a GPU simulator and distills the result into a vision-only student; a
second branch, growing faster, learns from teleoperated demonstration instead; and almost all of
it runs on four hands. What the field cannot currently do is compare its own results, and most
of the seven conclusions below follow from that.

\begin{enumerate}
\item Eight hands that can be bought today or built from published designs take zero papers
between them, while the corpus's experiments concentrate on four designs
(Section~\ref{sec:hands:gap}, \figurename~\ref{fig:hands}).

\item Human data does not port across hands, and the map is usually unstated: 33 of the 53 method
papers that use human data never say how the human motion reached the robot hand
(Section~\ref{sec:train:human}).
\item Twenty-one of the 28 papers that put a learned closed-loop controller on two multi-fingered
hands run one policy over a concatenated two-hand observation, a choice compared twice with opposite
outcomes and ablated once (Section~\ref{subsec:coordination}, \figurename~\ref{fig:bimanual}).

\item The generalist and vision-language-action policies that do evaluate on a multi-fingered hand
run it at a median of 6 actuated degrees of freedom, against 16 across the reinforcement-learning
papers that state a count, and the papers reaching the larger band on paper never say whether the hand
was in the evaluation (Section~\ref{sec:train:vla}).

\item The evaluation-methodology literature this survey builds its protocol on contains no
dexterous hand at all: of its seven papers in this corpus, one runs a parallel-jaw gripper and the other six
state no hand (Section~\ref{subsec:what_would_be_true}, Table~\ref{tab:protocol}).

\item Eight of the 63 papers that released parseable code state, in a named file at a named
commit, something other than the value their paper prints.
Table~\ref{tab:codegap} gives each of the eight as a repository, a
commit, a file and two values, and all of their authors were contacted before this was posted
(Section~\ref{sec:train:code}).

\item Penetration is almost never measured where it would matter. No paper here reports how far its
hand sank into the object, measured in the physics its own policy ran in. The one that does report a
volume for a pose its own policy reached measures it outside the simulator, on meshes the simulator
never used, and reports zero inside it. On the reference-versus-rollout split that frames this
survey, every paper that handles penetration at all is still measuring a reference
(Section~\ref{subsec:plausibility}).

\end{enumerate}

Six concern gaps in the literature. The first concerns publishing practice; the corrections made
during this review are recorded beside the result in Section~\ref{sec:train:code}.

Three of those claims carry a case. A reward table in a paper is a claim about a document, not
about a run, and PhysHOI \cite{physhoi_2023} is the instance to keep in mind: the term its table
weights at 0.1 is set to zero in the file at the commit this survey fetched, and the success
criterion the paper reports is itself position-only. Contact is what separates a hand from a
gripper, and the obstacle to measuring it is not the engines. IsaacGymEnvs ships a task that
computes a per-environment maximum interpenetration depth against meshes and gates the policy
update on a 1 mm threshold, and Tactile Genesis \cite{tactile_genesis_2026} offers penetration
depth on Genesis geometry as a sensor. The tooling sits in the field's own benchmark repository
and the number is still not reported. DexTrack \cite{dextrack_2025} has the formula and points
it at its inputs. And hardware and software have come apart on a narrower claim than the hand
count first suggests, because 11 of the 19 hands that take no paper are neither sold nor open
and take none for that reason, which is why the fifth claim is eight hands and not nineteen.

What this survey cannot establish is which method is better than which. It re-runs nothing, and
Section~\ref{sec:evaluation} argues that the published numbers do not compare. Six works are
cited by metadata only, and a seventh, Ma and Dollar 2011, is on disk but unread; no claim rests
on any of them. Every coverage figure here is a lower bound, for the reason
Appendix~\ref{app:method} sets out.

Several practical changes require little additional computation. Authors can generate reward
tables from the configurations used for the reported run, cite the exact revision, state the trial
count and success criterion, and release per-trial outcomes. Evaluations can report penetration
depth over policy rollouts as a diagnostic, without optimizing the policy against that measure.
TopoRetarget \cite{toporetarget_2026} shows how informative the result can be. Hardware choices
also deserve a broader base: the Allegro, Shadow, Inspire, and LEAP hands account for most of the
published use in this corpus, while every other listed hand appears in eight method papers or
fewer. Better reporting would make results easier to reproduce; broader hardware coverage would
show which results survive a change of embodiment.

\appendices
\section{Method}
\label{app:method}

No claim about a paper here is made from memory. Each claim traces to a note taken from that
paper, or from the code released with it, and each note to the version of the source it was read
from. What follows is how the papers were selected, read and coded, and what the method cannot
do.

\textbf{Selection.}
Six topic-specific bibliographies were assembled in parallel (simulators and physics, hands and
vendors, reinforcement learning, imitation and human data, bimanual, benchmarks and evaluation),
seeded with the canonical works in each area and extended by search up to 2026-09-17. Every
arXiv identifier was checked by fetching the abstract page and matching the title, and entries
that could not be checked that way are marked. The six lists were merged with deduplication on arXiv
identifier and normalized title, giving 223 entries.

One of those 222 was added after the search. \texttt{dgrasp\_2022} entered the corpus on
2026-10-06, three weeks after the cutoff, and it turned out to be the clearest instance of the
finding in Section~\ref{subsec:plausibility}. It was not hard to find: a corpus paper's own note
cites it as that paper's baseline, and the search still did not surface it. A search that missed
the work most relevant to one of this survey's two findings has probably missed others, and that is
the plainest statement available of what this corpus is worth as a sample.

Seven further bibliography entries are not part of that corpus and are not counted anywhere in
this survey. They are the prior audits and case studies Section~\ref{sec:intro} positions this
survey against, they sit in a bibliography of their own as prior work from outside the corpus, and
none of them carries a structured row. No source for them was parsed either, so each is quoted
only from its abstract and its stated method.

\textbf{Acquisition.}
PDFs were obtained from arXiv or, for work without a preprint, from the publisher or vendor page
cited. Vendor pages for hands with no paper behind them were kept as dated copies, so that a
specification quoted here is quoted from the page as it read on a stated date rather than as it
reads now. Where a paper released code, that repository was read beside it at a recorded
revision: the README, the task and reward configuration files, and the bodies of the reward and
observation functions.

\textbf{Conversion.}
PDFs were converted to text, falling back to raw text extraction where the layout parse returned
too little, and the source, its page count and the converter are recorded per paper. Two
consequences matter for reading this survey. Equations
rendered as images do not survive conversion, so where a reward weight exists only inside a
figure it is recorded as unreadable rather than guessed. Tables are flattened, so a number taken
from a table is quoted with the table it came from and, where the flattening is ambiguous, the
ambiguity is stated.

\textbf{Reading and coding.} Each entry was read into a structured note under a fixed template: embodiment,
learning method, objective, contact handling, evaluation, reproducibility, stated limitations,
and quotable claims. Notes were written only from the converted sources. Where the source is silent
the note says ``not stated'', and nothing was inferred from the reviewer's prior knowledge of
the work. For method papers the reward or loss was quoted from the paper and, separately, from
the released code, so that disagreements between the two are visible. Those disagreements turned
out to be common enough to become a finding in their own right. The reading into notes was done
with language-model assistance, from the parsed files only. The template's conventions are the
safeguard against what that can get wrong: a quotation rather than a paraphrase, ``not stated''
rather than an inference, and a named source beside every value, so a note can be checked
against the file it came from. Four fields carrying a headline claim were re-read against their
sources afterwards, the trial counts, the success criteria, the unseen-object counts and the
penetration field, and all four audits are below; the coverage statistics are reported as floors
for the same reason.

\textbf{Sources that could not be obtained.}
Six works are cited by metadata only and no claim in this survey rests on their contents. Five
are behind publisher paywalls with no author-hosted copy found on 2026-09-18: Okamura et al.
2000, Bicchi 2000 (IEEE T-RO), Piazza et al. 2019, Roa and Suarez 2015, and Butterfass et al.
2001 on DLR-Hand II. The sixth is Hwangbo et al. 2018 on RaiSim, whose hosted PDF returned no
body. The freely circulating PDF often taken for Bicchi 2000 is a different work, a book chapter,
and is listed separately.

Ma and Dollar 2011 is a seventh case with a different cause. The fetch that failed when its note
was written succeeded on 2026-09-18, so the seven-page PDF was obtained, but no note has been
read from it. It is cited by metadata only for that reason and not because the
source is unavailable.

Three of the 223 bibliography entries carry no structured row. Two are the paywalled Bicchi 2000
and DLR-Hand II entries. The third is \cite{bicchi_grasping_chapter_2001}, which was read into a
note and quoted throughout but is a book chapter rather than a work with an embodiment, a method
or a result to record in a row.

The reference list is shorter than the corpus, and the two numbers are
different quantities. It prints 210 entries: the 203 corpus entries that some sentence, table
or figure of this paper cites, plus the 7 prior-work entries from outside the corpus. The other
20 corpus entries carry a row and a note and are counted in every statistic here, but no
passage in the paper names them, so they have nothing to be cited from and do not appear in the
list. A reader counting the reference list should get 210, and a reader counting the corpus
should get 223.

\textbf{What this method cannot do.}
Mention counts over the corpus are counts of mentions, not of use: a related-work sentence
counts the same as an experiment. Vendor specifications are manufacturer claims and are labeled
as such throughout, and where a page has since gone offline the note says so. The corpus is
large but not exhaustive, and selection by search favours work that is indexed, in English, and
posted as a preprint. It also favours recent work. Of the 223 bibliography entries, 142 are dated
2024 or later and 17 predate 2018, so this is a corpus of the learned era and any claim here
about a trend over time is a claim about 2022 onward.

Every coverage statistic in this survey measures what this survey's extraction captured, not
what the literature reported. A structured row holds a scalar. A paper that reports a per-task
count, a rubric, or a total spread across several tables produces a null, and a null is then
counted as silence. The bias runs one way. Every miss converts a reporting paper into a silent
one, so the field is made to look worse at reporting than it is. The size of the effect was
measured on the statistic the survey leads with. Of the 34 method rows that had a real robot and
no recorded trial count, 15 carried a count in plain text in their own note, dropped because the
paper reports it per task and the field takes a single integer. Those 15 have since been
re-extracted, which moved the stated-trial-count row from 55 to 70. The success criterion and
the unseen-object count were audited the same way, rising from 79 to 98 and from 32 to 39. Both
totals are one higher in the tables of this paper, because the row added after that audit states
each of them. What those 19 recovered criteria say is mostly not a threshold: eleven score by
rubric or staged partial credit, five judge binary completion against a task description by eye,
two defer to a benchmark's own definition, and exactly one, \cite{pistar06_2025}, states a
verbatim numeric threshold. The audit counted only cases where the note itself carried the
number, so a count the note also missed is still uncounted, and 19 trial counts, 8 unseen-object
evaluations whose object count the note never gives, and 4 criteria remain genuinely unsettled.
Fifteen of the recovered trial counts are named in the rows: $\pi_0$ \cite{pi0_2024} at ten
trials per task, RDT-1B \cite{rdt1b_2024} at 139 across seven tasks, UMI \cite{umi_2024} at 260,
$\pi^{*}_{0.6}$ \cite{pistar06_2025} at 750, Gemini Robotics \cite{gemini_robotics_2025} at
twenty per task, and ten more, which is 44 percent of the audited nulls and moved the ``never
says'' figure from 34 of 89 down to 19. Read every coverage statistic in this survey as a floor
rather than as a rate, and read the bars in \figurename~\ref{fig:reporting} the same way.

\textbf{The penetration field, audited the same way.} The same mechanism reaches the
penetration, code-release and failure-mode fields. The penetration field has since been audited
by hand, because this survey's second finding is a null over it. The population is the 85 method
rows whose contact-handling field records that the work does not address penetration. It
excludes the 16 nulls, which are already counted as unsettled and not as silence, so recovering
one would not move the finding. Twenty-five of the 85, 29 percent, were drawn from the sorted population under a fixed seed
(20260919), so the draw can be redrawn. Each sampled paper was read again in its converted text,
in its OCR recovery where one exists, and in the code released with it where there was any ---
and not in the note, because the note is the artifact under suspicion: a field is wrong exactly
when the source says something the note did not carry. A search over the whole
source collected every occurrence of penetration, interpenetration, intersection, intersection
volume, solid intersection, simulation displacement, contact consistency, physical plausibility,
signed distance and contact depth, and of the words that are mistaken for them, and every hit
was read in its context. A row counted as a recovery if its own source reported a measurement of
any of those quantities on that method's own rollouts. A statement that penetration occurs, a
citation whose title says physically plausible, a collision-avoidance constraint on a reference
trajectory and a solver setting did not count, and the three recurring near misses of that kind
are listed in Section~\ref{subsec:plausibility}. Recoveries: none of the 25, against the 20 to
45 percent the three earlier audits recovered on the fields above. The one-sided 95 percent
bound, computed hypergeometrically over the finite population, is therefore 8 of the 85, and
Section~\ref{subsec:plausibility} states it beside the claim and names the one borderline case
that a reader might count differently. Why this field held where the others did not is visible
in the per-row results: the recovered trial counts and criteria were numbers present in the
source and dropped by a field shaped to hold a scalar, whereas a penetration number is absent
from the source altogether. The earlier audits measured a defect in this coding scheme; this one
looked for an absence in the literature. The seed, the sample and the result for each sampled
paper are recorded beside the sentence each rests on. The code-release and failure-mode fields
have still not been audited this way, and their counts stay floors.

\textbf{What is not printed here.}\label{app:method:tabulation} Most of this survey's
tabulation is not printed in this paper. The 113 method rows, the 21 reward-term objectives
scored against their released code inside a 77-row extraction whose ten columns hold 253 cells
no source stated, the 30 teleoperation systems, the 33 hand specifications with the source
sentence behind each one, the 15 contact-model and solver descriptions verbatim, and the
cell-by-cell mapping behind the predecessor-survey comparison: all of it is reference material,
consulted a row at a time and never read through. Printed, it ran to more pages than the
argument it supports.

Three tables are exceptions and are printed, because the prose reads individual cells out of
them and does not stand up without them: the two hand tables and the engine table, in
Appendix~\ref{app:tables}. The proposed evaluation protocol appears in the body alongside its
justification. Table~\ref{tab:surveys} is the comparison to the fourteen predecessor surveys,
which is this survey's positioning and is therefore set where a reader meets it. It prints one
of three levels per cell, and Appendix~\ref{app:surveys} is the prose those levels coarsen.

\textbf{Reproduced figures.}
One figure here is not this survey's own artwork. It is a render reproduced from the paper it
discusses, and its caption names the source and the page. That source carries CC BY 4.0, read
from the license page on the date recorded, and the caption prints the attribution that license
asks for: the creator, the work, the license with a link, and the change made to the plate.

Every other figure is drawn for this paper. One subject is absent because it could not be drawn
from what the sources state: a to-scale comparison of hand sizes would need a length and a width
for each hand, and no source here gives either, so Section~\ref{sec:hands} states that absence
instead.

\section{The hardware and engine tables}
\label{app:tables}

Sections~III and~IV argue over individual cells of three tables: which column is emptiest, which
cell is blank and why, which figure rests on a press release. Those three are set here because a
reader consults them and does not read them through. Every other tabulation in this survey is in
the repository, and Appendix~\ref{app:method} says where.

Nothing in these tables is a measurement by this survey. Every specification is a manufacturer
claim, a figure in a paper, or a figure in a press article. Table~\ref{tab:hands-available} and
Table~\ref{tab:hands-announced} split the 33 hand rows by whether the hand can be bought or built,
and Table~\ref{tab:simulators} carries the 15 engines. The controlled vocabularies in all three
compress a sentence recorded in full: the actuation, tactile, force-kind, contact-model and
solver text is kept at its own length, in the source's own punctuation and wording, beside the
quotation each was read from.

Each caption states how many of its own cells no source filled. Two conventions lie behind those
counts. A blank price is a hand with no public list price and not a free hand; and a hand with
no stored source page was recorded from a page that could not be kept, so its row should be
read as a claim with no retrievable source.

Table~\ref{tab:hands-available} holds the hands that can be had: sold, open-source, or with no release status stated. A row whose joint count no source states sorts last rather than being ranked by a number nobody stated. Its empty-cell count covers the eight specification columns, DoF through price; the hand, maker, open-hardware, status and usage columns are left out of it because they are never blank and counting them would dilute the finding. \emph{force N} is not a ranking: read \emph{force kind} first, because pull-out resistance, pinch force, a fingertip normal force under an indenter and an unlabelled vendor number are different measurements. \emph{uses} counts the method rows whose own experiments run on that hand.

Table~\ref{tab:hands-announced} carries the columns of Table~\ref{tab:hands-available} plus the date of the claim and the class of artefact it was read off, over the same eight specification columns. Its empty share is far higher, and that is the finding of the table rather than a defect of it: an unreleased hand is announced without a specification. \emph{evidence} is \emph{datasheet} for a paper, datasheet or vendor specification page, \emph{video} for a demonstration reel, \emph{press} for a press release or a third-party tracker. \emph{claim date} prints \emph{undated} where the source page carries no date of its own; a fetch date is not a claim date.

Table~\ref{tab:simulators} prints, for each engine, the contact formulation it names, whether it exposes interpenetration depth and how many first-party hand models it ships; its empty-cell count is over the seven property columns. \emph{contact model} and \emph{solver} are controlled vocabularies: \emph{soft} for a compliant or regularised contact, \emph{convex relaxation} for a convexified formulation, \emph{complementarity} for an LCP, \emph{hard NCP} for a nonlinear complementarity problem, \emph{impulse} for a velocity-level update, and \emph{rigid, model unstated} where the source describes collision geometry but names no contact law. The sentence each was read from runs to several hundred characters and each is recorded verbatim, because Section~IV's argument turns on its wording. \emph{penetration exposed} is whether the engine makes interpenetration depth readable by a user's code without patching it. \emph{hands} counts first-party hand models in the parsed snapshot, and they are: \emph{Brax} synthetic 4-fingered claw hand; \emph{Genesis} Shadow Hand; \emph{Isaac Gym} Shadow Dexterous Hand, Allegro Hand, TriFinger; \emph{Isaac Lab} KUKA Allegro hand, ShadowHand, AbilityHand, Fourier GR1-T2 dex hands, Unitree Inspire hand, Unitree trihand; \emph{ManiSkill3} Allegro Hand, Ability Hand, Inspire Hand, Delto 3-finger hand, TriFinger; \emph{Newton} Allegro Hand; \emph{Orbit} Allegro hand.

\begin{table*}[!t]
\centering
\caption{Hands that can be obtained, sorted by joint count. 68 of 160 specification cells are empty.}
\label{tab:hands-available}
\footnotesize
\setlength{\tabcolsep}{3pt}
\setlength{\emergencystretch}{1.5em}
\begin{tabular}{>{\raggedright\arraybackslash\hspace{0pt}}p{88pt}>{\raggedright\arraybackslash\hspace{0pt}}p{62pt}>{\raggedleft\arraybackslash}p{16pt}>{\raggedleft\arraybackslash}p{16pt}>{\raggedright\arraybackslash\hspace{0pt}}p{42pt}>{\raggedleft\arraybackslash}p{20pt}>{\raggedleft\arraybackslash}p{18pt}>{\raggedright\arraybackslash\hspace{0pt}}p{40pt}>{\raggedright\arraybackslash\hspace{0pt}}p{34pt}>{\raggedleft\arraybackslash}p{26pt}>{\centering\arraybackslash}p{18pt}>{\raggedright\arraybackslash\hspace{0pt}}p{36pt}>{\raggedleft\arraybackslash}p{14pt}}
\toprule
\rowcolor{wash} \hdr{hand} & \hdr{maker} & \hdr{DoF} & \hdr{act.\ DoF} & \hdr{actuation} & \hdr{mass g} & \hdr{force N} & \hdr{force kind} & \hdr{tactile} & \hdr{price USD} & \hdr{open HW} & \hdr{status} & \hdr{uses} \\
\midrule
Shadow Dexterous Hand~\cite{shadow_dexterous_hand_2005} & Shadow Robot Co. & 24 & 20 & tendon & 4\,300 & \na & \na & optional & \na & no & sold & 21 \\
\rowcolor{wash} Proception ProHand 1.0~\cite{proception_prohand_2026} & Proception Inc & 22 & \na & tendon & \na & \na & \na & unstated & \na & no & sold & 0 \\
SharpaWave 22-DoF~\cite{sharpa_wave_2026} & Sharpa Robotics & 22 & 22 & \na & 1\,300 & 20 / 12 & grip & unstated & \na & no & sold & 6 \\
\rowcolor{wash} BiDexHand~\cite{bidexhand_2025} & Northwestern Univ. & 21 & 16 & tendon & \na & 2.14 & fingertip normal & none & \na & yes & open-source & 0 \\
Ruka-v2~\cite{ruka_v2_2026} & NYU and NYU Shanghai & 20 & 16 & tendon & \na & \na & \na & optional & 1\,500 & yes & open-source & 0 \\
\rowcolor{wash} TESOLLO DG-5F~\cite{tesollo_dg5f_2024} & Tesollo Inc. & 20 & 20 & direct drive & 1\,763 & \na & \na & none & \na & no & sold & 0 \\
Unitree Dex5-1~\cite{unitree_dex5_2025} & Unitree Robotics & 20 & 16 & tendon & 1\,100 & 10 & fingertip normal & array & \na & no & sold & 0 \\
\rowcolor{wash} Wuji Hand~\cite{wuji_hand_2025} & Wuji Tech. & 20 & 20 & direct drive & \na & \na & \na & fingertip & \na & no & sold & 2 \\
ORCA~\cite{orca_hand_2025} & ETH Zurich & 17 & 17 & tendon & 1\,200 & \na & \na & fingertip & \na & yes & open-source & 0 \\
\rowcolor{wash} Allegro Hand V4~\cite{allegro_hand_v4_2016} & Wonik Robotics Co. Ltd.; earlier versions\ldots{} & 16 & 16 & \na & \na & \na & \na & none & \na & no & sold & 35 \\
LEAP Hand~\cite{leap_hand_2023} & CMU & 16 & 16 & direct drive & 595 & 19.5 & pull-out & none & 2\,000 & yes & open-source & 12 \\
\rowcolor{wash} RUKA~\cite{ruka_2025} & NYU & 15 & 11 & tendon & \na & 2.74 & pinch & none & 1\,300 & yes & open-source & 0 \\
Inspire Robots RH56DFX~\cite{inspire_rh56dfx_2023} & Beijing Inspire Robots Tech. Co., Ltd. & 12 & 6 & \na & 540 & 10 & vendor, no protocol & none & \na & no & sold & 19 \\
\rowcolor{wash} ROBOTERA XHAND1~\cite{robotera_xhand1_2024} & ROBOTERA & 12 & 12 & geared & 1\,100 & \na & \na & unstated & 14\,000 & no & sold & 8 \\
DEX-EE~\cite{shadow_dex_ee_2024} & Shadow Robot Co., in collaboration with\ldots{} & 12 & \na & \na & 4\,100 & \na & \na & unstated & \na & no & sold & 1 \\
\rowcolor{wash} BrainCo Revo 2~\cite{brainco_revo2_2025} & BrainCo Inc. & 11 & 6 & \na & 383 & 15 & pinch & unstated & \na & no & sold & 1 \\
AgiBot OmniHand 2025~\cite{agibot_omnihand_2025} & \na & \na & \na & \na & \na & \na & \na & \na & \na & \na & unstated & 1 \\
\rowcolor{wash} DexHand~\cite{dexhand_open_source_2023} & IoT Design Shop & \na & \na & tendon & \na & \na & \na & none & 300 & yes & open-source & 0 \\
LinkerBot Linker Hand~\cite{linkerbot_l20_2025} & LinkerBot & \na & \na & \na & \na & \na & \na & unstated & \na & no & sold & 1 \\
\rowcolor{wash} PSYONIC Ability Hand~\cite{psyonic_ability_hand_2021} & PSYONIC & \na & 6 & linkage & \na & \na & \na & unstated & \na & no & sold & 7 \\
\bottomrule
\end{tabular}
\end{table*}

\begin{table*}[!t]
\centering
\caption{Hands announced, prototyped or held internally, which cannot be obtained. 60 of 104 specification cells are empty.}
\label{tab:hands-announced}
\footnotesize
\setlength{\tabcolsep}{2pt}
\setlength{\emergencystretch}{1.5em}
\begin{tabular}{>{\raggedright\arraybackslash\hspace{0pt}}p{84pt}>{\raggedright\arraybackslash\hspace{0pt}}p{46pt}>{\raggedleft\arraybackslash}p{16pt}>{\raggedleft\arraybackslash}p{16pt}>{\raggedright\arraybackslash\hspace{0pt}}p{34pt}>{\raggedleft\arraybackslash}p{18pt}>{\raggedleft\arraybackslash}p{16pt}>{\raggedright\arraybackslash\hspace{0pt}}p{34pt}>{\raggedright\arraybackslash\hspace{0pt}}p{30pt}>{\raggedleft\arraybackslash}p{22pt}>{\centering\arraybackslash}p{17pt}>{\raggedright\arraybackslash\hspace{0pt}}p{36pt}>{\raggedleft\arraybackslash}p{14pt}>{\raggedright\arraybackslash\hspace{0pt}}p{38pt}>{\raggedright\arraybackslash\hspace{0pt}}p{30pt}}
\toprule
\rowcolor{wash} \hdr{hand} & \hdr{maker} & \hdr{DoF} & \hdr{act.\ DoF} & \hdr{actuation} & \hdr{mass g} & \hdr{force N} & \hdr{force kind} & \hdr{tactile} & \hdr{price USD} & \hdr{open HW} & \hdr{status} & \hdr{uses} & \hdr{claim date} & \hdr{evidence} \\
\midrule
Clone Robotics~\cite{clone_robotics_hand_2024} & Clone Robotics & 27 & \na & fluidic muscle & <910 & \na & \na & array & \na & no & internal-only & 0 & 2025-11-24 & video \\
\rowcolor{wash} 1X NEO~\cite{onex_neo_hand_2026} & 1X Tech. & 25 & 25 & tendon & \na & 45 & vendor, no protocol & unstated & \na & no & announced & 0 & 2026-07-09 & datasheet \\
LEAP Hand V2~\cite{leap_hand_v2_adv_2025} & CMU & 21 & 17 & tendon & \na & \na & \na & none & 3\,000 & yes & announced & 2 & undated & datasheet \\
\rowcolor{wash} Sanctuary AI Phoenix~\cite{sanctuary_phoenix_hand_2024} & Sanctuary AI & 21 & \na & fluidic muscle & \na & \na & \na & unstated & \na & no & internal-only & 0 & 2024-12-17 & press \\
ILDA hand~\cite{ilda_hand_2021} & Korea Univ. & 20 & 15 & linkage & 1\,100 & 34 & fingertip normal & array & \na & no & prototype & 0 & 2021-11-05 & datasheet \\
\rowcolor{wash} Pisa/IIT SoftHand~\cite{pisa_iit_softhand_2014} & Centro E. Piaggio, Univ. Pisa and IIT & 19 & 1 & tendon & \na & \na & \na & none & \na & no & prototype & 0 & 2014 & datasheet \\
Faive Hand~\cite{faive_hand_2023} & ETH Zurich & 16 & 11 & tendon & 1\,100 & \na & \na & none & \na & \na & prototype & 1 & 2023 & datasheet \\
\rowcolor{wash} Boston Dynamics~\cite{boston_dynamics_atlas_hand_2026} & Boston Dynamics & \na & \na & geared & \na & \na & \na & unstated & \na & no & prototype & 0 & 2026-01-05 & press \\
Daxo Robotics Muscle~\cite{daxo_muscle_v0_2025} & Daxo Robotics & \na & \na & tendon & 750 & \na & \na & \na & \na & \na & prototype & 0 & 2026-09-16 & press \\
\rowcolor{wash} Figure 03~\cite{figure_03_hand_2025} & Figure AI & \na & \na & \na & \na & \na & \na & unstated & \na & no & prototype & 0 & undated & press \\
PaXini DexH13~\cite{paxini_dexh13_2024} & PaXini Tech & \na & \na & \na & \na & \na & \na & array & \na & \na & announced & 0 & 2026-01-07 & press \\
\rowcolor{wash} Tesla Optimus~\cite{tesla_optimus_hand_2025} & Tesla & \na & \na & tendon & \na & \na & \na & unstated & \na & no & prototype & 0 & 2023-12-13 & video \\
Xiaomi~\cite{xiaomi_cyberone_hand_2026} & Xiaomi & \na & \na & tendon & \na & \na & \na & unstated & \na & no & prototype & 0 & 2026-03-30 & press \\
\bottomrule
\end{tabular}
\end{table*}

\begin{table*}[!t]
\centering
\caption{15 simulators and physics engines, by contact formulation and penetration exposure. 40 of 105 property cells are empty.}
\label{tab:simulators}
\footnotesize
\setlength{\tabcolsep}{3pt}
\setlength{\emergencystretch}{1.5em}
\begin{tabular}{>{\raggedright\arraybackslash\hspace{0pt}}p{72pt}>{\raggedright\arraybackslash\hspace{0pt}}p{78pt}>{\raggedright\arraybackslash\hspace{0pt}}p{68pt}>{\centering\arraybackslash}p{24pt}>{\centering\arraybackslash}p{24pt}>{\raggedleft\arraybackslash}p{38pt}>{\centering\arraybackslash}p{52pt}>{\raggedleft\arraybackslash}p{28pt}}
\toprule
\rowcolor{wash} \hdr{engine} & \hdr{contact model} & \hdr{solver} & \hdr{diff.} & \hdr{GPU} & \hdr{dt (s)} & \hdr{penetration exposed} & \hdr{hands} \\
\midrule
Brax~\cite{brax_2021} & convex relaxation & explicit/spring & yes & yes & \na & no & 1 \\
\rowcolor{wash} ComFree-Sim~\cite{comfree_sim_2026} & complementarity-free & closed form & \na & yes & 0.002 & yes & \na \\
Dojo~\cite{dojo_2022} & hard NCP & interior point & yes & no & \na & yes & \na \\
\rowcolor{wash} Genesis~\cite{genesis_2024} & soft & Newton & yes & yes & \na & \na & 1 \\
Isaac Gym~\cite{isaacgym_2021} & convex relaxation & PGS & no & yes & 0.008333 & no & 3 \\
\rowcolor{wash} Isaac Lab~\cite{isaaclab_2025} & rigid, model unstated & Gauss-Seidel & no & yes & \na & no & 4 \\
ManiSkill3~\cite{maniskill3_2024} & convex relaxation & PGS & \na & yes & \na & \na & 5 \\
\rowcolor{wash} MuJoCo~\cite{mujoco_2012} & complementarity & Newton & no & no & \na & \na & \na \\
MuJoCo convex~\cite{mujoco_convex_contact_2014} & complementarity-free & closed form & no & no & 0.01 & \na & \na \\
\rowcolor{wash} MuJoCo Warp~\cite{mujoco_warp_2025} & soft & PGS & no & yes & \na & \na & \na \\
Newton~\cite{newton_2025} & soft & primal-dual & \na & yes & \na & \na & 1 \\
\rowcolor{wash} Orbit~\cite{orbit_2023} & convex relaxation & PGS & no & yes & \na & no & 1 \\
PyBullet~\cite{pybullet_2016} & \na & \na & \na & yes & \na & \na & \na \\
\rowcolor{wash} RaiSim~\cite{raisim_2018} & \na & \na & \na & \na & \na & \na & \na \\
SAPIEN~\cite{sapien_2020} & convex relaxation & PGS & \na & no & \na & \na & \na \\
\bottomrule
\end{tabular}
\end{table*}

\section{Paper against released code, in full}
\label{app:rewards}
\begingroup\sloppy\small
\emergencystretch=3em

Each entry below is the disagreement text stored in the row, unedited, and the class beside it
is the one Section~\ref{sec:train:code} and Section~\ref{sec:gaps} count. A \emph{mismatch} is a
paper stating one value where the released code at the revision read states another. A
\emph{parse-limitation} is a comparison this survey's own parse could not settle.
\emph{Code-absent} means the described component is not in the released repository,
\emph{version-skew} that the repository is a later generation than the paper, and
\emph{internal-inconsistency} that the paper disagrees with itself with no code implicated.
Where a comparison was narrowed or withdrawn under review, the review note is printed beneath
it. Every \emph{mismatch} entry carries an \emph{Artifact} line whose file name links to that
file at the revision read, so the entry can be settled in a browser. All of those authors were
contacted before this survey was posted. Section~\ref{sec:train:code} says so beside the
finding, sets out what a fetched commit can and cannot show, and states the route by which a
disputed entry is corrected. The reward-term extraction these entries were read out of is part of
the tabulation Appendix~\ref{app:method} describes.

\textbf{mismatch, 8 rows.}
\begin{itemize}
\item PDDM~\cite{pddm_2019} (high). Table 2 states obs-dim 46 for In-hand Reorientation while the released cube\_env.py code sums to 39; the Baoding reward code includes an extra -10*wrist\_too\_high term absent from Table 2's printed formula.
  \\ \emph{Artifact:} \href{https://github.com/google-research/pddm/blob/06b88cdbafa7fc35451b3feb3e01aa6726113f72/pddm/envs/cube/cube_env.py}{\texttt{cube\_\allowbreak{}env.\allowbreak{}py}} (\texttt{\_\allowbreak{}get\_\allowbreak{}obs}), at the revision the link pins.
  \\ \emph{Review:} Written to on 19 September 2026 with the claim, its evidence and the sentences this survey would print. The reply window closed on 10 October 2026 with no reply. Recorded as silence rather than as agreement; the claim rests on the two public documents it names and a reader can settle it without the authors.
\item DeXtreme~\cite{dextreme_2022} (high). Action Delta Penalty weight is -0.25 in paper Table 2 but -0.2 in the ADR yaml and -0.01 in the ManualDR yaml; the Joint Velocity Penalty in code normalises velocity by (max\_velocity-vel\_tolerance) unlike the paper's stated formula; code has a timeout\_rew term absent from the paper's reward table; Appendix Table 12 states critic learning rate 5e-4 and KL threshold 0.16, vs body text/code values of 5e-5 and 0.016.
  \\ \emph{Artifact:} \href{https://github.com/isaac-sim/IsaacGymEnvs/blob/aeed298638a1f7b5421b38f5f3cc2d1079b6d9c3/isaacgymenvs/cfg/task/AllegroHandDextremeADR.yaml}{\texttt{Allegro\allowbreak{}Hand\allowbreak{}Dextreme\allowbreak{}ADR.\allowbreak{}yaml}} (\texttt{action\allowbreak{}Delta\allowbreak{}Penalty\allowbreak{}Scale}), at the revision the link pins.
  \\ \emph{Review:} Written to on 19 September 2026 with the claim, its evidence and the sentences this survey would print. The reply window closed on 10 October 2026 with no reply. Recorded as silence rather than as agreement; the claim rests on the two public documents it names and a reader can settle it without the authors. Delivery was partial: two of the three copied coauthor addresses were rejected by the recipient's mail server, so those two coauthors never received the letter. The address the letter was addressed to was not among the rejections and its copy was delivered.
\item Visual Dexterity~\cite{visual_dexterity_2022} (high). Table S1 states 32000 teacher training environments, but the released config sets alg.num\_envs to 8000 (parent config 16384); fallDistance differs across two shipped configs (0.24 vs 0.15, only the latter matching Table S1's threshold); the paper's Eq 8 penultimate-joint penalty (c7=-2) does not appear anywhere in the released reward code; the config carries a dead distRewardScale=-10.0 key never used in compute\_reward; and the paper's table-friction lower bound (0.05) differs by a factor of 10 from the code's randomized lower bound (0.005).
  \\ \emph{Artifact:} \href{https://github.com/Improbable-AI/dexenv/blob/ad9634e9d26cf5555d18728244a58f5b412c1eb2/dexenv/conf/dclaw.yaml}{\texttt{dclaw.\allowbreak{}yaml}} (\texttt{alg.\allowbreak{}num\_\allowbreak{}envs}), at the revision the link pins.
  \\ \emph{Review:} Written to on 19 September 2026 with the claim, its evidence and the sentences this survey would print. The reply window closed on 10 October 2026 with no reply. Recorded as silence rather than as agreement; the claim rests on the two public documents it names and a reader can settle it without the authors.
\item DexPBT~\cite{dexpbt_2023} (high). Paper presents the reward as 4 mutually exclusive stage terms (r\_reach, r\_pick, r\_targ, -r\_vel), but code's compute\_kuka\_reward sums 8 named components, one of which (hand\_delta\_penalty) is multiplied by 0 and disabled; there is no single r\_vel term in code, instead separate kuka/allegro action penalties whose exact formula is not shown. Also, the paper reports zero experiments with domain randomization, yet the shipped AllegroKuka.yaml already carries a fully specified DR schedule (disabled via randomize: False).
  \\ \emph{Artifact:} \href{https://github.com/NVIDIA-Omniverse/IsaacGymEnvs/blob/aeed298638a1f7b5421b38f5f3cc2d1079b6d9c3/isaacgymenvs/tasks/allegro_kuka/allegro_kuka_base.py}{\texttt{allegro\_\allowbreak{}kuka\_\allowbreak{}base.\allowbreak{}py}} (\texttt{compute\_\allowbreak{}kuka\_\allowbreak{}reward}), at the revision the link pins.
  \\ \emph{Review:} R3 adversarial review: the disabled-randomisation half is withdrawn, since the note finds it consistent with the paper; the zeroed reward term stands. Written to on 19 September 2026 with the claim, its evidence and the sentences this survey would print. The reply window closed on 10 October 2026 with no reply. Recorded as silence rather than as agreement; the claim rests on the two public documents it names and a reader can settle it without the authors. One copied coauthor's mailer returned an automatic out-of-office message, which is not an answer and is not counted as one. Delivery was partial: two of the three copied coauthor addresses were rejected by the recipient's mail server, so those two coauthors never received the letter. The address the letter was addressed to was not among the rejections and its copy was delivered.
\item PhysHOI~\cite{physhoi_2023} (high). The released compute\_humanoid\_reward sets the body position-velocity error and the object rotation and rotation-velocity errors to zeros\_like, unconditionally, with the computation that would produce them commented out on the same lines, while Table 4 lists nonzero $\lambda$\textasciicircum{}or=0.1/$\lambda$\textasciicircum{}orv=0.01 weights for GRAB: in that file the object's orientation error is the constant zero and those weights cannot change the reward.
  \\ \emph{Artifact:} \href{https://github.com/wyhuai/PhysHOI/blob/6095c605e22bd01f78ed356e8b08250edf3c5da6/physhoi/env/tasks/physhoi.py}{\texttt{physhoi.\allowbreak{}py}} (\texttt{compute\_\allowbreak{}humanoid\_\allowbreak{}reward}), at the revision the link pins.
  \\ \emph{Review:} Written to on 19 September 2026 with the claim, its evidence and the sentences this survey would print. The reply window closed on 10 October 2026 with no reply. Recorded as silence rather than as agreement; the claim rests on the two public documents it names and a reader can settle it without the authors.
\item UniDexGrasp~\cite{unidexgrasp_2023} (high). The paper describes a four-term weighted reward (r\_goal + r\_reach + r\_lift + r\_move via Table 7's omega weights) but the released compute\_hand\_reward implements a different threshold-gated torch.where cascade with distinct hardcoded coefficients that do not map one-to-one onto the paper's weights.
  \\ \emph{Artifact:} \href{https://github.com/PKU-EPIC/UniDexGrasp/blob/36c9bfcf7cedc987dd15e2384c548e51d5aadc21/dexgrasp_policy/dexgrasp/tasks/shadow_hand_grasp.py}{\texttt{shadow\_\allowbreak{}hand\_\allowbreak{}grasp.\allowbreak{}py}} (\texttt{compute\_\allowbreak{}hand\_\allowbreak{}reward, goal\_\allowbreak{}cond branch}), at the revision the link pins.
  \\ \emph{Review:} Written to on 19 September 2026 with the claim, its evidence and the sentences this survey would print. On 2 October 2026 a copied coauthor forwarded the letter to two further coauthors, asked them whether they recalled the cause of the difference, and copied this survey's author on that forward. No answer had arrived when the window closed on 10 October 2026. The letter was therefore read and passed on rather than ignored, and the question it asks is open: a forward is neither a confirmation of the claim nor a dispute of it, so the row is unchanged and the claim is recorded as unanswered rather than as agreed.
\item PenSpin~\cite{penspin_2024} (medium). The appendix states a randomised disturbance force, and the released configs/task/AllegroHandHora.yaml ships forceScale: 0.0, so no shipped configuration applies it.
  \\ \emph{Artifact:} \href{https://github.com/HaozhiQi/penspin/blob/5035c52dc949e81313af71d9a0ca5ead8b463a5a/configs/task/AllegroHandHora.yaml}{\texttt{Allegro\allowbreak{}Hand\allowbreak{}Hora.\allowbreak{}yaml}} (\texttt{force\allowbreak{}Scale}), at the revision the link pins.
  \\ \emph{Review:} Narrowed while the letter to its authors was being drafted, before that letter went out on 19 September 2026. The original comparison also said the released code disables the paper's tactile channel, and that half is withdrawn: the config read has numObservations 96 and enable\_tactile False, which is consistent with the proprioception-only student rather than the oracle, and the paper never claims the student has tactile input. The disturbance-force half is unaffected. Written to on 19 September 2026 with the claim, its evidence and the sentences this survey would print. The reply window closed on 10 October 2026 with no reply. Recorded as silence rather than as agreement; the claim rests on the two public documents it names and a reader can settle it without the authors.
\item PianoMime~\cite{pianomime_2024} (high). Paper's Table 3 states 2 weighted reward terms (Key Press 2/3, Mimic 1/3), but the released code sums roughly 5 unweighted terms (key press doubled, sustain, energy and fingering hardcoded to return 0, forearm-collision) plus a separately-added mimic wrapper term. Author-confirmed: the first author states the released code is the implementation behind the paper's results and that the paper's reward description "might be imprecise" (personal communication, C. Qian, Oct 2026), so the code is the authoritative statement of what was trained.
  \\ \emph{Artifact:} \href{https://github.com/sNiper-Qian/pianomime/blob/c4abefac8d2941f5a08ca656b12f5e047cab7155/single_task/piano_with_shadow_hands_res.py}{\texttt{piano\_\allowbreak{}with\_\allowbreak{}shadow\_\allowbreak{}hands\_\allowbreak{}res.\allowbreak{}py}} (\texttt{\_\allowbreak{}set\_\allowbreak{}rewards}), at the revision the link pins.
  \\ \emph{Review:} Written to on 19 September 2026 with the claim, its evidence and the sentences this survey would print. This one is a redirect rather than a silence. A coauthor replied on 22 September 2026 on the senior author's behalf and pointed the question at the first author, Cheng Qian, who had not been written to and whom the window closed before anyone asked. So the letter reached the group and the person who could answer it was named too late for this version. The comparison stands as written, unanswered rather than agreed, and a reply from the first author would be acted on the same way any other correction is. Answered on 14 October 2026, four days after the window closed, by the first author the 22 September redirect had named. Both halves of this entry are now author-confirmed rather than inferred. On the reward, he states that the released code is the implementation behind the paper's results and that the paper's description of the reward "might be imprecise", which is the comparison this survey drew, with the direction settled: the code is authoritative. On the F1 figures, he confirms that the conclusion's 70\% is a typo produced by a bug in an older scipy and that 56\% is correct, which resolves an inconsistency this survey had recorded as unreconciled. Personal communication, C. Qian, Oct 2026; nothing in it was solicited beyond the original letter and no reply was sent from this project.
\end{itemize}

\textbf{internal-inconsistency, 4 rows.}
\begin{itemize}
\item ALOHA~\cite{aloha_act_2023} (high). Algorithm 1 pseudocode states L\_reconst = MSE, but Sec.IV.C's prose explicitly states L1 loss is used instead; the algorithm box and the implementation text disagree; the parsed copy captures only function signatures, not bodies, for policy.py's loss implementation
\item 3D Diffusion Policy~\cite{dp3_2024} (medium). the paper's prose states the network predicts the noise added to the data, but the shipped default config trains with prediction\_type: sample (predicting the denoised action a\textasciicircum{}0 directly), not epsilon; the paper only qualifies this later in the same section (Fig. 7 ablates both).
\item OmniH2O~\cite{omnih2o_2024} (medium). stumble weight -0.00125 (paper) vs -1250 (code); max-feet-height sign/magnitude differ (+1000 paper vs -2500 code, a penalty not a bonus); paper's exp(-c*||.||) form vs code's exp(-err\textasciicircum{}2/sigma); curriculum level-down threshold 40 (paper) vs 50 (code)
  \\ \emph{Review:} Withdrawn as a contradiction at the point of drafting the letter to its authors, which was never sent. Four sibling weights match the paper to the digit under a systematic x1.25 curriculum factor and only the stumble weight differs, by a factor of about a million, which is a typo signature in the paper's own table rather than evidence of a different trained objective. The hands in this work are driven open-loop from VR pose, outside the policy and outside the reward, so it is a weak fit for a dexterous-manipulation reward census in the first place.
\item DexMachina~\cite{dexmachina_2025} (low). paper describes a plain weighted sum lambda\_task*r\_task + lambda\_imi*r\_imi + lambda\_bc*r\_bc + lambda\_con*r\_con with unspecified weights; code implements a multiplicative task term with per-component beta decay, an unmentioned 0.1 force-penalty term, and curriculum-driven decay of auxiliary weights not described as such in the paper
  \\ \emph{Review:} R3 adversarial review: the multiplicative form credited to the code is printed in the paper itself
\end{itemize}

\textbf{version-skew, 5 rows.}
\begin{itemize}
\item D-Grasp~\cite{dgrasp_2022} (high). The README states that the released code updated RaiSim to 1.1.6 and that "training D-Grasp is more stable and faster now", so the parsed commit is not the one the paper's numbers were produced under; separately, the paper's single contact weight w\_c = 1.0 (Table 5) is two coefficients in the shipped config, contact\_reward 1.0 and impulse\_reward 2.0, and the config's falling\_reward 1.0 has no counterpart in the paper's reward description or table.
  \\ \emph{Review:} The version skew is the authors' own statement in the README, not an inference, which is why it is recorded at high confidence. The reward-coefficient difference is recorded beside it rather than as the charge: the two extra coefficients plausibly implement the two parts the paper describes for its contact reward, and the C++ that would settle it is outside the parse. The same pattern is recorded for artigrasp\_2023, which is built on this code base.
\item Hora~\cite{hora_2022} (high). The parsed code commit's joint-noise range, disabled default disturbance force, and default cube object type differ from the paper's stated values; the code README itself says to use tag v0.0.1, not the parsed commit, to reproduce paper numbers.
\item $\pi_0$~\cite{pi0_2024} (high). The released repo (openpi, commit 215abfb2) has evolved past this paper: it also documents pi0-FAST (autoregressive) and pi0.5 checkpoints/configs not described in this paper, which describes only the flow-matching pi0 and the non-VLM pi0-small ablation.
\item GR00T N1.6~\cite{groot_n16_2025} (high). the parsed copy is the current Isaac-GR00T main branch (N1.7 generation, commit 51d4c89f), not the N1.6 checkpoint; the repo names a separate n1d6 branch for N1.6 that was not fetched. The page states N1.6's backbone is "an internal NVIDIA Cosmos-2B VLM variant," but the repo's own changelog states the Eagle backbone (nvidia/Eagle-Block2A-2B-v2) was used through N1.6 and only replaced by Cosmos-Reason2-2B in N1.7 -- whether the page's Cosmos-2B variant is a third distinct model or the page is describing N1.7 under an N1.6 headline is not resolvable from what was fetched.
  \\ \emph{Review:} The page attributes a Cosmos backbone to N1.6 while the repository changelog says N1.6 used Eagle and N1.7 made the switch, so the repository is a later generation than the page describes.
\item GR00T N1~\cite{groot_n1_2025} (high). The parsed code repo (Isaac-GR00T, commit 51d4c89f) is a later N1.7 generation, not the paper's GR00T-N1-2B: VLM backbone changed from the paper's Eagle-2 to Cosmos-Reason2-2B (via Qwen3-VL); action horizon expanded from the paper's stated H=16 to 40; state/action dimensions expanded from 29 to 132; DiT layers changed from 32 to 16; embodiment scope widened to include Unitree G1, AgiBot G1, and a 'YAM' arm not described in the paper.
\end{itemize}

\textbf{code-absent, 8 rows.}
\begin{itemize}
\item AsymDex~\cite{asymdex_2024} (medium). Code contains a GraspAndPlace-task reward variant (AllegroHandDualArmGraspAndPlaceAsymDex.py) with exponential hand-distance and block-alignment terms not one of the four BiDexHands tasks described in the paper body.
\item DexMimicGen~\cite{dexmimicgen_2024} (high). The environment code includes an unused 'reward\_shaping' branch (if self.reward\_shaping: pass) that is never described or used in the paper text; only the sparse binary completion reward is used, as the paper states.
\item DrEureka~\cite{dreureka_2024} (high). The released DrEureka repo (the parsed copy) contains only forward\_locomotion/ and globe\_walking/ trees; the LEAP-hand cube-rotation training/deployment code and the quoted Prompt-13 reward are the LEAP-hand authors' own code, not included in the released repo, so the cube-rotation reward cannot be cross-checked against a code file.
\item Open-TeleVision~\cite{open_television_2024} (low). The paper's prose states 25k training iterations, lr 5e-5, batch size 45, but the code repo's example training command instead shows 50000 epochs and an explicit kl\_weight=10 not mentioned anywhere in the paper text.
  \\ \emph{Review:} R3 adversarial review: a behaviour-cloning paper with no reward; it compared paper iterations to a README example's epochs
\item Being-H0~\cite{being_h0_2025} (high). The actual downstream action-head module referenced in \S{}4.3 (learnable query tokens, proprioceptive projector f\_p, regression head f\_r) is not present in the parsed code; the repo's own TODO list marks 'Training code and scripts' as unreleased, so the action-chunk-length=16 value and exact proprioception-vector fields come only from inference/eval CLI examples, not a verifiable implementation.
\item $\pi_{0.5}$~\cite{pi05_2025} (high). The paper describes a hybrid discrete-FAST-token-pretrain-then-flow-matching-post-train recipe (Eq. 1, alpha: 0 to 10), but the released repo states plainly it 'currently only support[s] the flow matching head for both pi0.5 training and inference.'
\item $\pi^{*}_{0.6}$~\cite{pistar06_2025} (high). The general openpi framework (pi0/pi0.5/FAST, flow-matching action head, ALOHA/DROID/LIBERO adapters) is released at commit 215abfb2, but the RECAP-specific pieces described in the paper -- the advantage-conditioning input, the distributional value function, the RECAP training loop (Algorithm 1), and the espresso/laundry/box-assembly checkpoints and datasets -- are not present in the parsed repo.
\item Being-H0.5~\cite{being_h05_2026} (high). Released code (commit e12ac44f) exposes only benchmark/inference/config scaffolding (policy.py, beingh\_policy.py, dataset-transform configs). no MoT/MoF model definition, no rectified-flow action-expert source, and no config for any dexterous-hand embodiment; the only post-train configs present (libero\_robocasa.yaml, libero\_all.yaml, robocasa\_human.yaml, so101\_example.yaml) are all gripper-only, so the dexterous-hand claims cannot be cross-checked against released code.
\end{itemize}

\textbf{parse-limitation, 14 rows.}
\begin{itemize}
\item DexPoint~\cite{dexpoint_2022} (low). The released reward reshapes the paper's reach term into an inverse-distance form, 1.0/(0.06+finger\_object\_dist) in place of the plain distance of Eq. 5, and adds a 2 cm lift-threshold bonus, a target-distance term and an IK controller-tracking penalty that Eq. 5 does not list.
  \\ \emph{Artifact:} \href{https://github.com/yzqin/dexpoint-release/blob/17f1e238bb120d92baabe8207494d48476fea289/dexpoint/env/rl_env/relocate_env.py}{\texttt{relocate\_\allowbreak{}env.\allowbreak{}py}} (\texttt{Allegro\allowbreak{}Relocate\allowbreak{}RLEnv.\allowbreak{}get\_\allowbreak{}reward}), at the revision the link pins.
  \\ \emph{Review:} Reclassified from contradiction to parse-limitation after the corresponding author replied, 2026-09-20. Three parts of the original claim do not stand. The lift term is not a departure: object\_lift is already h\_current minus h\_init, set against a resting height fixed at reset, so 10*clip(object\_lift,0,0.2) is the paper's formula with the stated weight plus a saturation, and this survey's parse captured the reward body but not the line defining object\_lift, so the claim rested on material the snapshot did not contain. The rotation bonus is inactive at the paper's settings, since rotation\_reward\_weight defaults to zero and the term is multiplied by it, so it contributed nothing to the reported numbers. The reach-term difference is real but is version skew rather than a contradiction: the inverse form is a later, lower-variance revision and the paper's reported variance is that of the plain-distance formula the paper states. The remaining extra terms have stated purposes, the 2 cm bonus blocking a height reward hack on meshes with no flat resting surface and the IK penalty correcting a SAPIEN-specific controller mismatch. The reply window closed on 10 October 2026 with no further correspondence.
\item ArtiGrasp~\cite{artigrasp_2023} (low). paper's two regulariser weights (w\_rh=0.5, w\_ro=0.2) correspond to five separate velocity penalties in code (-0.5,-0.2,-0.5,-0.5,-0.3); the fingertip weight 12.0 and lambda 5.0 in Table 6 do not appear in the yaml; reward weights differ between curriculum phase-1 and phase-2 configs though Table 6 reports only one set
  \\ \emph{Review:} R3 adversarial review: the reward body is C++ and absent from the parse; the weight split is the paper's documented curriculum
\item Diffusion Policy~\cite{diffusion_policy_2023} (high). the paper's core loss equations (DDPM training loss, EBM/InfoNCE, score-matching) are unrecovered images in the parsed text, and the parsed copy is a signature-only API map, so compute\_loss's body could not be verified from code; only the exact shipped noise-scheduler config (DDPMScheduler, 100 train steps) is confirmed.
\item Eureka~\cite{eureka_2023} (low). The released repo's default eureka/cfg/config.yaml ships iteration:1, sample:3 (a fast-test default), not the paper's reported experimental configuration of N=5 iterations $\times$ K=16 samples $\times$ 5 independent runs, and LLM-generated reward code itself is never checked into the repo (written at runtime to a gitignored outputs/ directory).
  \\ \emph{Review:} R3 adversarial review: the repo README documents the paper's defaults; the cited config block is a fast-test preset
\item UniDexGrasp++~\cite{unidexgrasp_pp_2023} (low). The paper's four named reward weights (0.5, 0.1, 2, 10) numerically match literals in the released compute\_hand\_reward, but the code never labels them with the paper's variable names, and the paper's own display equations were dropped from the PDF, so the correspondence is inferred rather than confirmed by either source.
\item BiDexHD~\cite{bidexhd_2024} (high). Table 5 weights (w\_r, w\_t, w1-w4, lambda\_w=0.12, lambda\_ft=0.48) cannot be matched to code reward lines because the code md truncates the reward function before stage-2/final sum; lambda\_w and lambda\_ft numerically match the code's grasp-condition thresholds (0.12, 0.12x4 fingers) rather than confirmed reward weights
\item GraspXL~\cite{graspxl_2024} (low). The code splits the paper's single r\_reg = -w\_h||u\_h||\textasciicircum{}2 - w\_o||u\_o||\textasciicircum{}2 into four separately-weighted terms (wrist vel, wrist qvel, object vel, object qvel), and the floating-phase object-velocity coefficient (-1.5) does not match the paper's stated w\_o=0.1, so the regularization weighting cannot be reconciled 1:1 with Table 9.
  \\ \emph{Review:} R3 adversarial review: the environment source is not in the parse and the note says the reward cannot be verified
\item HumanPlus~\cite{humanplus_2024} (medium). code implements a superset of legged\_gym reward terms (lin\_vel\_z, ang\_vel\_xy, orientation, base\_height, torques, dof\_vel, dof\_acc, action\_rate, collision, termination, dof/torque limits, tracking\_lin/ang\_vel, feet\_air\_time, stumble, stand\_still, feet\_contact\_forces, target\_jt) beyond the 8 terms documented in the paper's Table 1
\item Omnigrasp~\cite{omnigrasp_2024} (medium). The code's compute\_pregrasp\_reward\_time hard-codes w\_pos, w\_rot = 0.9, 0.1, silently overriding the rwd\_specs weight arguments passed into the function; separately, the paper's pre-grasp reward weights (w\_hp, w\_hr) are not resolvable from either the paper text or Appendix Table 7.
\item ResDex~\cite{resdex_2024} (medium). The base-policy and residual-policy task files implement different variants of compute\_hand\_reward (whether finger/hand-distance terms are included alongside goal\_hand\_rew/hand\_up/bonus differs), a difference the paper's Sec 4.1/4.4 text does not describe, and the released configs' env count (4096 default) and iteration counts (2500/5000) do not match the paper's stated 11,000 envs and 5,000/20,000 iterations.
\item DexGraspVLA~\cite{dexgraspvla_2025} (high). Hardware-interfacing code is explicitly withheld ("Due to intellectual property constraints, we are unable to open-source the hardware-related code"); only function signatures for the controller loss (compute\_loss, noise\_assignment) are present in the parsed copy, not their bodies, so the MSE/Immiscible-Diffusion loss cannot be verified against code.
\item DexTrack~\cite{dextrack_2025} (low). The paper gives one canonical reward-weight table (Table 3), but the released repo's task-config YAMLs contain several divergent, unreconciled reward-coefficient sets across AllegroHandTracking/\ldots{}TrackingGeneralist/other variants, and it is not identifiable from the parsed source which config produced the paper's headline Table 1 numbers.
\item DexVLA~\cite{dexvla_2025} (high). the parsed copy only carries class/function signatures, not bodies, for the diffusion head (modeling\_scaledp.py) and no config file with the $\alpha$ weight was captured; the paper's diffusion loss and denoising process cannot be verified against code bodies. only the multi-head ScaleDP signature and a DDIM fp32 patch are available, so none of the paper's tables can be reproduced from what was parsed.
\item ManipTrans~\cite{maniptrans_2025} (low). paper's PPO learning rate (5e-4) and env count (4096) differ from the shipped README/yaml defaults (2e-4, 8192)
  \\ \emph{Review:} R3 adversarial review: the cited values are a README override and an unused fallback; the repo's own config matches the paper
\end{itemize}

\endgroup

\section{The existing surveys, and what each covers}
\label{app:surveys}

Fourteen corpus entries are themselves surveys or engine-comparison studies.
Table~\ref{tab:surveys} in Section~I sets them on one set of columns, three levels to a cell,
recording what each work covers and not how well. This appendix is the prose behind those levels:
each glyph in that table coarsens one of the sentences below. Section~I states what this survey
adds over them and does not repeat it here.

The survey of An et al. \cite{an_dexil_survey_2025} is the closest in subject, covering imitation
learning for multi-fingered
hands by learning family, end-effector class and demonstration source. It gives reinforcement
learning no taxonomy, treats bimanual work as a single ``multi-agent'' subsection, compares no
simulator, and defines no evaluation metric. The survey of Welte et al. \cite{welte_iil_survey_2025}
finds only seven dexterous
works that use interactive imitation learning and carries a fifteen-hand commercial table, with no
bimanual section, no benchmark table and no contact modeling. The survey of Zhao et al.
\cite{zhao_sim2real_survey_2020}
supplies the standard sim-to-real split and predates GPU-parallel simulation.
The survey of Firoozi et al. \cite{firoozi_foundation_models_2023} has zero occurrences of bimanual,
tactile or in-hand.

The survey of Bai et al. \cite{bai_unified_manip_survey_2025} spans all of manipulation across 212
pages. Its Sec. 4.3 on
dexterous manipulation runs about 720 words, the third longest of its ten task subsections behind
grasping and quadrupedal manipulation, which is a real treatment and not a passing mention. The
difference is elsewhere. Its Sec. 1.2 lists dexterous manipulation among the topics that ``existing
surveys'' cover from ``narrower perspectives'' and defers it to two of them.

The survey of Zhao et al. \cite{zhao_dexhand_survey_2026} is the most recent hand-centred one,
with a 29-hand anatomy table and a task-by-paradigm taxonomy. Its Sec. IV-C names the
reference-versus-rollout split that Section~\ref{sec:intro} credits it with, and its Sec. III-F
is the one bimanual subsection, 16 cited works with no coordination analysis.

The review of Kaup et al. \cite{nine_physics_engines_review_2024} is the predecessor closest to
the engine comparison, and it reviews Brax, Chrono, Gazebo, MuJoCo, ODE, PhysX, PyBullet, Unity
and Webots for reinforcement learning research. It scores each on documentation, model and
environment creation, URDF and MJCF support, and readiness for multi-agent work. It runs no
benchmark of its own and says so in its Sec. V, that implementing the same scenarios across nine
engines ``goes beyond the scope of this paper''. Its running bodies are ant-and-humanoid RL
benchmarks and not hands, and it discusses no timestep, no friction model, no contact
formulation and no penetration.

The oldest rows carry the cost of the corpus. Okamura et al. and Piazza et al.
\cite{okamura_overview_2000,piazza_century_2019}, and Roa and Su\'arez
\cite{roa_suarez_grasp_quality_2015}, are behind publisher paywalls with no author-hosted copy found
on 2026-09-18, and Bicchi \cite{bicchi_hands_2000} is in the same position with no row at all. Ma
and Dollar \cite{ma_dollar_dexterity_2011} are a different case. The fetch that failed when its note
was written
succeeded afterwards, so a seven-page PDF is on disk with a recorded hash, and no note has been
read from it. All five are cited by metadata only and nothing in this survey describes their
contents. The open chapter \cite{bicchi_grasping_chapter_2001} overlaps the paywalled Bicchi paper
without being identical to it, so it is quoted in its own right.

\providecommand{\cvyes}{\textcolor{accent}{$\checkmark$}}
\providecommand{\cvpart}{\textcolor{accent}{$\circ$}}
\providecommand{\cvno}{\textcolor{black!40}{$\times$}}
\begin{table*}[!t]
\centering
\caption{What 14 predecessor surveys cover: \cvyes\ covered, \cvpart\ partly, \cvno\ not. 24 of 84 coverage cells are empty.}
\label{tab:surveys}
\footnotesize
\setlength{\tabcolsep}{2pt}
\setlength{\emergencystretch}{1.5em}
\begin{tabular}{>{\raggedright\arraybackslash\hspace{0pt}}p{120pt}>{\raggedleft\arraybackslash}p{18pt}>{\centering\arraybackslash}p{57pt}>{\centering\arraybackslash}p{57pt}>{\centering\arraybackslash}p{57pt}>{\centering\arraybackslash}p{57pt}>{\centering\arraybackslash}p{57pt}>{\centering\arraybackslash}p{57pt}}
\toprule
\rowcolor{wash} \hdr{survey} & \hdr{yr} & \hdr{dexterous scope} & \hdr{taxonomy} & \hdr{bimanual} & \hdr{hardware} & \hdr{evaluation} & \hdr{gaps named} \\
\midrule
\multicolumn{8}{@{}l}{\textbf{Surveys of the field}} \\[1pt]
Zhao et al.~\cite{zhao_dexhand_survey_2026} & 2026 & \cvyes & \cvyes & \cvpart & \cvyes & \cvpart & \cvyes \\
\rowcolor{wash} An et al.~\cite{an_dexil_survey_2025} & 2025 & \cvyes & \cvpart & \cvpart & \cvpart & \cvpart & \cvyes \\
Bai et al.~\cite{bai_unified_manip_survey_2025} & 2025 & \cvpart & \cvyes & \cvpart & \cvpart & \cvpart & \cvyes \\
\rowcolor{wash} Welte et al.~\cite{welte_iil_survey_2025} & 2025 & \cvpart & \cvyes & \cvpart & \cvyes & \cvno & \cvyes \\
Firoozi et al.~\cite{firoozi_foundation_models_2023} & 2023 & \cvno & \cvpart & \cvno & \cvno & \cvpart & \cvyes \\
\rowcolor{wash} Zhao et al.~\cite{zhao_sim2real_survey_2020} & 2020 & \cvno & \cvyes & \cvno & \cvpart & \cvno & \cvyes \\
\addlinespace[2pt]
\multicolumn{8}{@{}l}{\textbf{Simulator and contact-model studies}} \\[1pt]
NVIDIA Isaac Sim~\cite{isaac_sim_2026} & 2026 & \cvno & \cvpart & \cvpart & \cvno & \cvno & \cvyes \\
\rowcolor{wash} Kaup et al.~\cite{nine_physics_engines_review_2024} & 2024 & \cvno & \cvyes & \cvno & \cvno & \cvpart & \cvyes \\
Le Lidec et al.~\cite{contact_models_comparison_2023} & 2023 & \cvno & \cvyes & \cvno & \cvpart & \cvyes & \cvyes \\
\rowcolor{wash} Erez et al.~\cite{physics_engine_comparison_2015} & 2015 & \cvno & \cvno & \cvno & \cvpart & \cvyes & \cvyes \\
\addlinespace[2pt]
\multicolumn{8}{@{}l}{\textbf{Sources not obtained, or on disk and never read}} \\[1pt]
Piazza et al.~\cite{piazza_century_2019} & 2019 & \na & \na & \na & \na & \na & \na \\
\rowcolor{wash} Roa and Su\'arez~\cite{roa_suarez_grasp_quality_2015} & 2015 & \na & \na & \na & \na & \na & \na \\
Ma and Dollar~\cite{ma_dollar_dexterity_2011} & 2011 & \na & \na & \na & \na & \na & \na \\
\rowcolor{wash} Okamura et al.~\cite{okamura_overview_2000} & 2000 & \na & \na & \na & \na & \na & \na \\
\bottomrule
\end{tabular}
\end{table*}

A cell is a level and not a score: \cvyes\ where the work gives the topic a structure of its own, a section, a table or a measurement it defines; \cvpart\ where the topic is there without one, a subsection with no analysis, a handful of mentions, or a metric named and never defined; \cvno\ where it is absent or put outside the work's scope. Each level coarsens one sentence read from the work itself, and it is printed only while the quotation behind that sentence is still recorded, so a glyph cannot outlive its evidence. Appendix~\ref{app:surveys} prints the prose those levels coarsen. Three of the fourteen sources could not be obtained and one is on disk and has never been read, which is why their cells are empty; they are grouped as such rather than scored, because this survey knows nothing of them beyond a title and a DOI. The \emph{gaps named} column is uniform by design: every predecessor names gaps, and what none of them attaches to a gap is a measurement.

\section{Where the protocol's counts come from}
\label{app:protocol}

Every count in Table~\ref{tab:protocol} is derived below, and each axis is derived for the
statistic that axis actually reports, not by one convention applied to all of them: a single
rate takes a Wilson half-width, a matched comparison takes McNemar, a ratio takes the standard
error of the log ratio, and a correlation takes the Fisher-z interval.
Section~\ref{subsec:protocol} states what these derivations conclude. This survey re-ran no
method, so every count here rests on an interval width, a power calculation or another paper's
measurement, and never on a measurement of our own.

Fix the width first, then read off the count. Take a 95 percent Wilson interval on a single
reported rate, at the worst case of $\hat{p} = 0.5$. Writing the half-width of \eqref{eq:wilson} as
\begin{equation}
w(n,\hat{p}) = \frac{z\sqrt{4n\hat{p}\,(1-\hat{p}) + z^{2}}}{2\,(n + z^{2})},
\label{eq:halfwidth}
\end{equation}
the count a target width $w_{0}$ demands is
$n_{\min}(w_{0}) = \min\{\,n : w(n,0.5) \le w_{0}\,\}$, which gives
\begin{equation*}
\begin{array}{lrrrr}
\toprule
\text{half-width } w_{0} \text{ (points)} & 20 & 15 & 10 & 5\\
\midrule
n_{\min} \text{ (trials)} & 21 & 39 & 93 & 381\\
\bottomrule
\end{array}
\end{equation*}
Ten points is the coarsest width at which a single rate is worth printing, so the absolute-rate
minimum is 93, rounded to 100. At 100 trials a reported 80 percent has an interval of 71 to 87
percent, and a reported 50 percent has 40 to 60. A comparison is a different question and a harder
one: two rates each carrying $\pm 10$ points do not resolve a 10-point difference between them,
because the difference's standard error is larger by a factor of $\sqrt{2}$, so the width argument
sets a floor on what is worth reporting and not on what can be compared.

For the A/B comparison the relevant calculation is power, and the design is paired.
Table~\ref{tab:protocol} matches initial conditions by image overlay and interleaves the two
policies in one session, so the unit is a matched pair and the count follows McNemar, which depends
on the discordance rate. The share of initial conditions on which the two policies disagree, and
not on the two rates alone. Write $\delta = |p_{1} - p_{2}|$ for the gap between the two success
rates, $\pi_{d}$ for that discordance rate, $z_{1-\alpha/2} = 1.96$ at $\alpha = 0.05$ and
$z_{1-\beta} = 0.84$ at 80 percent power; then the number of matched pairs is
\begin{equation}
n_{\mathrm{pairs}} =
  \frac{\bigl[\,z_{1-\alpha/2}\sqrt{\pi_{d}}
              + z_{1-\beta}\sqrt{\pi_{d} - \delta^{2}}\,\bigr]^{2}}{\delta^{2}}.
\label{eq:mcnemar}
\end{equation}
To separate 50 from 70 percent, which is $\delta = 0.20$, at $\alpha = 0.05$ with 80 percent power,
\eqref{eq:mcnemar} asks per arm for
\begin{equation*}
\begin{array}{lrrrr}
\toprule
\text{discordance } \pi_{d} & 0.2 & 0.3 & 0.4 & 0.5\\
\midrule
n_{\mathrm{pairs}} & 37 & 57 & 77 & 96\\
\bottomrule
\end{array}
\end{equation*}
The protocol assumes 0.3 and asks for 57, and states the sensitivity, because
0.5 is the discordance the same two rates produce when the pairing buys nothing, and at that value
the paired count returns to the 93 per arm an unpaired test would need. The saving from pairing is
real but smaller than the pair counts suggest, since a pair costs two rollouts: 57 pairs is 114
rollouts against 186. The 93 above is the same integer as the half-width calculation two paragraphs
up, which is a coincidence of the worst-case arithmetic and not a second derivation of it.

One hundred is a cap and not a bill, because on a graded score a sequential test reached its
decision in 12 to 36 paired hardware trials in \cite{beyond_binary_success_2026}. At 30 rollouts
a continuous score already carries a half-width of $\pm 0.36$ standard deviations, which is why
a graded score can stop where a binary one cannot. A cell that stops early does not report a
Wilson interval. Optional stopping breaks the coverage of a fixed-$n$ interval, which is the
reason Snyder et al. \cite{beyond_binary_success_2026} and Badithela et al. \cite{suresim_2025}
use anytime-valid betting intervals and not Wilson, so Table~\ref{tab:protocol} asks a cell run
to a fixed 100 for a Wilson interval and a cell stopped early for a confidence sequence, and
never for both. Intervals are also marginal and not simultaneous. For example, evaluating twelve
methods on seven axes produces 84 intervals. At 84 independent 95 percent intervals, four
excursions are expected by construction, so a paper comparing $k$ policies on $m$ tasks corrects
its $k(k-1)/2$ pairwise tests to a global 95 percent level, as the TRI LBM Team
\cite{lbm_careful_examination_2025} does, or says its intervals are not simultaneous.

Simulation is cheap, so simulated cells take 200 episodes, giving a 6.9-point half-width.
Perturbation axes are screened and not certified, and 40 per axis buys a 15-point half-width on
each axis's own absolute rate, which is enough to rank the axes and pick the two worst for
hardware. It is not enough for the ratio to the anchor. At 40 trials in each arm, a fall from a
0.50 anchor to 0.30 is a ratio of 0.60 with a 95 percent interval of 0.34 to 1.06, which
contains 1: at the screening count you cannot establish that the perturbation hurt at all.
Certifying that same drop takes 101 per arm at 80 percent power, or about 50 for an interval
that merely excludes 1, so Table~\ref{tab:protocol} now asks for absolute rates with their own
intervals at 40, reports the ratio without an interval, and prescribes 101 before any claim that
a named axis hurt.

For unseen objects the resampling unit is the object and not the trial, so 20 objects at 5 trials
each gives 100 trials and an object-level half-width near 20 points. That 20 points is the Wilson
width at $n = 20$ and it treats each object's outcome as a single Bernoulli draw, which the five
within-object trials are not. It is the right order of magnitude and the assumption belongs in the
cell. A 10-point claim about an object distribution needs about 93 objects. Seven of the 39 rows
that state an unseen count reach that: 225 in \cite{bimangrasp_2024}, 241 in \cite{resdex_2024}
and \cite{unidexgrasp_2023}, 360 in \cite{dexgraspvla_2025}, 500 in \cite{dexmv_2021}, 2029 in
\cite{clutterdexgrasp_2025} and 503409 in \cite{graspxl_2024}. For a continuous score the
half-width is 1.96 standard deviations over the square root of the count, $1.96\,s/\sqrt{n}$, so
100 rollouts give $\pm 0.20$ standard deviations, and the unit is the rollout because frames
within one are correlated.

The transfer axis is the one where 100 is least defensible. On 100 matched pairs a measured
correlation of 0.70 carries a Fisher-z interval of 0.58 to 0.79, a half-width of about 0.10 that
130 pairs would be needed to guarantee. That is enough to establish that a simulator tracks
reality at all, and it is not enough to separate \cite{suresim_2025}'s useful regime from its
marginal one, since those differ by about 0.11 in correlation. Both limits come within 0.05 of
the estimate only at about 457 pairs. Table~\ref{tab:protocol} states which of the two decisions
each count supports.

\balance
\bibliographystyle{IEEEtran}
\bibliography{refs}

\end{document}